\documentclass[final]{elsarticle}
\usepackage{caption}
\usepackage{subcaption}
\usepackage{amsmath,amsthm,amssymb,amsfonts}
\usepackage{xcolor}
\usepackage{caption}
\usepackage{graphicx}
\usepackage{subcaption}
\usepackage{mdframed}
\usepackage{pifont}
\usepackage[shortlabels]{enumitem}
\usepackage{indentfirst}
\usepackage{float}
\usepackage{bbm}
\usepackage{threeparttable}
\usepackage{svg}
\usepackage{lipsum}
\usepackage{cite}
\usepackage{hyperref}
\usepackage{tcolorbox}
\usepackage[ruled,vlined]{algorithm2e}
\SetAlgoSkip{1.5ex} 
\SetAlgoInsideSkip{1.5ex}
\usepackage{tikz}
\usetikzlibrary{arrows.meta,positioning,calc,fit,shapes.misc}


\usepackage{mdframed}
\hypersetup{
    colorlinks=true,
    linkcolor=blue,
    filecolor=magenta,      
    urlcolor=blue,
    citecolor=blue
}
\usepackage{geometry}

\graphicspath{{Images/}}

\newcommand{\RR}{\mathbb{R}}

\newcommand{\bfx}{\mathbf{x}}
\newcommand{\bfu}{\mathbf{u}}

\newcommand{\bftheta}{\boldsymbol \theta}

\newcommand{\GG}{\mathcal{G}}
\newcommand{\FF}{\mathcal{F}}
\newcommand{\OO}{\mathcal{O}}
\newcommand{\CC}{\mathcal{C}}

\usepackage{booktabs}

\newcommand{\rf}[1]{{\color{black}  #1}} 
\newcommand\norm[1]{\left\lVert#1\right\rVert}
\makeatletter
\def\ps@pprintTitle{%
  \let\@oddhead\@empty
  \let\@evenhead\@empty
  \let\@oddfoot\@empty
  \let\@evenfoot\@empty
}
\makeatother

\begin{document}
\title{DeepONet-accelerated Bayesian inversion for moving boundary problems}
\author[addr1]{M. A. Iglesias}
\ead{Marco.Iglesias@nottingham.ac.uk}

\author[addr1]{M. E. Causon}

\author[addr2]{M. Y. Matveev}

\author[addr2]{A. Endruweit}

\author[addr1]{M.V. Tretyakov}
                 
\address[addr1]{School of Mathematical Sciences, University of Nottingham, UK}
\address[addr2]{Composites Group, Faculty of Engineering, University of Nottingham, UK}

\begin{abstract}
 This work demonstrates that neural operator learning provides a powerful and flexible framework for building fast, accurate emulators of moving boundary systems, enabling their integration into digital twin platforms. To this end, a Deep Operator Network (DeepONet) architecture is employed to construct an efficient surrogate model for moving boundary problems in single-phase Darcy flow through porous media. The surrogate enables rapid and accurate approximation of complex flow dynamics and is coupled with an Ensemble Kalman Inversion (EKI) algorithm to solve Bayesian inverse problems. 

The proposed inversion framework is demonstrated by estimating the permeability and porosity of fibre reinforcements for composite materials manufactured via the Resin Transfer Moulding (RTM) process. Using both synthetic and experimental in-process data, the DeepONet surrogate accelerates inversion by several orders of magnitude compared with full-model EKI. This computational efficiency enables real-time, accurate, high-resolution estimation of local variations in permeability, porosity, and other parameters, thereby supporting effective monitoring and control of RTM processes, as well as other applications involving moving boundary flows. Unlike prior approaches for RTM inversion that learn mesh-dependent mappings, the proposed neural operator generalises across spatial and temporal domains, enabling evaluation at arbitrary sensor configurations without retraining, and represents a significant step toward practical industrial deployment of digital twins.

\end{abstract}


\maketitle

\section{Introduction} \label{sec: Intro}

Moving boundary problems in flow through porous media arise in numerous applications, including tissue growth in biomedical modelling \citep{Goriely2017}, composites manufacturing \citep{Advani}, and groundwater hydrology related to oil and gas recovery and geothermal power production \citep{hydro,Woods2015}. These problems are frequently modelled as single-phase flow of a viscous liquid through a porous medium with homogenised properties, governed by Darcy’s law \citep{Darcy}. Methods for predictive simulation of moving boundary problems for a prescribed geometry, material parameters, and initial and boundary conditions are well established and are commonly referred to as the \emph{forward problem}. Such problems are typically solved using implementations (e.g. \citep{Zenodo}) of the Control Volume Finite Element Method (CVFEM) \citep{Bruschke:1990ii,Advani,MichaelMinho}.
In many practical settings, however, the relevant properties of the porous medium, in particular porosity and permeability, exhibit stochastic spatial variability, and boundary conditions are uncertain. As a result, the inputs to the forward problem may deviate from the true system, leading to discrepancies between experimental observations and model predictions. This motivates the formulation of the corresponding \emph{inverse problem}, in which unknown material parameters are estimated from in-process sensor data. These data typically consist of sparse and noisy measurements of fluid pressure and the moving boundary location.

\subsection{Resin Transfer Moulding}

As a case study of industrial relevance, a moulding process for manufacturing fibre-reinforced composites is considered here. Combining an exceptional strength-to-weight ratio with long fatigue life and the ability to form complex geometries, composites are mainly used in lightweight structural applications across the aerospace, automotive, and marine sectors \citep{Advani}. Among a variety of composites manufacturing processes, Resin Transfer Moulding (RTM) has emerged as a common choice, as it enables the production of components at relatively high rate and low cost (compared to other processes). The process involves placing a dry fibrous reinforcement (typically from glass or carbon fibre, but also from natural fibres such as flax or hemp) in a rigid mould tool. After the tool is closed, a liquid thermoset resin (such as epoxy or polyester) is injected under a flow-driving pressure gradient, permeating the porous reinforcement. Once the initially dry reinforcement is completely impregnated, the part is left to cure before being demoulded, trimmed, and inspected for quality assurance in accordance with its end-use requirements. Process simulations are implemented to predict the component quality, estimate the process cycle time, and optimise the tool lay-out. Usually, the components can be treated as thin shell-like structures, meaning that through-thickness effects can be ignored in the simulations.

The mechanical properties of a component made through RTM are highly sensitive to the completeness and uniformity of reinforcement impregnation during resin injection. In practice, stochastic fibre arrangements, deformation of the fibre structure introduced during preforming of the initially flat sheet of reinforcement to a 3D component shape, and \textit{race-tracking phenomena} (i.e. uncontrolled resin flow in gaps along the mould edges) \citep{Advani3,Andreas,Andreas2,Matveev2,Long2} promote irregular flow front propagation which may result in partially impregnated parts that are typically scrapped \citep{Advani,Long2}. To mitigate the effect of these issues on the component quality, there is considerable interest in developing methods capable of solving the inverse problem associated with RTM rapidly, ideally in real time, to enable active control of flow front propagation during the manufacturing process. Furthermore, achieving real-time inversion would facilitate the integration of RTM with a digital twin, allowing for non-destructive evaluation to be performed concurrently with the injection process. In this paper, a highly efficient inversion framework for moving boundary problems is developed using recent advances in neural operator learning \citep{lu2021learning}.

\subsection{Machine Learning for inversion in RTM.}

For inversion in RTM that is rapid enough to allow active process control, application of Machine Learning (ML) is essential. The application of ML to improve the RTM process dates back to the early 2000s, when methods such as decision trees \citep{Advani4} and genetic algorithms \citep{Advani6,Advani7} were employed to identify and mitigate irregular resin flow patterns caused by race tracking. However, these methods demonstrated limited generalisability beyond the specific scenarios on which they were trained. The advent of the third wave \citep{goodfellow2016deep} of machine learning has since renewed attention to data-driven approaches in RTM, fuelled by advances in deep neural networks, graphics processing units (GPUs), and the availability of high-fidelity simulation and experimental data. These developments have enabled models capable of learning complex flow behaviours and inferring unknown material properties directly from in-process measurements.

A series of studies applied deep learning to infer reinforcement permeability from fluid pressure measurements recorded during injection. The initial work used a Convolutional Neural Network (CNN) \citep{goodfellow2016deep} to estimate rectangular-shaped defects \citep{González} in the reinforcement, which was later extended to capture race tracking \citep{González2}. More recently, coupled encoder–decoder architectures have been proposed, in which one network predicts the location and severity of race-tracking channels from pressure data, while a second network emulates the resin injection process using the inferred permeability field. Together, these models form a digital twin capable of reproducing flow-front evolution in the presence of race tracking \citep{González4}.

In other studies, images capturing flow front propagation (acquired using a transparent mould) have been used to infer the anisotropic permeability fields of fibre reinforcements. In particular, Physics-Informed Neural Networks (PINNs) \citep{PINNs} have been employed to invert images of flow fronts in radial injection scenarios to estimate permeability fields that are then used to predict resin flow front evolution \citep{Hanna}. A comparative investigation evaluated the performance of CNNs, transformers and convolutional long short-term memory architectures for predicting fibre volume content and permeability tensors from injection images \citep{stieber2023inferring}. The authors also demonstrated that enriching simulation-trained models with real experimental data through transfer learning can enhance predictive accuracy.

While these methods have demonstrated success in numerous cases, they do not resolve the ill-posedness of inverse problems \citep{Kaipio,Stuart}. In particular, such problems are typically characterised by: (i) non-uniqueness, in that multiple permeability and porosity fields can reproduce the same observed data; and (ii) instability, meaning that small perturbations (or noise) in the data can lead to disproportionately large variations in the inferred permeability or porosity fields. Consequently, purely deterministic estimates of permeability fail to capture the fundamental uncertainty inherent in the solution.

\subsection{Bayesian inversion in RTM}

The Bayesian approach to inverse problems \citep{Stuart} provides a framework for addressing the aforementioned ill-posedness issues. In this approach, the reinforcement's permeability and porosity are modelled as random fields, initially through prior distributions reflecting design specifications and their anticipated variability, which are subsequently updated, or conditioned, on in-process measurements that are themselves corrupted by random noise. To circumvent the computational burden of sampling the full posterior via Markov chain Monte Carlo (MCMC) methods, which typically requires upwards of $O(10^5)$ forward simulations, the ensemble Kalman inversion (EKI) algorithm \citep{MarcoYang} has been found applicable to approximate the posterior at a fraction of the computational cost, for example $O(10^3$–$10^4)$ simulations. In addition to its computational efficiency, EKI is derivative free and does not require access to the linearised forward model or its adjoint, which are often unavailable for complex moving boundary problems. EKI has been successfully employed for RTM applications \citep{Iglesias_2018,Matveev}, enabling probabilistic identification of local variations in material properties in both flat and preformed reinforcements without imposing assumptions on their geometry, location, or number, for both virtual and laboratory experiments.

Despite the computational advantage of EKI over MCMC, achieving real-time estimation of local material properties has remained challenging for RTM applications. Recent developments, however, have demonstrated that this limitation can be overcome by replacing the costly forward solver with a trained neural surrogate \citep{MC}. By emulating the simulator response to permeability and porosity fields, this surrogate-accelerated EKI framework has shown strong performance in two-dimensional settings, validated against both virtual and laboratory experiments. While even simple neural architectures with a single hidden layer were sufficient to uncover defects in reinforcements, this relied on a low-dimensional representation of the permeability and porosity fields. The resulting estimates were confined to a fixed spatial partition, comprised of a 9×9 grid of `central zones' and four race tracking zones, which limited the framework's flexibility and expressiveness. To overcome this constraint, the purpose of this paper is to train a neural operator surrogate capable of handling a much broader range of input representations for the material properties.

\subsection{Neural operators}\label{subsec:neural_operators}

Neural operators are a class of deep learning architectures that learn maps between function spaces and therefore provide discretisation-invariant surrogates for parametric PDE solution operators, enabling evaluation at arbitrary query points once trained \citep{kovachki2023neuraloperator}. Prominent architectures include the Fourier Neural Operator (FNO), which performs global convolution in the Fourier domain and has been shown to scale effectively to high-dimensional parametric PDE families \citep{li2021fourierneuraloperatorparametric}, and the Deep Operator Network (DeepONet), which represents operators via coupled branch and trunk networks \citep{lu2021learning} and has motivated a broad set of variants for complex multi-input settings. In inverse problems, neural operators are increasingly used as forward surrogates to accelerate Bayesian inference, including MCMC and geometric MCMC, by amortising repeated solves of expensive forward maps \citep{kaltenbach2023semisupervised,cao2024dino}. Existing applications of neural operator surrogates for inverse problems have focused predominantly on fixed-domain settings, including elliptic and parabolic PDEs arising in elasticity, diffusion, and related problems. More recently, neural operators have also been applied to problems involving evolving interfaces, such as free-surface and multiphase flows \citep{LANTERI2025106773}. Nevertheless, applications to Bayesian inverse problems governed by moving boundary PDEs remain comparatively limited. In particular, the systematic use of operator-learning surrogates for Bayesian inversion in porous-media moving boundary problems has not yet been extensively explored.

\subsection{Contributions}\label{subsec:ContribOutline}

This work demonstrates that neural operators, and in particular the DeepONet architecture \citep{lu2021learning,LI2025128675,Lu2022}, can be used to construct efficient and accurate surrogates for moving boundary problems, enabling a substantial acceleration of the corresponding inverse problems. The proposed framework employs DeepONet surrogates within a surrogate-accelerated Bayesian inversion setting to achieve real-time estimation of spatially varying permeability and porosity fields in RTM using in-process measurements. The considered material properties are modelled as heterogeneous random fields that may exhibit sharp discontinuities associated with the possible presence of defects in the reinforcement and/or race-tracking phenomena, where permeability can vary by several orders of magnitude. In addition to identifying material parameters, the framework also infers the geometry of defective regions.
Although training is performed on a fixed mesh, the DeepONet surrogate can be evaluated at arbitrary spatial locations, enabling its use with any sensor configuration without retraining.


The proposed framework is validated using both synthetic data and real laboratory experiments, showing that the surrogate not only generalises beyond training but also remains robust under experimental conditions. The numerical and experimental results demonstrate that the framework can reduce the runtime of EKI posterior approximations from hours to seconds, while preserving accuracy in characterising both the heterogeneous material properties and the defective regions. As a result, characterisation of local permeability and porosity can be performed in real time and used for active process control, non-destructive examination of produced parts, and potentially for certification purposes.

Although the focus of this paper is on real-time inverse problems, the demonstrated DeepONet surrogates are also well suited for other computationally intensive RTM tasks, including rapid evaluation for optimal process design and both passive and active control strategies (see e.g. \citep{ourcontrol22} and references therein), such as optimising inlet configurations, inlet opening and closing sequences, and pressure schedules to minimise deviations of the manufactured component from its design specifications.

The remainder of the manuscript is organised as follows. In Section \ref{sec:mainInversion}, the inverse problem for the moving boundary model, with a focus on RTM, is introduced. In Section~\ref{sec:virtual_EKI}, synthetic experiments using EKI with the \emph{full model} (i.e. simulations of the moving boundary problem) are presented to illustrate the associated computational cost. This motivates the need for a DeepONet surrogate, which is introduced in Section \ref{sec: Surrogate}. In Section~\ref{subsec: SurrogateEKI}, the inversion problem is revisited using the DeepONet surrogate for both synthetic and real data from laboratory experiments. Finally, conclusions and directions for future work are presented in Section \ref{sec: Conclusions}. 

\section{Inversion framework for resin infusion in RTM}\label{sec:mainInversion}


\subsection{The forward operator for resin injection}\label{sec:forward_operator}

The reinforcement spans the domain $D$, whose boundary, $\partial D$, is partitioned into an inlet, $\partial D_I$, an outlet, $\partial D_0$, and impermeable sections, $\partial D_N$ (on which the normal flow velocity vanishes). Here, the permeability is assumed to be isotropic, so that it can be described by a scalar field denoted by $K(\bfx)$. The reinforcement porosity is denoted by $\phi(\bfx)$. Both $K(\bfx)$ and $\phi(\bfx)$ are spatially varying functions defined for every $\bfx\in D$. 

Resin, with viscosity $\mu$, is injected into the domain $D$ at an inlet pressure, $p_I(t)$, that is uniform along the inlet boundary $\partial D_I$. On the moving flow front, the pressure is maintained at a constant value, $p_0$, which for simplicity is taken as $p_0 = 0$ Pa (corresponding to vacuum in the mould tool). The flow is assumed to be isothermal, and curing of the resin is neglected, so that the viscosity, $\mu$, remains constant throughout the process. As the injection proceeds, the resin fills a time-dependent subdomain, $D(t) \subseteq D$, enclosed by the inlet, $\partial D_I$, the evolving front, $\Upsilon(t)$, and the corresponding parts of the impermeable boundary, $\partial D_N$.

Within the saturated portion $D(t)$ of the domain $D$, the phase-averaged flow velocity field is governed by Darcy’s law, which combined with the continuity equation yields

\begin{align}
    -\nabla \cdot \Bigg[\frac{K(\bfx)}{\mu}\nabla p(\bfx,t)\Bigg] = 0, \ \ \ \bfx \in D(t), \label{eq: incompressibility}
\end{align}
where $p$ represents the resin pressure. To complete the formulation, the governing equation is accompanied by the following initial and boundary conditions:
\begin{align}
    &V(\bfx,t) = -\Bigg[\frac{K(\bfx)}{\mu\,\phi(\bfx)}\nabla p(\bfx,t)\Bigg] \cdot \textbf{n}_\Upsilon(\bfx,t),\,\,\ \bfx \in \Upsilon(t),\ \ \,\, t \geq 0, \label{eq:PDE_start}\\
    &p(\bfx,t) = p_I(t), \,\,\bfx \in \partial D_I,\ \ \,\, t \geq 0, \notag \\
    &\nabla p(\bfx,t) \cdot \textbf{n}(\bfx) = 0, \,\, \bfx \in \partial D_N,\ \ \,\, t \geq 0, \notag \\
    & p(\bfx,t) = p_0, \,\, \bfx \in \Upsilon(t),\ \,\, 
    p(\bfx,t) = p_0, \,\, \bfx \in \partial D_0,\ \ \,\, t > 0, \notag \\
    & p(\bfx,0) = p_0, \,\, \bfx \in D,\,\,\,
    \Upsilon(0) = \partial D_I, \notag 
\end{align}
where $\textbf{n}_{\Upsilon}(\bfx,t)$ and $\textbf{n}(\bfx)$ denote the unit outward normals to the moving and impermeable boundaries, respectively, and $V(\bfx,t)$ denotes the component of the velocity of the moving boundary, $\Upsilon(t)$, along $\textbf{n}_{\Upsilon}(\bfx,t)$. In practice, the governing equations \eqref{eq: incompressibility} to \eqref{eq:PDE_start} are solved over a finite time horizon $[0,T]$, corresponding to the duration of the filling process.

In previous studies \citep{Iglesias_2018,Matveev,MC}, the inlet pressure was usually prescribed as a constant $p_{I}(t)= P_I$, since it can, in principle, be fully controlled. In laboratory and manufacturing settings, however, there is typically a short delay between opening the gates for resin to flow into the preform and the inlet pressure reaching a steady-state pressure. Consequently, a time-dependent representation of the inlet pressure is necessary for an accurate description of the flow, and hence for accurate inversion. The following relation is adopted:
\begin{eqnarray}\label{eq:press}
p_{I}(t)=\chi P_{I}+(P_{I}-\chi P_{I})\Big(1-e^{-(t/\lambda)^\beta}\big),
\end{eqnarray}
where $P_{I}$ denotes the asymptotic inlet pressure; $\chi\in [0,1]$  is a dimensionless parameter representing the initial fraction of $P_{I}$ at $t=0$; and $\lambda>0$ and $\beta>0$ are calibration parameters controlling the time scale and shape of the pressure rise, respectively. This formulation captures the gradual increase of the inlet pressure observed in practice and provides a more realistic description of resin flow propagation.

Given a fixed tool geometry, the Darcy flow model expressed in Eqs.~\eqref{eq: incompressibility} to \eqref{eq:press} defines a mapping from the inputs $\bfu_{\text{FM}}$ to the solution of the moving–boundary problem:
\begin{align}
\bfu_{\text{FM}}(\bfx)
:= \big(\log K(\bfx),\, \phi(\bfx),\, \mu,\, P_I,\, \lambda,\, \beta,\, \chi\big)
\longmapsto \big(p(\bfx,t),\, \Upsilon(t)\big).
\label{eq:fwd_inputs}
\end{align}
The permeability is parametrised on the natural log scale, using $\log K(\bfx)$ in place of $K(\bfx)$, since the log parametrisation enforces positivity, accommodates large variations commonly observed in permeability fields, and simplifies statistical modelling (see Section \ref{subsec:BayesApproach}). 

For a fixed saturated domain, the pressure equation is a uniformly elliptic boundary value problem and is well posed under standard assumptions on the permeability field. 
For the considered moving boundary problem, involving strongly heterogeneous and possibly discontinuous permeability and porosity fields, the existence and uniqueness question for its solution remains open and challenging. Analytical solutions are only available in the highly idealised one-dimensional setting \citep{Advani,MichaelMinho}.

\subsection{The forward operator}\label{sec:fop}
Since the inlet boundary $\partial D_I$ and the impermeable boundary $\partial D_N$ are known, the output $\big(p(\bfx,t),\, \Upsilon(t)\big)$ in Eq. \eqref{eq:fwd_inputs} can equivalently be expressed as $\big(p(\bfx,t),\, D(t)\big)$. To obtain a formulation amenable to operator learning, the problem is expressed in terms of a functional representation of the evolving saturated region, given by a continuous filling factor $f : D \times [0,T] \to [0,1]$ that represents resin saturation and characterises the saturated domain via
\[
D(t) = \{\bfx \in D \mid f(\bfx,t) = 1\}.
\]
This representation provides a functional description of the moving geometry that is well suited for learning with neural operators. In this work, the filling factor is chosen to be consistent with the numerical filling factor produced by the control-volume finite element method (CVFEM) used to discretise and solve Eq.~\eqref{eq:fwd_inputs} on a stationary mesh \citep{Zenodo,Bruschke:1990ii,Advani,MichaelMinho}. Alternative functional representations of the moving domain, such as those arising from level-set or phase-field formulations, could equally be employed within the proposed operator-learning framework.

With the aid of the filling factor, the resin infusion problem is formulated as the evaluation of an operator $\mathcal{F}:\mathcal{U}\to\mathcal{Y}$, mapping admissible inputs $\bfu_{\mathrm{FM}}\in\mathcal{U}$ to admissible outputs $(p,f)\in\mathcal{Y}$, such that
\begin{align}
\mathcal{F}[\bfu_{\mathrm{FM}}]
:= 
\begin{bmatrix}
p[\bfu_{\mathrm{FM}}]\\[2pt]
f[\bfu_{\mathrm{FM}}]
\end{bmatrix}.
\label{eq:fwd_inputs2}
\end{align}
Here, $\mathcal{U}$ and $\mathcal{Y}$ denote spaces of physically admissible input parameters and corresponding solution fields, respectively. Owing to the limited availability of rigorous analytical results for this type of moving boundary problem with strongly heterogeneous coefficients, the precise functional characterisation of $\mathcal{U}$ and $\mathcal{Y}$ is left generic. Throughout this work, it is assumed that, for each admissible input $\bfu_{\mathrm{FM}}\in\mathcal{U}$, the forward operator $\mathcal{F}$ is well defined and admits a unique solution over the time interval of interest.

In Eq. \eqref{eq:fwd_inputs2} the explicit dependence on the inputs $\bfu_{\text{FM}}$ is included in $p$ and $f$, since these are determined by the set of inputs. For a given $\bfu_{\text{FM}}$, the pointwise evaluation at any $\bfx \in D$ and $t>0$ is denoted by
\begin{align}
    \FF[\bfu_{\text{FM}}](\bfx,t)\; := \left[\begin{array}{c}
    p[\bfu_{\text{FM}}](\bfx,t)\\
    f[\bfu_{\text{FM}}](\bfx,t)\end{array}\right].
    \label{eq:fwd_inputs3}
\end{align}
The mapping $\FF$ is referred to as the \emph{forward operator}, and 
its evaluation for a given set of inputs is defined as the solution of the forward 
problem, i.e.\ solving Eqs.~(\ref{eq: incompressibility}) to (\ref{eq:press}) 
with input $\bfu_{\text{FM}}$ from Eq. (\ref{eq:fwd_inputs}) which, again, involves the porosity and permeability functions defined at every point of the domain, $D$.

Conversely, the natural \emph{inverse problem} is to estimate $\bfu_{\text{FM}}$ given space- and time-discrete observations of $p[\bfu_{\text{FM}}]$ and $f[\bfu_{\text{FM}}]$ collected during the resin injection process (i.e., by using pressure measurements from sensors and/or flow-front tracking). The central objective is recovery of the reinforcement properties $K(\bfx)$ and $\phi(\bfx)$, since solving the inverse problem enables the identification and characterisation of potential material heterogeneities (i.e., local variations in these properties). However, in contrast to previous studies in which the viscosity $\mu$ and the inlet pressure parameters $(P_{I}, \lambda, \beta, \chi)$ are assumed to be known, since they can in principle be experimentally prescribed, these quantities are included in the inversion. This accounts for the uncertainty in these quantities and avoids the potential biases that may arise if they are fixed a priori. For instance, $\mu $ depends on the temperature of the resin which can vary from one experiment to another, as the lab conditions may vary. This can be captured by allowing $\mu$ to be an input parameter which is to be recovered in the inversion.

While, for simplicity, only the case of isotropic permeability is considered here, the approach can be generalised to the anisotropic case \citep{Advani,MichaelMinho}. Furthermore, in this work, $K(\bfx)$ and $\phi(\bfx)$ are treated as independent input parameters to reflect that there is no unique dependence of $K$ on $\phi$, but the framework can readily be extended to link these properties through a geometry factor that encodes information on fibre arrangement, orientation, and cross-sectional area \citep{Matveev}.

\subsection{Parametrisation and priors of permeability and porosity} \label{sec:parameterisation}

This work focuses on the two-dimensional case, where $D=[0,D_{x}]\times [0,D_{y}]$ is a rectangular domain. A linear resin inlet, $\partial D_I$, and a linear vent, $\partial D_0$, are positioned on opposite edges of $D$. The reinforcement is assumed to have a nominal permeability, $K_{\text{nom}}$, and porosity, $\phi_{\text{nom}}$.

Two types of possible defects are considered in the reinforcement during the resin infusion process: (1) race-tracking (RT), modelled by assigning a substantially higher permeability and porosity than the nominal values of the reinforcement to the affected regions (see similar treatments in \citep{Advani,Matveev,MC}), and (2) regions with lower permeability/porosity than the nominal values. RT may only occur in narrow regions along the boundaries of the domain parallel to the applied pressure gradient. These RT regions are parametrised by two one-dimensional random fields, $\xi_{T}(x)$ and $\xi_{B}(x)$, which specify the variable widths in the $x$-direction. Hence, the possible RT regions are
\begin{eqnarray}\label{eq:geo1}
\mathcal{RT}_{T}=&\big\{(x,y)\in D~~\vert ~~y>D_{y}-\xi_{T}(x)  \big\},\\
\mathcal{RT}_{B}=&\big\{(x,y)\in D~~\vert ~~y<\xi_{B}(x)  \big\}. \label{eq:geo2}
\end{eqnarray}
The log-permeabilities in the RT regions are denoted by $\log K_{T}(x,y)$ and $\log K_{B}(x,y)$. These are spatially varying functions, reflecting the severity of the RT defect within the reinforcement. Internal (i.e. non-RT) deviations in log-permeability are denoted by $\log K_{\text{def}}(x,y)$, which is again treated as a spatially varying function to account for the dependence on the severity of the defect.

\begin{figure*}
    \centering
    \includegraphics[width=0.9\linewidth]{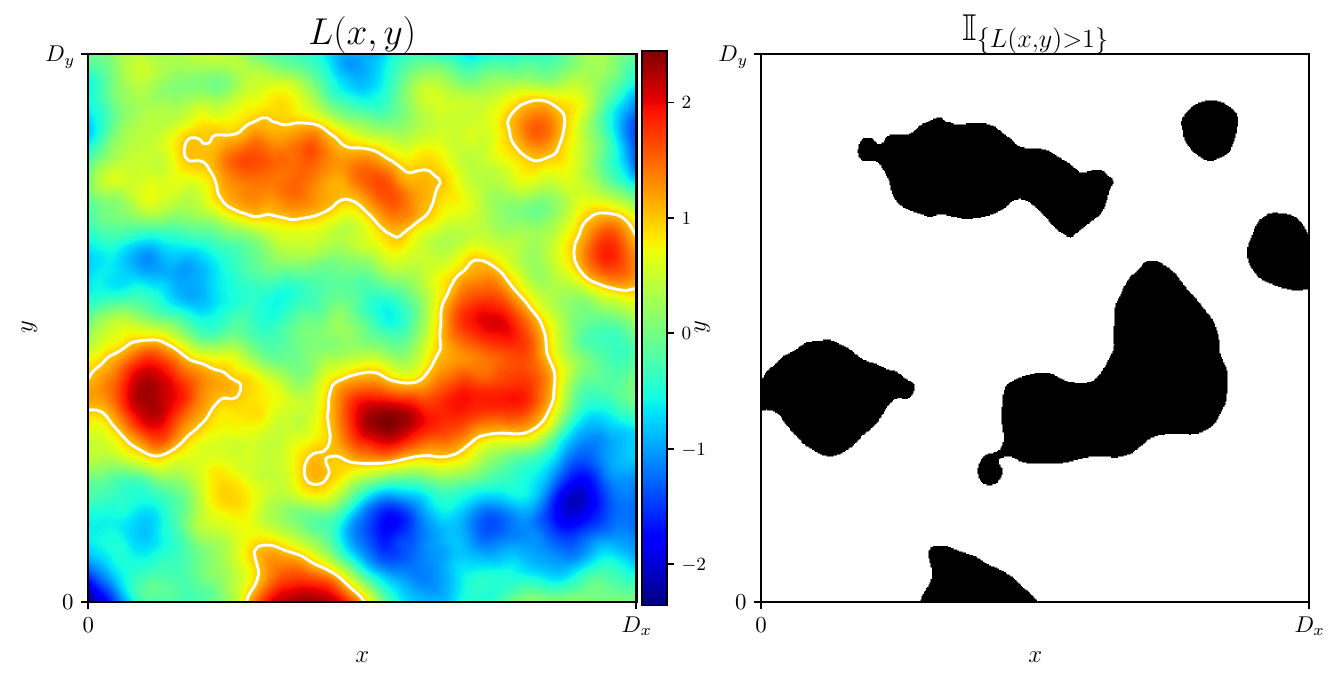}
    \caption{Left: realisation of a Gaussian random field used as the level-set function. Right: corresponding thresholded geometry with  $L_{*}=1$.}
    \label{fig:level_set}
\end{figure*}

\begin{figure*}
    \centering
    \includegraphics[width=0.8\linewidth]{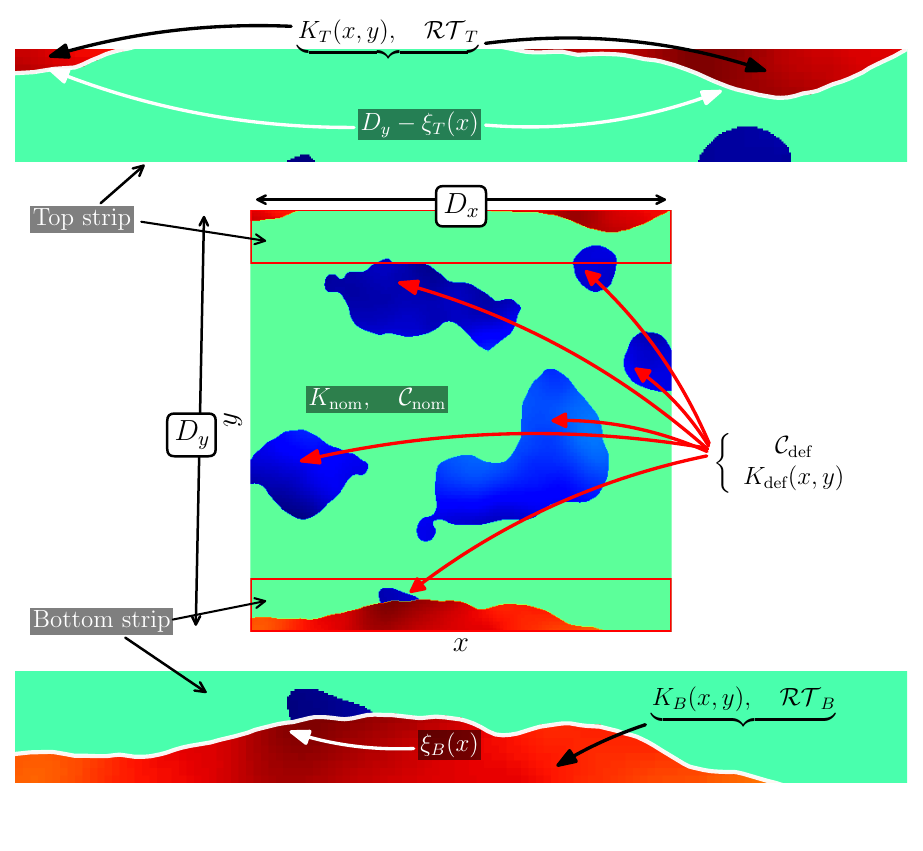}
    \caption{Configuration of the geometry: central defect region obtained from the level-set function together with the RT regions $\mathcal{RT}_{B}$ and $\mathcal{RT}_{T}$ defined by the random fields $\xi_{B}$ and $\xi_{T}$.}
    \label{fig:GeometryConfig}
\end{figure*}

To represent the random geometry of the central defects, the domain is parametrised using a two-dimensional random field $L(x,y)$, referred to as the level-set function, together with a given threshold value $L_{*}\in \mathbb{R}$. The central defect region is then defined as
\begin{eqnarray}\label{eq:C_def}
\mathcal{C}_{\text{def}}=\Big\{(x,y)\in D~~\vert ~~L(x,y)>L_{*}  \Big\}\setminus\big(\mathcal{RT}_{T}\cup \mathcal{RT}_{B}\big),
\end{eqnarray}
while the nominal (defect-free) region of the reinforcement is given by
\begin{eqnarray}\label{eq:geo3}
\mathcal{C}_{\text{nom}}=\Big\{(x,y)\in D~~\vert ~~L(x,y)<L_{*}  \Big\}\setminus\big(\mathcal{RT}_{T}\cup \mathcal{RT}_{B}\big).
\end{eqnarray}
To illustrate this parameterisation, Fig.~\ref{fig:level_set} displays a realisation of a 2D Gaussian random field (GRF) employed as the level-set function for this example (left) which is thresholded at $L^*=1$ to produce the corresponding central defect region (right). In Fig.~\ref{fig:GeometryConfig}, 1D GRFs $\xi_{B}$ and $\xi_{T}$ are used to define the RT regions, respectively. The resulting configuration shows both the central defect region and the RT regions $\mathcal{RT}_{B}$ and $\mathcal{RT}_{T}$.

Given the above configuration, the log-permeability field at any point $\mathbf{x}=(x,y)\in D$ is defined as
\begin{align}\label{eq:param1}
    \log K(\bfx)=&\log (K_{\text{nom}})\,\mathbb{I}_{\mathcal{C}_{\text{nom}}}(\bfx)+\log K_{\text{def}}(\bfx)\,\mathbb{I}_{\mathcal{C}_{\text{def}}}(\bfx)+\log K_{B}(\bfx)\,\mathbb{I}_{\mathcal{RT}_{B}}(\bfx)+\log K_{T}(\bfx)\,\mathbb{I}_{\mathcal{RT}_{T}}(\bfx),
\end{align}
where $\mathbb{I}_{A}(\bfx)$ denotes the indicator function of the set $A$, which is equal to $1$ if $\bfx \in A$ and $0$ otherwise. Therefore, each term in Eq.~\eqref{eq:param1} assigns the appropriate log-permeability depending on whether the point $\bfx$ lies in the nominal region, the central defect region, or one of the RT regions.

For the porosity, the representation:
\begin{align}\label{eq:param2}
\phi(\bfx) = \phi_{\text{nom}}\,\mathbb{I}_{\mathcal{C}_{\text{nom}}}(\bfx)
+ \phi_{\text{def}}\,\mathbb{I}_{\mathcal{C}_{\text{def}}}(\bfx)
+ \phi_{B}\,\mathbb{I}_{\mathcal{RT}_{B}}(\bfx)
+ \phi_{T}\,\mathbb{I}_{\mathcal{RT}_{T}}(\bfx),
\end{align}
is adopted. In contrast to the permeability, the porosity is assumed to be uniform within each region. This choice is motivated by the fact that, across the regions considered, the porosity does not exhibit large variability (unlike the permeability, which may vary by several orders of magnitude). Moreover, it was shown in previous work \citep{MC} that identifying small variations in porosity is particularly challenging and often not practically achievable.

Using Eqs. \eqref{eq:geo1} to \eqref{eq:geo3}, the log-permeability and porosity fields can be written in piecewise form as
\begin{eqnarray}\label{eq:param3}
    \log K(x,y)=&\begin{cases}
    \log K_{T}(x,y) & \text{if}~y>D_{y}-\xi_{T}(x)\\
    \log K_{B}(x,y) & \text{if}~y<\xi_{B}(x)\\
    \log K_{\text{nom}} &  \text{if}~L(x,y)\leq L_{*}~ \text{and}~~y\in [\xi_{B}(x),D_y-\xi_{T}(x)]\\ 
    \log K_{\text{def}}(x,y) &  \text{if}~L(x,y)> L_{*}~ \text{and}~~y\in [\xi_{B}(x),D_y-\xi_{T}(x)],
        \end{cases}\\        
        \phi(x,y)=&\begin{cases}
    \phi_{T} & \text{if}~y>D_{y}-\xi_{T}(x)\\
    \phi_{B} & \text{if}~y<\xi_{B}(x)\\
    \phi_{\text{nom}} &  \text{if}~L(x,y)\leq L_{*}~ \text{and}~~y\in [\xi_{B}(x),D_y-\xi_{T}(x)]\\ 
    \phi_{\text{def}} &  \text{if}~L(x,y)> L_{*}~ \text{and}~~y\in [\xi_{B}(x),D_y-\xi_{T}(x)].
        \end{cases}
\end{eqnarray}
This provides an explicit parametrisation, denoted by $\mathcal{P}_{K,\phi}$, which maps the collection of unknown quantities comprised in
\begin{eqnarray}\label{eq:param4}
\mathbf{u}_{K,\phi}:= \Big(\log K_{T},\log K_{B},\log K_{\text{def}},K_{\text{nom}},\phi_{T},\phi_{B},\phi_{\text{nom}},\phi_{\text{def}},L,\xi_{T},\xi_{B}\Big)  
\end{eqnarray}
to the physical permeability and porosity fields within the reinforcement:
\begin{eqnarray}\label{eq:param5}
(\log K,\phi)=\mathcal{P}_{K,\phi}[\bfu_{K,\phi}].
\end{eqnarray}

By specifying a \emph{prior} distribution over the inputs collected in $\mathbf{u}_{K,\phi}$, an initial characterisation of the reinforcement porosity and permeability is encoded that reflects prior knowledge before any measurements of the pressure field (and possibly the flow front position) are obtained. The parametrisation already accounts for the possible presence of RT regions as well as central defects, while the prior distribution prescribes the variability, smoothness, and size of these geometric regions. The prior also sets nominal baseline values and anticipated ranges for the defect regions. Throughout, it is assumed that the components of $\mathbf{u}_{K,\phi}$ are a priori independent. Full details of the prior specifications and the choice of hyperparameters are provided in~\ref{app: Priors}.

\begin{figure*}
    \centering
    \includegraphics[width=0.9\linewidth]{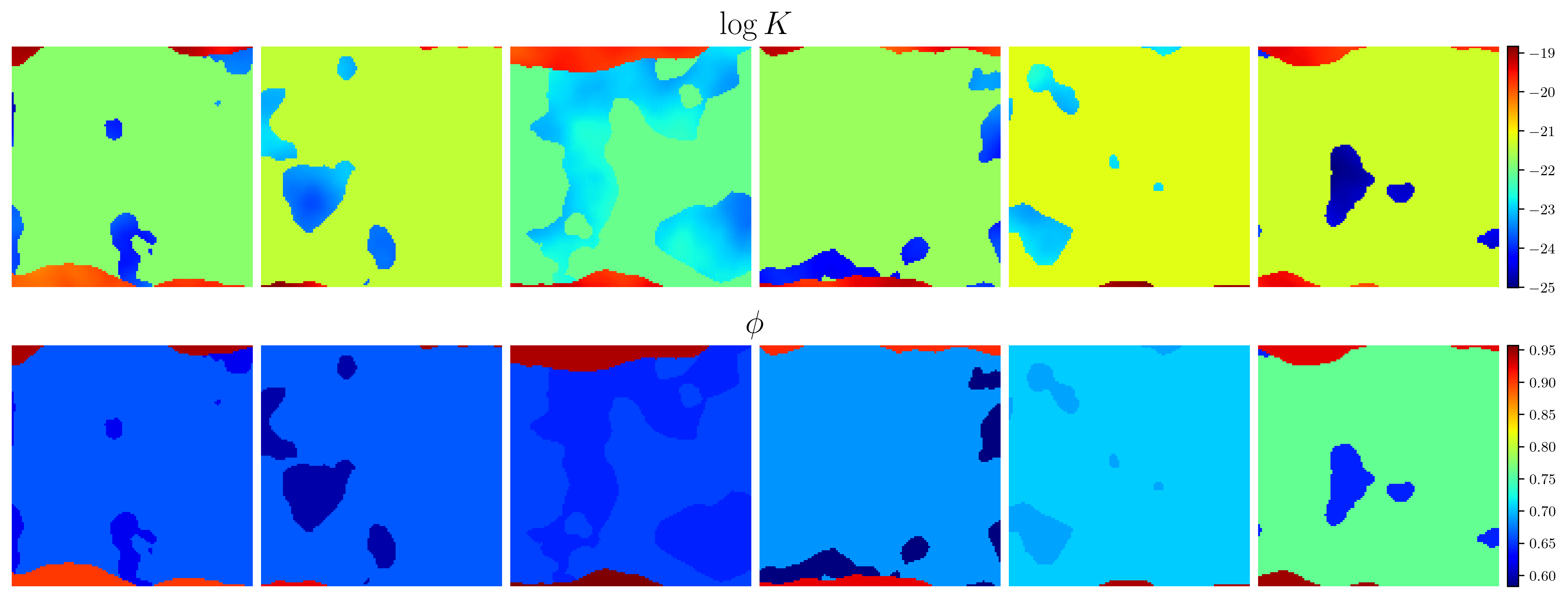}
    \caption{Five prior samples of log-permeability $\log K^{(j)}$ (top) and porosity $\phi^{(j)}$ (bottom) obtained via Eq.~\eqref{eq:param12}, given $\bfu_{K,\phi}^{(j)} \sim \mathbb{P}(\bfu_{K,\phi})$.}
    \label{fig:priorK_phi}.
\end{figure*}

Let $\mathbb{P}(\bfu_{K,\phi})$ denote the joint prior on $\bfu_{K,\phi}$, defined as the product of the individual priors. Samples from the prior, $\bfu_{K,\phi}^{(j)} \sim \mathbb{P}(\bfu_{K,\phi})$ induce samples from a push-forward prior on the physical fields $(\log K,\phi)$ via
\begin{equation}\label{eq:param12}
(\log K^{(j)},\,\phi^{(j)})=\mathcal{P}_{K,\phi}\big(\bfu_{K,\phi}^{(j)}\big),
\end{equation}
which can then be used as inputs for the forward model (CVFEM simulations of the resin injection). Fig.~\ref{fig:priorK_phi} shows random samples of permeability and porosity generated using Eq.~\eqref{eq:param12} under the priors specified in \ref{app: Priors} on a square domain with $D_x=D_y=0.3\,\mathrm{m}$, discretised on a $120\times120$ regular grid. Realisations of the 2D GRFs were obtained by standard sampling from a Cholesky factorisation of the covariance on the grid. For the 1D fields $\xi_T$ and $\xi_B$, a one-dimensional Mat\'ern covariance is constructed on the domain $[0,D_{x}]$, and it is discretised on a grid of 120 points that coincides with the corresponding boundaries. For the level-set threshold, $L_{*}=1$ is employed which, for the selection of $\sigma=1$ for the random fields of the level-set function, corresponds to a prior probability of approximately 16 \% of central defects. 

The samples shown in Fig.~\ref{fig:priorK_phi} demonstrate the range of spatial behaviours captured by the prior, with central defects of varying shape, size and intensity, arising naturally from the level-set construction. The width and permeability of RT channels also varies substantially. The level-set technique described here can also be used to model a wider variety of defect types. For instance, multiple level sets may be employed to represent more complex RT configurations, such as those interacting with the central region. Moreover, the framework readily accommodates high-permeability central defects, as well as the low-permeability defects considered here.




\subsection{The parametrised inverse problem} \label{sec:inverse_parameterised}
As $(\log K,\phi)$ is parametrised in terms of $\bfu_{K,\phi}$ defined in Eq.~\eqref{eq:param4}, it is further convenient to collect all unknowns into a single variable:
\begin{align}\label{eq:param6}
    \bfu = (\bfu_{K,\phi},\,\mu,\,P_{I},\,\lambda,\,\beta,\chi).
\end{align}
Then, the operator
\begin{align}\label{eq:param7}
\mathcal{P}[\bfu]:=(\mathcal{P}_{K,\phi}[\bfu_{K,\phi}],\mu,P_{I},\lambda,\beta,\chi)=(\log K,\,\phi,\,\mu,P_{I},\,\lambda,\,\beta,\,\chi)=\bfu_{\text{FM}}
\end{align}
is defined, which acts as $\mathcal{P}_{K,\phi}$ on $\bfu_{K,\phi}$ and as the identity on $(\mu,P_{I},\lambda,\beta,\chi)$, and thus it returns the parameters of the flow model, $\bfu_{\text{FM}}$, that were defined in Eq.~\eqref{eq:fwd_inputs}. This expresses the forward operator given in Eq.~\eqref{eq:fwd_inputs2} as the composition
\begin{align}\label{eq:param8}
     \FF \circ \mathcal{P}(\bfu)
    \;=\;   
    \FF[\bfu_{FM}]
    \;=\; \left[\begin{array}{c}
    p[\bfu_{\text{FM}}]\\
    f[\bfu_{\text{FM}}]\end{array}\right].
\end{align}

\begin{figure}[h]
    \centering
    \includegraphics[width=1.0\linewidth]{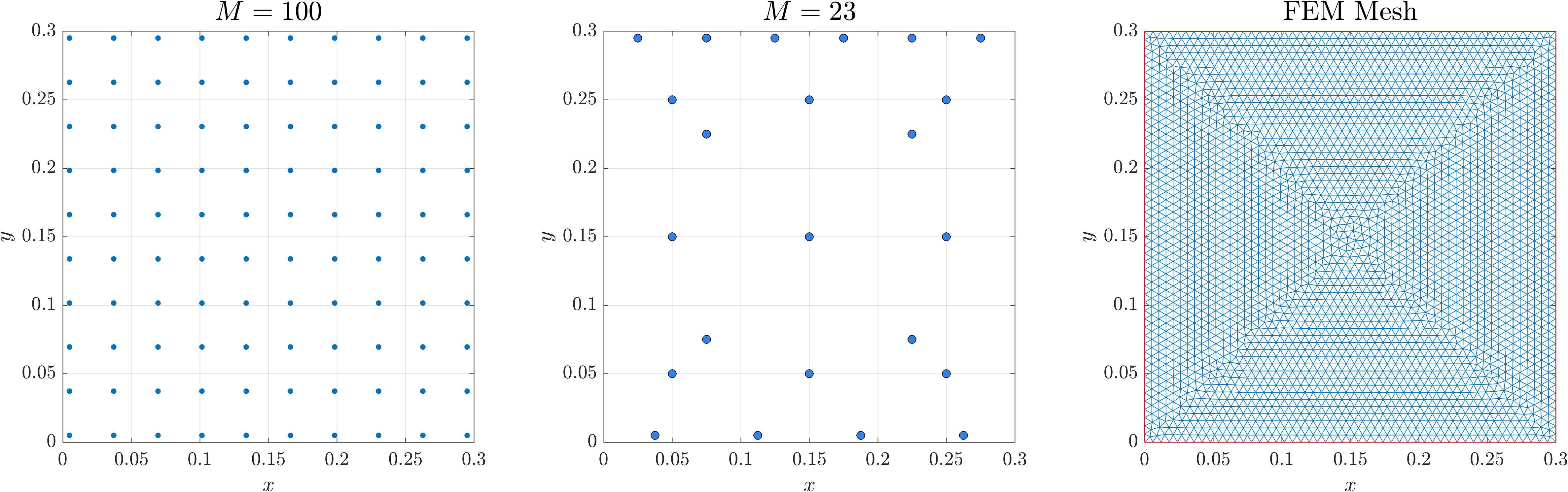}
 \caption{Left: locations of pressure sensors for a dense configuration with $M=100$. 
Middle: locations of pressure sensors for a sparse configuration with $M=23$, corresponding to the experimental setup. 
Right: FEM mesh used for the CVFEM simulations.}
    \label{fig:Meshes}
\end{figure}

To infer $\bfu$, experimental measurements of $(p,f)$ obtained only from pressure sensors are used, although the framework can be extended to incorporate flow-front data \citep{Matveev,Iglesias_2018}. The measurement process is described by an \emph{observation operator}, $\OO:\mathcal{Y}\to \mathbb{R}^{MN}$, acting on the output of the forward operator:
\begin{align}\label{eq:param9}
    \OO\left[\begin{array}{c}
    p[\bfu_{\text{FM}}]\\
    f[\bfu_{\text{FM}}]\end{array}\right] : = \Bigg\{\Big\{p[\bfu_{\text{FM}}](\bfx_m,t_1)\Big\}_{m=1}^{M},\ldots,\Big\{p[\bfu_{\text{FM}}](\bfx_m,t_N)\Big\}_{m=1}^{M}\Bigg\},
\end{align}
where $M$ denotes the number of pressure sensors located at $\{\bfx_m\}_{m=1}^M \subset D$, and $N$ is the number of observation times $\{t_n\}_{n=1}^N$ within the time horizon $[0,T]$ which is chosen based on forward simulations for samples from the prior to ensure that the reinforcement is impregnated with high probability. Thus, the measurement process corresponds to point-wise measurements of the pressure at a finite set of sensor locations and times. In this work, a collection of 34 observation times is considered within horizon $T=110\text{s}$. For the sensors, two configurations are studied: one with $M=100$ sensors uniformly distributed across the domain, and a second one with $M=23$ sensors, shown in Fig.~\ref{fig:Meshes}. The first configuration is used primarily for benchmarking the inverse method, while the second corresponds to the experimental setting described in Section~\ref{sec: Lab}.

Combining Eqs.~\eqref{eq:param8} and \eqref{eq:param9},  the \textit{parameter-to-measurements map},
\begin{align}
\GG[\bfu]:=    (\OO\circ\FF\circ \mathcal{P})\,[\bfu]= \Bigg\{\Big\{p[\bfu_{\text{FM}}](\bfx_m,t_1)\Big\}_{m=1}^{M},\ldots,\Big\{p[\bfu_{\text{FM}}](\bfx_m,t_N)\Big\}_{m=1}^{M}\Bigg\},\label{eq:param10}
\end{align}
is obtained, which will be used for the inversion.  More specifically, given experimental measurements of the right-hand side of Eq. \eqref{eq:param10}, denoted by $\mathbf{d}^{\dagger}$, the aim of the inverse problem is to recover $\bfu$. To address this inverse problem, the Bayesian approach discussed in the following section is employed.  

\subsection{The Bayesian approach} \label{subsec:BayesApproach}

A Bayesian framework is employed to identify $\bfu$ for given measurement data $\mathbf{d}^{\dagger}$, where both quantities are treated as random variables. Following standard assumptions \citep{Stuart,Iglesias_2018,Matveev,MarcoYang}, the measurements $\mathbf{d}^{\dagger}$ are regarded as a realisation of
\begin{align}
\mathbf{d} = \mathcal{G}[\bfu] + \eta, \label{eq: inverse_problem_unparameteriseddublicate}
\end{align}
where $\mathcal{G}$ denotes the parameter-to-measurements map from Eq. \eqref{eq:param10} and $\eta$ represents measurement noise. The noise term accounts for sensor inaccuracies and is assumed to follow a Gaussian distribution with zero mean and known diagonal covariance $\Gamma$, i.e. $\eta \sim \mathcal{N}(0,\Gamma)$. Extensions to correlated noise structures (non-diagonal $\Gamma$) can also be incorporated if such information is available.

Prior to collecting measurements, a \textit{prior} probability measure, denoted by $\mathbb{P}(\bfu)$, is placed on $\bfu$ to encode existing knowledge of the unknown inputs. For simplicity, it is assumed that the components of $\bfu$ in Eq.~\eqref{eq:param6} are independent under the prior, so that 
\begin{align}
\mathbb{P}(\bfu)= \mathbb{P}(\bfu_{K,\phi})\mathbb{P}(\mu)\mathbb{P}(P_{I})\mathbb{P}(\lambda)\mathbb{P}(\beta)\mathbb{P}(\chi),\label{eq:prior_u}
\end{align}
where, as before, $\mathbb{P}(\bfu_{K,\phi})$ is the prior on the parameters of the reinforcement log-permeability and porosity detailed in Section \ref{sec:parameterisation} (see also \ref{app: Priors}). The priors for $\mu,P_{I},\lambda,\beta,\chi$ are chosen as uniform with the values displayed in Table \ref{tab:prior_mu_inlet}.

\begin{table}
    \centering
    \caption{Range for priors on viscosity and inlet pressure.}
    \begin{tabular}{c|c}
         Parameter & Range\\
         \hline
         $\mu$ & $[0.085,0.12]$ Pa$\cdot$s\\
         $P_I$ & $[92, 120]$ kPa\\
         $\lambda$ & $[0.6, 1.25]$\\
         $\beta$ & $[0.2, 0.7]$\\
         $\chi$ & $[0.35, 0.75]$ 
    \end{tabular}
    \label{tab:prior_mu_inlet}
\end{table}

The solution of the Bayesian inverse problem is the \textit{posterior}, which represents the updated knowledge of $\bfu$ after assimilating the measurements. By Bayes’ theorem \citep{Stuart} and under the Gaussian noise assumption, the posterior density is given by
\begin{align}
\mathbb{P}(\bfu|\mathbf{d}^{\dagger}) \;=\; \frac{1}{Z}\,\mathbb{P}(\bfu)\exp\Big(-\tfrac{1}{2}\big\lVert \Gamma^{-1/2}\big(\mathbf{d}^{\dagger}- \mathcal{G}[\bfu]\big)\big\rVert^2 \Big), \label{eq:pos}
\end{align}
where $Z$ is the normalising constant ensuring that the posterior integrates to one.

The formulation above is understood in a formal Bayesian sense. Rigorous existence and well-posedness results for the posterior typically rely on continuity properties of the parameter-to-measurements map $\mathcal{G}$ with respect to the unknown parameters. For the moving boundary problem under consideration, such results have been established only for the 1D problem and under additional regularity assumptions on the permeability field \citep{Iglesias_2018}. In the present work, which considers moving boundary problems with strongly heterogeneous and possibly discontinuous material properties, a general theoretical analysis of posterior well-posedness is not available. Nevertheless, following common practice in applied Bayesian inversion, it is assumed that for admissible inputs the forward model defines a measurable map and that the posterior is well defined.

\subsection{Approximation of the posterior}\label{subsec:approx}
The posterior given by Eq. \eqref{eq:pos} is fully specified up to the normalising constant. However, since the parameter-to-measurements map $\GG(\bfu)$ is nonlinear, no explicit analytical expression for $Z$ is available. Moreover, direct numerical integration is infeasible in high-dimensional settings. For example, assuming a $120\times 120$ discretisation of $(L, \log K_{T}, \log K_{B}, \log K_{\mathrm{def}})$, a 120-point discretisation of $(\xi_{T}, \xi_{B})$, together with the scalar parameters $(K_{\mathrm{nom}}, \phi_{T}, \phi_{B}, \phi_{\mathrm{nom}}, \phi_{\mathrm{def}}, \mu, P_{I}, \lambda, \beta, \chi)$ yields a total dimension of $4\times 120^2 + 2\times 120 + 10 = 57{,}847.$

Consequently, the posterior distribution must be explored using sampling-based approaches such as Markov chain Monte Carlo (MCMC) \citep{Liu} or Sequential Monte Carlo (SMC) \citep{kantas2014smchighdim}, both of which avoid the explicit computation of $Z$. As stated in Section \ref{sec: Intro}, sampling approaches for such high-dimensional problems often require millions of forward simulations and are therefore not conducive to real-time estimation. 



In this work, ensemble Kalman inversion (EKI) is adopted as the inversion method \citep{MarcoYang}. This choice is primarily motivated by practical considerations arising from the structure of the forward model and the objectives of the study. The forward problem involves a moving boundary formulation for which linearised solvers and adjoint models are not readily available. Moreover, the parametrisation of the permeability and porosity fields, which incorporates level-set representations and piecewise-defined random fields to capture defects and race-tracking regions, is not differentiable. Although smooth approximations, such as regularised Heaviside functions, could be introduced, doing so would substantially increase the complexity of the formulation without being essential to the aims of this work. 

It is emphasised, however, that EKI yields an approximate characterisation of the Bayesian posterior and does not, in general, produce exact posterior samples. More accurate Bayesian inference could, in principle, be obtained using advanced methods, including geometric or likelihood-informed MCMC techniques \citep{beskos2016geometric,cui2016dimensionindependent,lan2019adaptive} and variational inference frameworks \citep{povala2022variational,frank2022geometricVI}. These approaches typically rely on gradient information of the parameter-to-observable map and therefore require access to linearised forward models or adjoint solvers, which are not available in the present setting. Consequently, a derivative-free ensemble-based inversion method, requiring only forward model evaluations, is adopted. The primary focus of this work is on demonstrating the effectiveness of DeepONet surrogates for accelerating inverse problems governed by moving boundary flows, rather than on the development or comparison of inversion algorithms.


The employed EKI algorithm is outlined in \ref{app: EKI}, and further methodological details are discussed in previous work by the authors \citep{MarcoYang,Iglesias_2018}. Briefly, the method starts with an ensemble of $J$ samples $\{\bfu^{(j)}\}_{j=1}^J$ drawn from the prior $\mathbb{P}(\bfu)$, which are iteratively updated to gradually improve approximation of the experimental data.
Upon convergence, the ensemble $\{\bfu^{(j)}\}_{j=1}^J$ provides an approximation of the posterior. The transition between iterations is controlled by a regularisation parameter which is selected adaptively.  

Although the optimal choice of ensemble size $J$ remains an open question, in practice only a few hundred particles are sufficient for EKI to produce posterior estimates (and hence posterior means and variances) of the reinforcement permeability with accuracy comparable to SMC, but at a computational cost reduced by approximately two orders of magnitude \citep{Iglesias_2018}.

Once the posterior ensemble $\{\bfu^{(j)}\}_{j=1}^J$ is obtained, it is mapped to the physical parameters via Eq. \eqref{eq:param7}, i.e.  
\[
\bfu_{\text{FM}}^{(j)} = \big(\log K^{(j)},\, \phi^{(j)},\, \mu^{(j)},\, P_{I}^{(j)},\, \lambda^{(j)},\, \beta^{(j)},\, \chi^{(j)}\big) = \mathcal{P}[\bfu^{(j)}],
\]  
from which the ensemble mean,
\begin{align}\label{eq:pos_mean1}
\overline{\log K}(\bfx) &= \frac{1}{J}\sum_{j=1}^{J} \log K^{(j)}(\bfx), \qquad     
\overline{\phi}(\bfx) = \frac{1}{J}\sum_{j=1}^{J} \phi^{(j)}(\bfx),
\end{align}
is computed, as well as the pointwise standard deviation:
\begin{align}\label{eq:pos_std1}
\text{STD}(\log K)(\bfx) &= \Bigg[\frac{1}{J-1}\sum_{j=1}^{J} \Big(\log K^{(j)}(\bfx)-\overline{\log K}(\bfx)\Big)^2\Bigg]^{1/2}, \\
\text{STD}(\phi)(\bfx) &= \Bigg[\frac{1}{J-1}\sum_{j=1}^{J} \Big(\phi^{(j)}(\bfx)-\overline{\phi}(\bfx)\Big)^2\Bigg]^{1/2}.\label{eq:pos_std2}
\end{align}

Beyond estimating permeability and porosity, inferring the geometry of defective regions under the posterior is also of interest. To this end, the geometric parametrisation introduced in Eqs.~\eqref{eq:geo1} to \eqref{eq:geo3} is leveraged. Each ensemble member can be written as  
\[
\bfu^{(j)}=\big(\bfu_{K,\phi}^{(j)},\,\mu^{(j)},\,P_{I}^{(j)},\,\lambda^{(j)},\,\beta^{(j)}, \chi^{(j)}\big),
\]  
with  
\[
\mathbf{u}_{K,\phi}^{(j)} = \Big(\log K_{T}^{(j)},\,\log K_{B}^{(j)},\,\log K_{\text{def}}^{(j)},\,K_{\text{nom}}^{(j)},\,\phi_{T}^{(j)},\,\phi_{B}^{(j)},\,\phi_{\text{nom}}^{(j)},\,\phi_{\text{def}}^{(j)},\,L^{(j)},\,\xi_{T}^{(j)},\,\xi_{B}^{(j)}\Big).
\]

For each member of the posterior ensemble, $(\xi_{T}^{(j)}, \xi_{B}^{(j)})$ may be used to construct the RT regions
\begin{align}
\mathcal{RT}_{T}^{(j)} &= \big\{(x,y)\in D \,\big|\, y > D_{y}-\xi_{T}^{(j)}(x)\big\}, \\
\mathcal{RT}_{B}^{(j)} &= \big\{(x,y)\in D \,\big|\, y < \xi_{B}^{(j)}(x)\big\},
\end{align}
while from the ensemble of level-set functions $L^{(j)}$ the central defect regions are computed:
\begin{align}
\mathcal{C}_{\mathrm{def}}^{(j)} = \big\{(x,y)\in D \,\big|\, L^{(j)}(x,y) > L_{*}\big\}\setminus\big(\mathcal{RT}_{T}^{(j)}\cup \mathcal{RT}_{B}^{(j)}\big).
\end{align}

The pointwise posterior probabilities of central defects of RT (top or bottom) are respectively approximated as \citep{MarcoYang,Matveev} 
\begin{align}\label{eq:pos_probdef}
 \mathbb{P}_{\text{def}}(\bfx) := \frac{1}{J}\sum_{j=1}^{J}  \mathbb{I}_{\mathcal{C}_{\mathrm{def}}^{(j)}}(\bfx) \text{ and }
  \mathbb{P}_{\text{RT}}(\bfx) := \frac{1}{J}\sum_{j=1}^{J}  \mathbb{I}_{\mathcal{RT}_{T}^{(j)}\cup \mathcal{RT}_{B}^{(j)}}(\bfx).
\end{align}


In addition, the posterior ensemble allows to extract and visualise marginal distributions of the scalar parameters, e.g.\ $K_{\text{nom}}$, $\phi_{T}$, $\phi_{B}$, $\phi_{\text{nom}}$, $\phi_{\text{def}}$, the viscosity $\mu$, and the inlet-pressure parameters $(P_{I},\,\lambda,\,\beta,\,\chi)$.  

Finally, posterior uncertainty in the parameter space can be propagated to the measurement space by evaluating the forward operator on the posterior ensemble:
\begin{align}
\mathbf{d}^{(j)} = \mathcal{G}[\bfu^{(j)}] + \eta^{(j)}, \qquad \eta^{(j)}\sim N(0,\Sigma), \label{eq: inverse_problem_unparameterised}
\end{align}
for $j=1,\ldots,J$, thereby yielding predictive distributions for the sensor data.

\section{Full model Benchmark}\label{sec:virtual_EKI}
For benchmarking purposes, EKI is first applied to the \emph{full} forward model, and virtual experiments are considered  with synthetic data being generated from a known ground–truth reinforcement under prescribed injection conditions.

\subsection{Ground truth and synthetic data}
For a reinforcement with nominal permeability  $K_{\mathrm{nom}}^{\dagger}=4\times10^{-10}\,\mathrm{m}^2$ and porosity $\phi_{\mathrm{nom}}^{\dagger}=0.73$, which are typical values for random glass fibre mats, the true log-permeability $\log K^{\dagger}$ and porosity $\phi^{\dagger}$ are shown in the left panels of Fig.~\ref{fig:full_EKI1}. These are:
(i) a circular inclusion with permeability $1.2\times10^{-10}\,\mathrm{m}^2$ and porosity $0.62$, and
(ii) a rectangular inclusion with permeability $4\times10^{-11}\,\mathrm{m}^2$ and porosity $0.62$. 
There are RT regions ($\mathbf{A}$, $\mathbf{B}$, $\mathbf{C}$; see Fig.~\ref{fig:full_EKI1}) of elevated permeability along the boundaries parallel to the direction of the applied pressure gradient:
$\mathbf{A}$ has width $7.5\,\mathrm{mm}$ and permeability $2.5\times10^{-9}\,\mathrm{m}^2$;
$\mathbf{B}$ has width $15\,\mathrm{mm}$ and permeability $4.0\times10^{-9}\,\mathrm{m}^2$;
$\mathbf{C}$ has width $7.5\,\mathrm{mm}$ and permeability $4.0\times10^{-9}\,\mathrm{m}^2$.
For all RT regions, the porosity values are $\phi_T^{\dagger}=\phi_B^{\dagger}=0.91$.

Although the geometry is intentionally simple and not representative of real preforms, it was chosen deliberately because the resulting porosity and permeability fields lie outside the range of typical draws from the prior in Section~\ref{sec:parameterisation}. The prior, which combines random geometries with Mat\'ern GRFs, inherently favours smooth spatial variations and therefore cannot fully capture features with sharp corners and long straight edges. Using this out-of-prior (“out-of-sample”) truth provides a valuable stress test for the inversion, enabling the assessment of the framework's robustness and generalisation under covariate shift.

The true resin viscosity is $\mu^{\dagger}=0.092\,\mathrm{Pa}\cdot\mathrm{s}$. The true inlet pressure model is given by Eq. \eqref{eq:press} with the following selection of parameters: 
\[
(P_I^{\dagger},\,\lambda^{\dagger},\,\beta^{\dagger},\,\chi^{\dagger})
= (109.12\,\mathrm{kPa},\,1.114,\,0.42,\,0.66).
\]
The full set of ground–truth inputs is 
\[
\bfu_{\mathrm{FM}}^{\dagger}
=\big(\log K^{\dagger},\,\phi^{\dagger},\,\mu^{\dagger},\,P_I^{\dagger},\,\lambda^{\dagger},\,\beta^{\dagger},\,\chi^{\dagger}\big).
\]

The inputs $\log K^{\dagger}$ and $\phi^{\dagger}$ are first defined on a $120\times120$ regular grid and are subsequently interpolated onto the unstructured mesh of 2{,}973 nodes shown in Fig.~\ref{fig:Meshes} using nearest-neighbour interpolation, which preserves sharp material interfaces arising from RT regions and defects. 

The same mesh is then employed in the CVFEM solver from \citep{Zenodo} to generate the reference fields, $\FF[\bfu_{\text{FM}}^{\dagger}] = \big(p[\bfu_{\text{FM}}^{\dagger}],\, f[\bfu_{\text{FM}}^{\dagger}]\big)$. Virtual measurements are obtained by applying the observation operator $\mathcal{O}$ (cf.\ Eq.~\eqref{eq:param9}) to the reference solution under two sensor layouts with $M=100$ and $M=23$ sensors, respectively. Computationally, this corresponds to extracting the pressure values at the mesh nodes nearest to the prescribed sensor locations, while the observation times are chosen as a set of $N=34$ time steps selected within the horizon $[0,110]\,\mathrm{s}$ in order to capture the main stages of the resin front progression during the infusion. Synthetic data, $\mathbf{d}^{\dagger}$, are then generated by adding Gaussian noise to these simulation-based pressures, i.e.
\[
\mathbf{d}^{\dagger} \;=\; \Big\{p[\bfu_{\text{FM}}^{\dagger}](\bfx_{m},t_{1}),\dots, p[\bfu_{\text{FM}}^{\dagger}](\bfx_{m},t_{N})\Big\}_{m=1}^{M}+ \eta^{\dagger}, 
\]
where $\eta^{\dagger}\in \RR^{MN}$ is a realisation of centred Gaussian noise with diagonal covariance
\begin{equation}\label{eq:noise}
\Gamma 
= \text{diag}\Big(\big[\sigma_0\,\max\big\{\GG(\bfu_{\text{FM}}^\dagger),\,100\,\mathrm{Pa}\big\}\big]^2\Big).
\end{equation}
The variance is set to $\sigma_0=0.025$, corresponding to adding noise at a $2.5\%$ level, relative to noise-free measurements, with a minimum variance floor equivalent to $100\,\mathrm{Pa}$ to prevent unrealistically small uncertainties for weak signals. This choice is consistent with the specified precision of the pressure sensors later employed in the laboratory experiments (Section~\ref{sec: Lab}).

\subsection{Inversion setup and outputs}

An all-at-once inversion strategy is adopted, whereby measurements from all observation times are assimilated simultaneously. The framework could also be adapted to sequential inversion \citep{Iglesias_2018,MC}. The EKI algorithm is applied as described in Section~\ref{subsec:BayesApproach}. Although the framework is formulated to infer $\bfu$, the ultimate goal is to infer the physical parameters $\bfu_{\text{FM}}$. As discussed earlier, once the posterior of $\bfu$ is obtained, the posterior of $\bfu_{\text{FM}}$ can be recovered through the parametrisation $\bfu_{\text{FM}}=\mathcal{P}[\bfu]$. In practice, once EKI has converged, the posterior ensemble $\{\bfu^{(j)}\}_{j=1}^J$ is mapped into physical space via $\mathcal{P}$ to produce $\{\bfu_{\mathrm{FM}}^{(j)}\}_{j=1}^J$. From this ensemble, posterior means and standard deviations of the permeability and porosity fields are computed, as well as posterior probabilities of geometric defects. These include central defects and RT regions, which constitute key diagnostic outputs of the proposed inversion framework.

To benchmark performance and assess the effect of ensemble size, results were obtained for three ensemble sizes, $J\in\{500,\,1000,\,5000\}$. All computations in this section were run on the University of Nottingham HPC system using a MATLAB implementation of the CVFEM forward solver (see \citep{Zenodo}) coupled with EKI. At each EKI iteration, the forward evaluations (one per ensemble member) were parallelised and distributed across 90 CPU cores on a single node with 1536\,GB of RAM. The number of EKI iterations required for convergence, the wall–clock time, and the number of parameter-to-measurement evaluations are reported in Table~\ref{tab:full_EKI}.

\begin{figure*}
    \centering
    \includegraphics[width=0.9\linewidth,clip,trim=0 2cm 0 0]{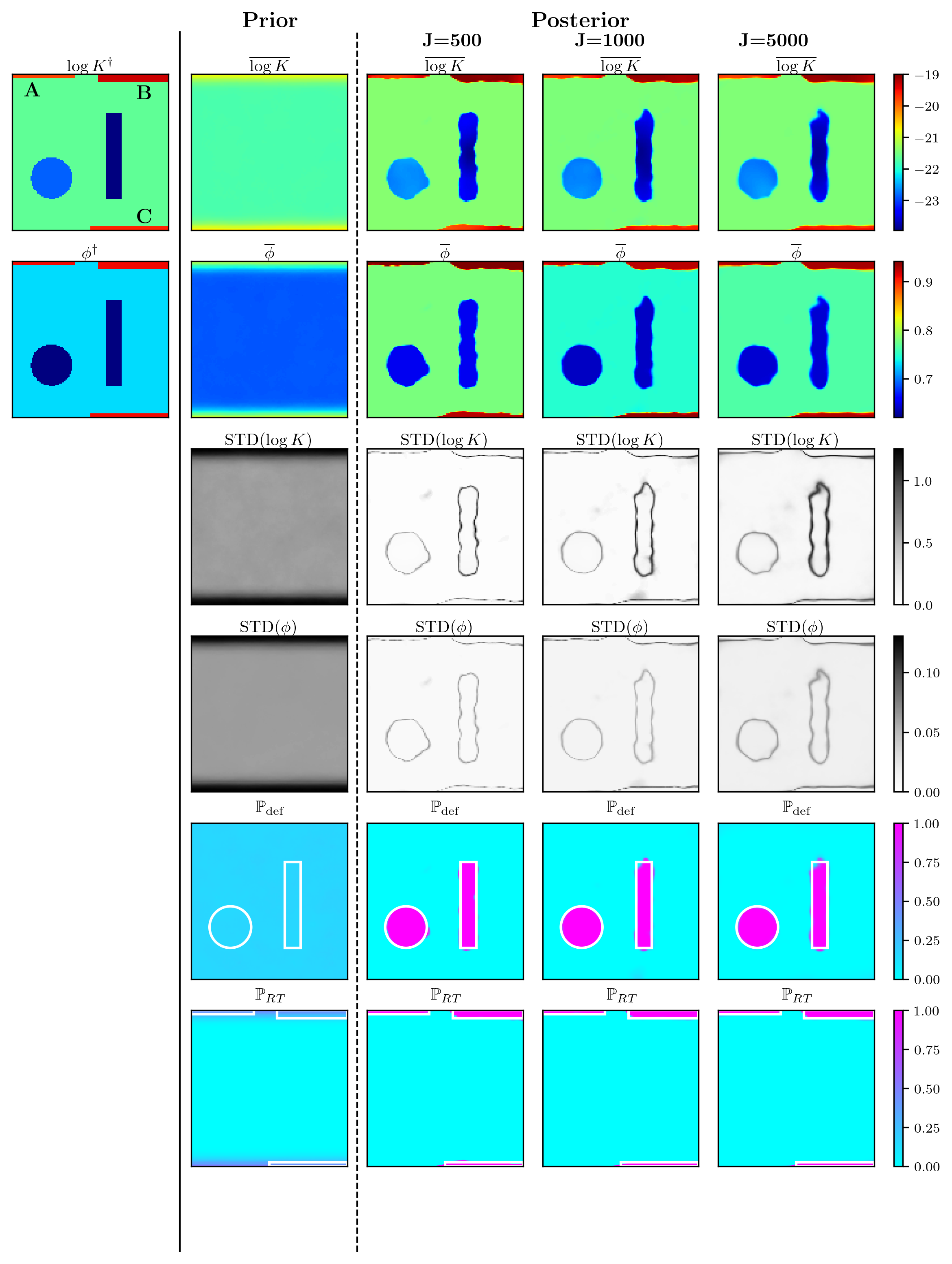}
    \caption{Results for the $M=100$ sensor configuration. First column: ground–truth log-permeability, $\log K^{\dagger}$, and porosity, $\phi^{\dagger}$. Second column: prior statistics. Third to fifth columns: posterior statistics obtained with ensemble sizes $J=500$, $J=1000$, and $J=5000$, respectively. From top to bottom: mean of log-permeability, $\overline{\log K}$, mean of porosity, $\overline{\phi}$, standard deviation of log-permeability, $\mathrm{STD}(\log K)$, standard deviation of porosity, $\mathrm{STD}(\phi)$, probability of central defects, $\mathbb{P}_{\mathrm{def}}$, and probability of RT, $\mathbb{P}_{\mathrm{RT}}$. The geometry of the true central defects is shown in white in the fifth row, while the true RT geometry is shown in the bottom row.}    \label{fig:full_EKI1}
\end{figure*}

\begin{figure*}
    \centering
    \includegraphics[width=0.9\linewidth,clip,trim=0 2cm 0 0]{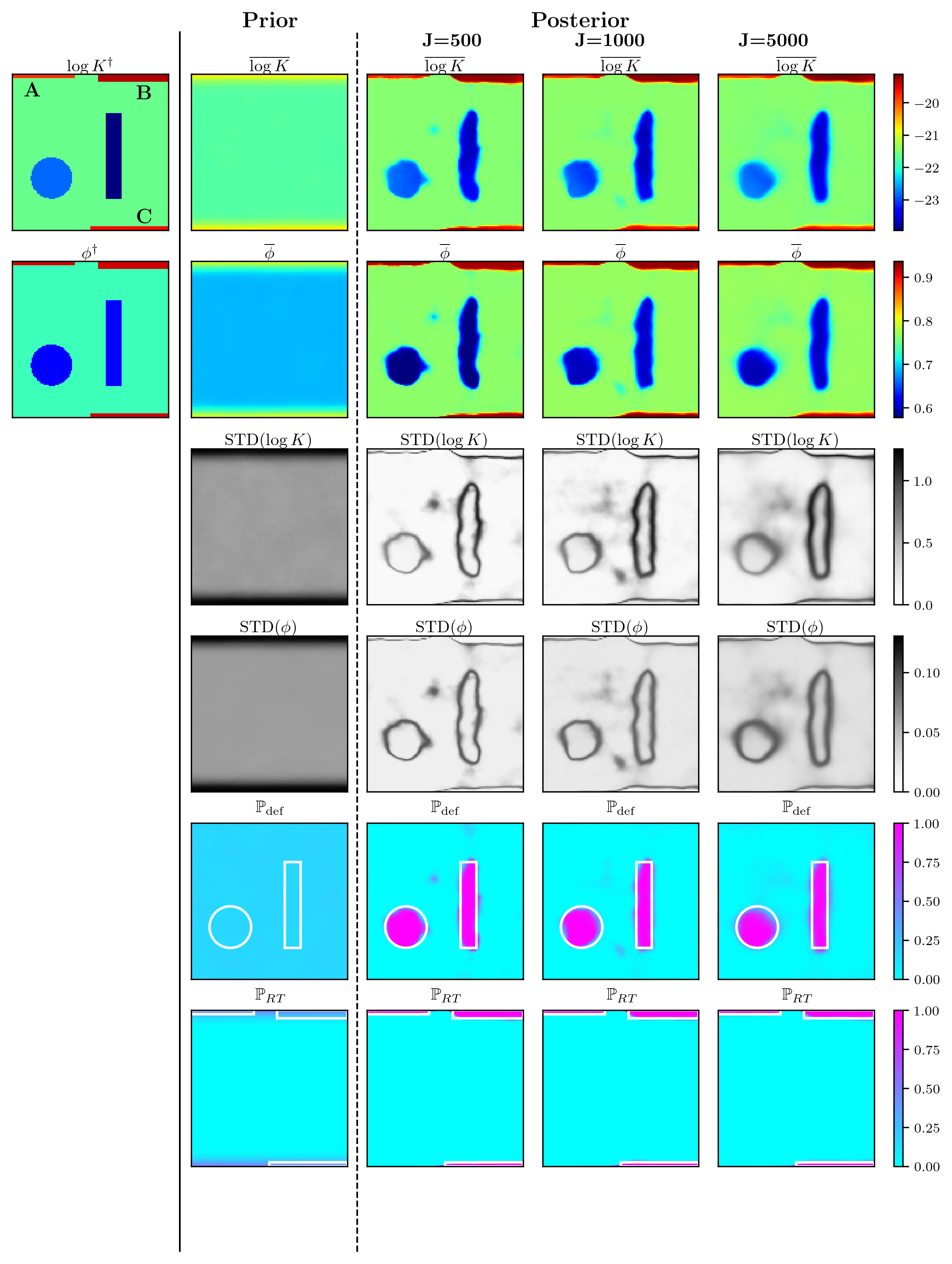}
     \caption{Results for the $M=23$ sensor configuration. See detailed description in Fig.~\ref{fig:full_EKI1}.}   
    \label{fig:full_EKI2}
\end{figure*}

Posterior means and standard deviations of log–permeability and porosity, as well as posterior probabilities of central defects and RT, are shown in Fig.~\ref{fig:full_EKI1} for the dense sensor configuration ($M=100$) and in Fig.~\ref{fig:full_EKI2} for the sparse configuration ($M=23$). As a reference, prior statistics are shown which were computed from $J=5000$ prior samples generated as described in Section~\ref{sec:parameterisation}. As expected, the dense sensor layout ($M=100$) yields reconstructions in closest agreement with the ground–truth porosity and log–permeability. The posterior probabilities of the central defect and the RT regions align closely with their true geometries. For the more realistic sparse layout ($M=23$), accuracy degrades, but the geometry of both the RT paths and the central defect remain reasonably well captured, suggesting that the framework is robust under practical sensing constraints.  

Regarding ensemble size, for $M=100$ the posterior statistics in Fig.~\ref{fig:full_EKI1} are relatively insensitive to $J$; even $J=500$ performs satisfactorily. By contrast, for $M=23$ (Fig.~\ref{fig:full_EKI2}), performance improves more noticeably with larger ensembles, with $J=5000$ yielding the most accurate reconstructions. This suggests that under sparse sensing, larger ensembles are preferable. Nevertheless, the wall–clock times in Table~\ref{tab:full_EKI} highlight the high computational cost of full–model EKI, even with significant parallelisation, underscoring the need for surrogate–accelerated inversion.

\begin{figure*}
    \centering
    \includegraphics[width=0.9\linewidth]{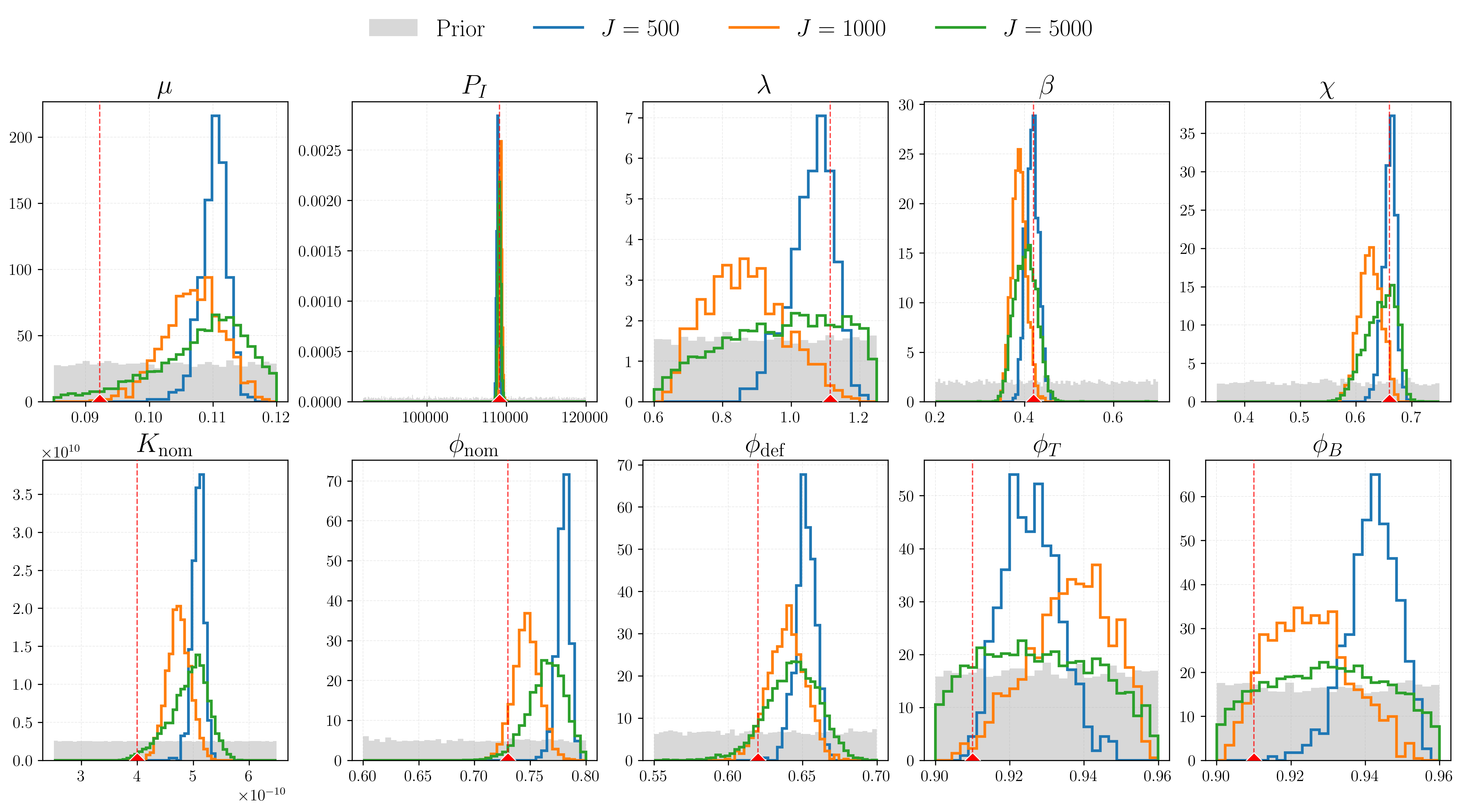}
    \caption{Histogram of priors and posteriors of scalar parameters for the $M=100$ sensor configuration, along with the true values (red).}
    \label{fig:full_EKI2B}
\end{figure*}

\begin{figure*}
    \centering
    \includegraphics[width=0.9\linewidth]{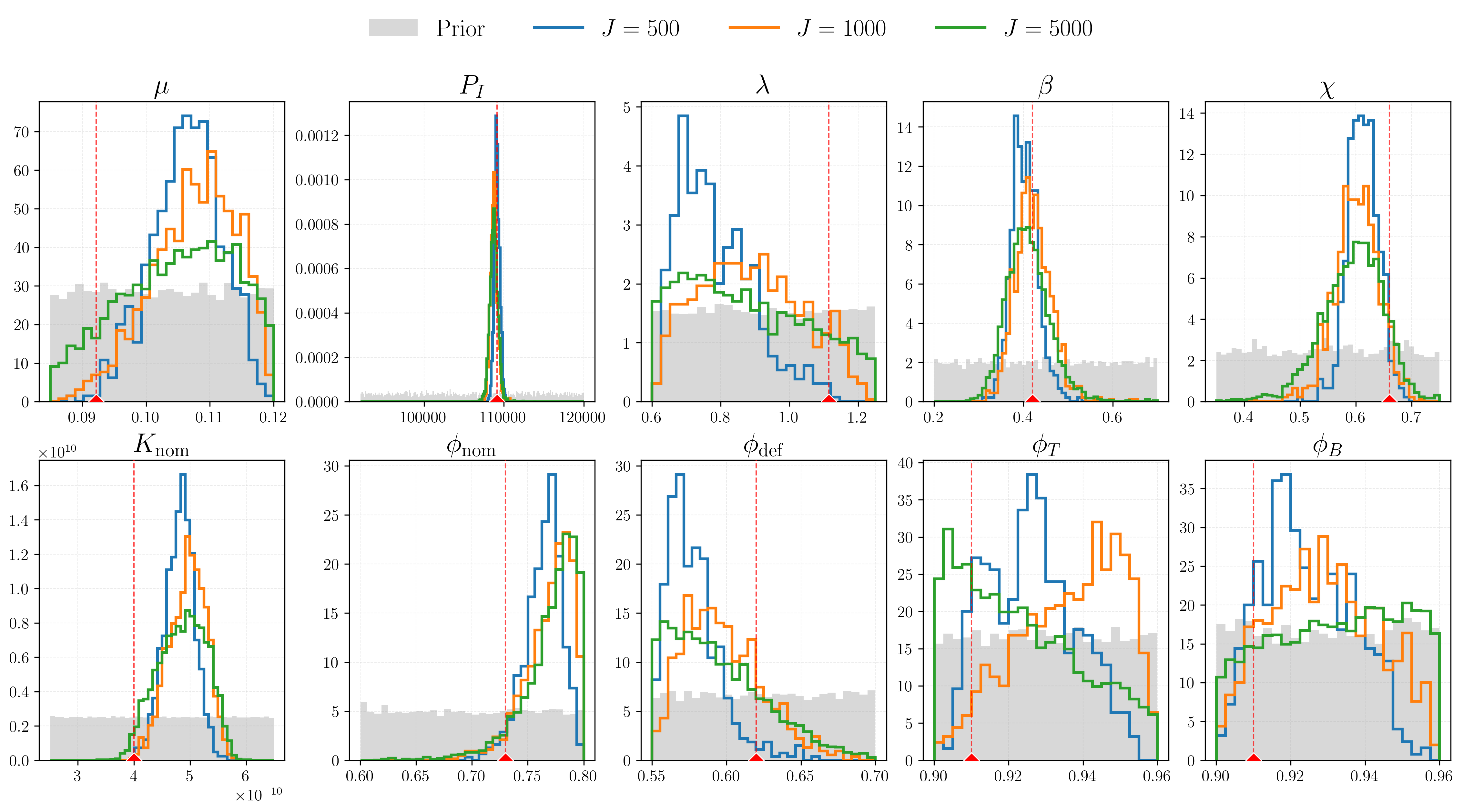}
    \caption{Histogram of priors and posteriors of scalar parameters for the $M=23$ sensor configuration, along with the true values (red).}
    \label{fig:full_EKI1B}
\end{figure*}

Figures ~\ref{fig:full_EKI2B} and \ref{fig:full_EKI1B} show the marginal posterior histograms of the scalar parameters, with the prior (based on $J=5000$ samples) shown for reference and the true values indicated in red. These results reveal that posterior estimates of scalar parameters are considerably sensitive to the ensemble size, even for the denser sensing configuration with $M=100$ sensors. While some parameters, such as $P_{I}$ and $\beta$, are well identified, with posteriors sharply concentrated around the true values, others, including $\lambda$ and the porosities $\phi_{T}$ and $\phi_{B}$ in the race-tracking regions, remain close to their priors even for $J=5000$. This indicates that these parameters are only weakly informed by the available pressure data and remain highly uncertain. 

Additional parameters, such as the viscosity and nominal permeability and porosity, are only partially constrained, with the true values lying in the tails of their posterior distributions and, for smaller ensemble sizes, not captured accurately at all. These findings highlight that, even under dense sensing configurations, the inverse problem is not uniquely identifiable and that substantial posterior uncertainty persists for certain parameters.

This behaviour can be further understood by inspecting the governing equations, i.e. Eqs. \eqref{eq: incompressibility} to \eqref{eq:press}. For a fixed time and a given saturated domain, the pressure equation does not depend explicitly on the viscosity $\mu$, and the pressure field is therefore invariant under simultaneous scalings of permeability and viscosity. Instead, the evolution of the moving front depends on the ratio $K/\mu$ (hydraulic conductivity), which determines the Darcy velocity. As a consequence, multiple combinations of permeability and viscosity, corresponding to the same hydraulic conductivity, can give rise to indistinguishable pressure measurements and similar front dynamics. This intrinsic non-identifiability explains the weak posterior contraction observed for $\mu$ and certain permeability parameters when considered individually.

\begin{figure*}
    \centering
    \includegraphics[width=0.6\linewidth]{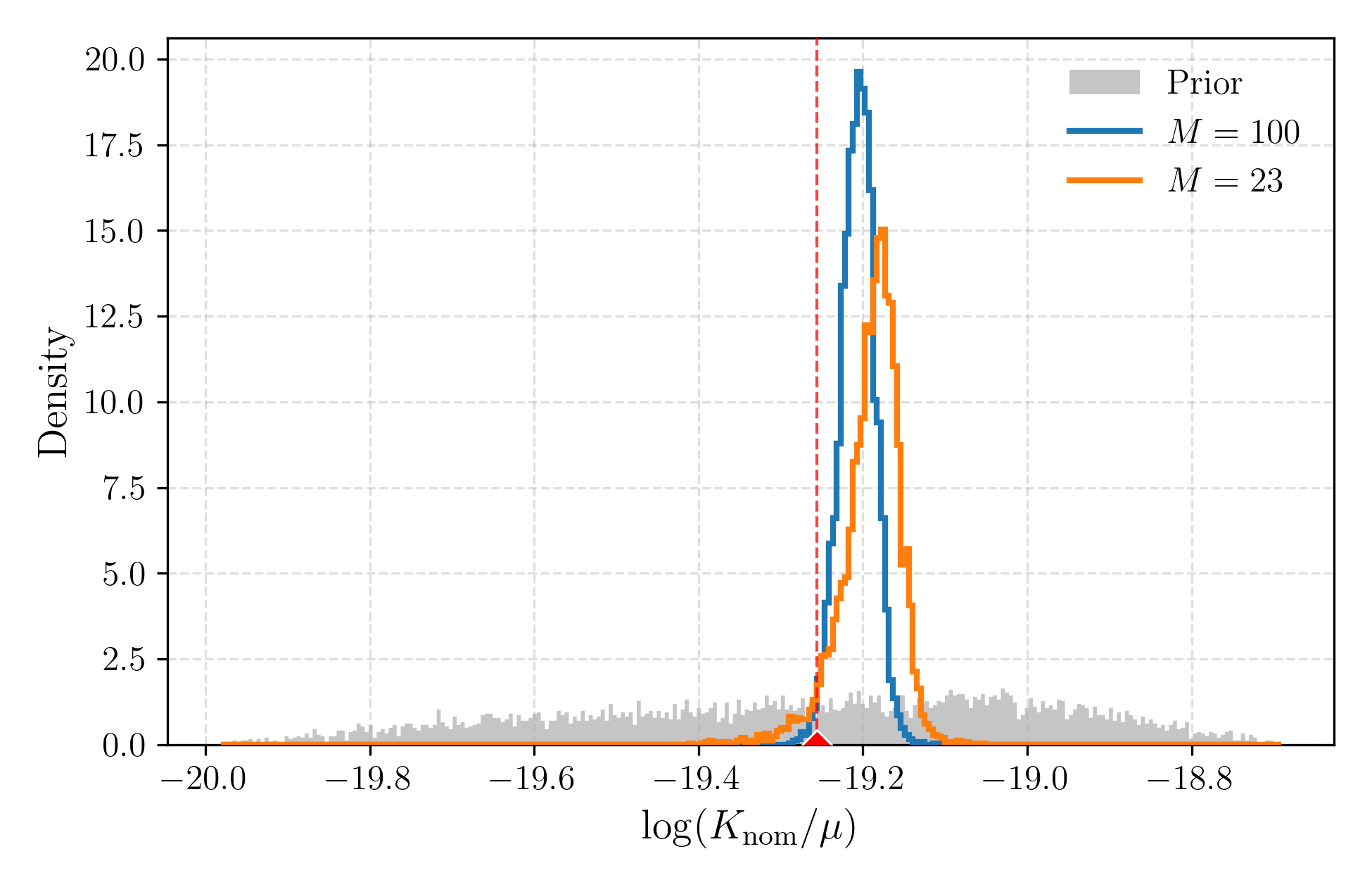}
    \caption{Histogram of priors and posteriors of log hydraulic conductivity for the $M=23$ and $M=100$ sensor configurations, along with the true values (red).}
    \label{fig:full_EKI_hydro}
\end{figure*}

To illustrate this effect, Fig.~\ref{fig:full_EKI_hydro} shows the marginal posterior histograms of $\log(K_{\mathrm{nom}}/\mu)$ for an ensemble size $J=5000$ and for the two sensor configurations considered. Although the true value remains located in the tails of the posterior distribution, the variance of the posterior is substantially reduced compared to the prior. This indicates that while the individual parameters $K_{\mathrm{nom}}$ and $\mu$ are only weakly identifiable, their ratio is significantly better constrained by the available data.

While reasonable accuracy can be achieved with relatively small ensemble sizes, the results of this section indicate that larger ensembles are preferable in practice, particularly under sparse sensing configurations. However, the associated computational cost rapidly becomes prohibitive. In the following section, a DeepONet is trained as a neural operator to approximate $\mathcal{G}$, thereby reducing the computational cost of EKI by several orders of magnitude while retaining accuracy in posterior inference.


\begin{table}[ht]
\centering
\caption{EKI performance versus ensemble size $J$ for two sensor layouts.}
\label{tab:eki_perf_J}
\begin{tabular}{@{}r ccc ccc@{}}
\toprule
& \multicolumn{3}{c}{$M=100$ sensors} & \multicolumn{3}{c}{$M=23$ sensors} \\
\cmidrule(lr){2-4}\cmidrule(lr){5-7}
$J$ & Time (min) & $N_{\text{iter}}$ & \#~$\GG$ evals & Time (min) & $N_{\text{iter}}$ & \#~$\GG$ evals \\
\midrule
500  & 12.99 & 9 & 4500 & 18.48 & 13 & 6500 \\
1000 & 30.11 & 11 & 11000 & 43.52 & 16 & 16000\\
5000 & 160.24 & 12 & 60000 & 212.96 & 16 & 80000 \\
\bottomrule
\end{tabular}\label{tab:full_EKI}
\end{table}

\section{Surrogate modelling via neural operators} \label{sec: Surrogate}

The primary computational bottleneck of EKI, and of any other sampling method in general, arises from the need to evaluate the parameter-to-measurement map $\GG(\bfu)$ a large number of times. Following from Eq. \eqref{eq:param10}, this involves the following compositions:
\[
\GG : \bfu 
   \xrightarrow{\;\;\mathcal{P}\;\;} (K,\phi,\mu,P_{I},\lambda,\beta,\chi)
   \xrightarrow{\;\;\FF\;\;} (p,f)
   \xrightarrow{\;\;\OO\;\;} \GG(\bfu).
\]
Thus, the forward operator is invoked for every proposed sample. Note that the cost of evaluating $\mathcal{P}$ and  $\OO$ is negligible compared to the evaluation of $\FF$, which requires solving the moving boundary problem described in Section~\ref{sec:forward_operator}, making it computationally intensive. Since $\GG$ must be recomputed for every proposed sample during inference, direct use of $\FF$ becomes the principal bottleneck.

To reduce this burden, a neural operator $\FF_{s}$ is introduced, with trainable parameters $\theta$ designed to approximate $\FF$:
\begin{eqnarray}
\FF[\bfu_{\text{FM}}](\bfx,t)=\left[\begin{array}{c}
    p[\bfu_{\text{FM}}](\bfx,t)\\
    f[\bfu_{\text{FM}}](\bfx,t)\end{array}\right]\approx \left[\begin{array}{c}
    p_{s}[\bfu_{\text{FM}}](\bfx,t)\\
    f_{s}[\bfu_{\text{FM}}](\bfx,t)\end{array}\right]=:\FF_{s}[\bfu_{\text{FM}};\theta](\bfx,t), \quad \forall (\bfx,t)\in D\times [0,T].
\end{eqnarray}
In contrast to most existing approaches for RTM inversion (see e.g. \citep{stieber2023inferring,MC,González} and references therein), the goal is not to approximate a fixed discretisation of the PDE, but rather to approximate the operator $\FF$ itself, that is, the pressure and filling factors at any point in space and time. This operator-learning perspective ensures that the surrogate captures the mapping between inputs and outputs at a functional level, thereby generalising across discretisations, domains, and experimental setups. In particular, different inlet pressure conditions, fluid viscosities, and sensor configurations can be treated within a single emulator.

A powerful approach for such operator learning is the DeepONet, first proposed by \citet{lu2021learning}. DeepONet extends classical neural networks by learning mappings between infinite-dimensional Banach spaces, thereby providing a principled way to approximate nonlinear operators. Since its introduction, DeepONet and its variants have been successfully applied to a broad range of problems, including solving parametric PDEs \citep{lu2021learning,LI2025128675,Lu2022}. In this work, a variant of DeepONet is employed to construct $\FF_{s}$ as a surrogate for the costly operator $\FF$. Once trained offline, $\FF_{s}$ yields sub-second evaluations of the forward map, making it ideally suited for use in EKI. 


\subsection{Building the DeepONet} \label{subsec: Data}

The DeepONet architecture consists of a branch network that encodes the input $\bfu_{\text{FM}}$ and a trunk network that encodes the query spatio-temporal location $(\bfx,t)$ on which the predictions (pressure and filling factor) are required. Both trunk and branch networks typically contain several hidden layers and, in the standard version of DeepONet, their output layers are combined through an inner product. This work adopts a variant inspired by the Extended DeepONet recently proposed by \citet{LI2025128675}, where the output of the branch network is not only combined with the trunk output but is also used to modulate each of the hidden layers of the multilayer perceptron (MLP) for the trunk network. This richer interaction between branch and trunk improves expressiveness and allows the surrogate to better capture complex nonlinear operator mappings.

\paragraph{Branch network}  
The branch network encodes the inputs $\bfu_{\text{FM}}$ into a latent representation $\mathcal{B}(\bfu_{\text{FM}}) \in \mathbb{R}^{N_{\text{out}}}$. Recall that each input $\bfu_{\text{FM}} = (\log K, \phi, \mu, P_{I}, \lambda, \beta, \chi)$ contains two functional components $(\log K, \phi)$ and five scalar components $(\mu, P_{I}, \lambda, \beta, \chi)$. Accordingly, two input branch networks are employed: (1) a convolutional U-Net encoder applied to the field inputs $(\log K,\phi)$, yielding $\mathcal{B}_{\text{fields}}$, and (2) a MLP applied to the scalar inputs $(\mu, P_{I}, \lambda, \beta, \chi)$, yielding $\mathcal{B}_{\text{scalars}} \in \mathbb{R}^{N_{\text{out}}}$. The scalar MLP consists of 3 hidden layers.  

For the field encoder, the raw inputs $(\log K, \phi)$ are augmented with positional encodings of the spatial grid coordinates before being passed to the U-Net. Incorporating positional information is standard practice in neural operator learning and related architectures, as it provides the network with explicit knowledge of spatial location and scale. This enrichment helps the branch network disentangle spatial variability in the input fields and facilitates alignment with the trunk network that processes spatio-temporal queries.  

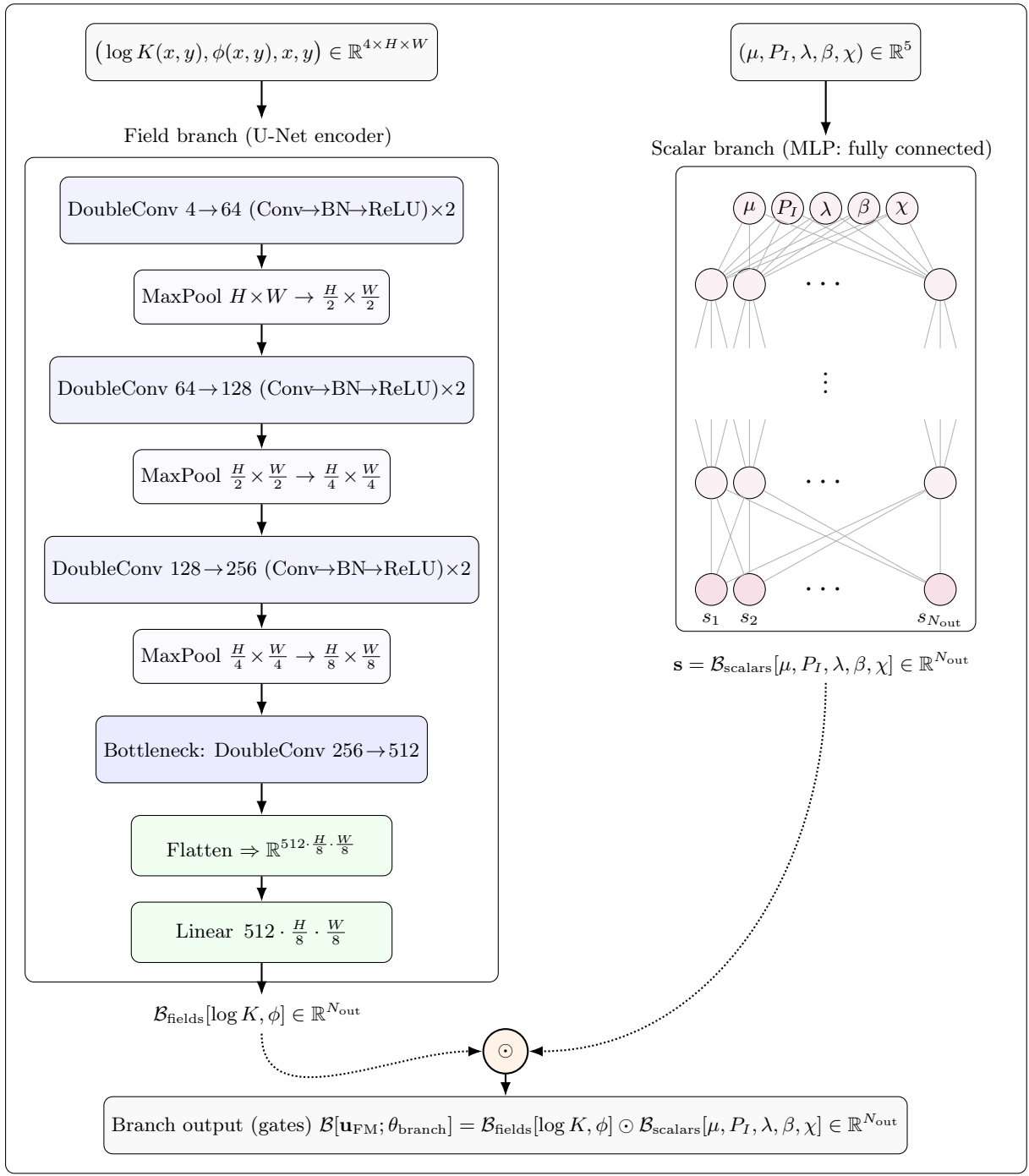
\begin{figure}
\centering
\begin{tikzpicture}[
  font=\small,
  >=Latex,
  arr/.style={-{Latex[length=2.2mm]}, thick},
  blk/.style={draw, rounded corners, minimum width=2.9cm, minimum height=0.9cm, fill=gray!5},
  conv/.style={draw, rounded corners, minimum width=3.8cm, minimum height=1.05cm, fill=blue!5},
  pool/.style={draw, rounded corners, minimum width=3.8cm, minimum height=0.85cm, fill=blue!2},
  bott/.style={draw, rounded corners, minimum width=3.8cm, minimum height=1.05cm, fill=blue!8},
  head/.style={draw, rounded corners, minimum width=4.1cm, minimum height=0.95cm, fill=green!6},
  prod/.style={draw, circle, minimum size=7mm, thick, fill=orange!10},
  neuron/.style={circle, draw, fill=purple!6, minimum size=5.0mm, inner sep=0pt},
  dots/.style={font=\Large, inner sep=0pt},
  box/.style={draw, rounded corners, inner sep=3mm}
]

\node[blk] (field_in) {$\displaystyle \big(\log K(x,y),\phi(x,y),x,y\big) \in \mathbb{R}^{4\times H \times W}$};

\node[conv, below=15mm of field_in] (enc1) {DoubleConv $4\!\to\!64$ (Conv$\!\to$BN$\!\to$ReLU)$\times 2$};
\node[pool, below=4mm of enc1] (pool1) {MaxPool $H\!\times\!W \to \frac{H}{2}\!\times\!\frac{W}{2}$};

\node[conv, below=5mm of pool1] (enc2) {DoubleConv $64\!\to\!128$ (Conv$\!\to$BN$\!\to$ReLU)$\times 2$};
\node[pool, below=4mm of enc2] (pool2) {MaxPool $\frac{H}{2}\!\times\!\frac{W}{2} \to \frac{H}{4}\!\times\!\frac{W}{4}$};

\node[conv, below=5mm of pool2] (enc3) {DoubleConv $128\!\to\!256$ (Conv$\!\to$BN$\!\to$ReLU)$\times 2$};
\node[pool, below=4mm of enc3] (pool3) {MaxPool $\frac{H}{4}\!\times\!\frac{W}{4} \to \frac{H}{8}\!\times\!\frac{W}{8}$};

\node[bott, below=5mm of pool3] (bottleneck) {Bottleneck: DoubleConv $256\!\to\!512$};

\node[head, below=5mm of bottleneck] (flatten) {Flatten $\Rightarrow \mathbb{R}^{512\cdot \frac{H}{8}\cdot \frac{W}{8}}$};
\node[head, below=4mm of flatten] (headlin) {Linear $\,512\cdot\frac{H}{8}\cdot\frac{W}{8}$};

\node[box, fit=(enc1)(pool1)(enc2)(pool2)(enc3)(pool3)(bottleneck)(flatten)(headlin)] (unetbox) {};

\node[below=6mm of field_in.south, align=center] (unetlabel) {%
  Field branch (U-Net encoder)
};

\draw[arr] (field_in.south) -- (unetlabel.north);  

\node[below=2mm of unetbox.south, align=center] (unetlabel2) {%
  $\mathcal{B}_{\text{fields}}[\log K,\phi]\in \RR^{N_{\text{out}}}$
};

\draw[arr] (enc1) -- (pool1);
\draw[arr] (pool1) -- (enc2);
\draw[arr] (enc2) -- (pool2);
\draw[arr] (pool2) -- (enc3);
\draw[arr] (enc3) -- (pool3);
\draw[arr] (pool3) -- (bottleneck);
\draw[arr] (bottleneck) -- (flatten);
\draw[arr] (flatten) -- (headlin);
\draw[arr] (headlin) -- (unetlabel2);

\node[blk, right=4.6cm of field_in] (scalar_in)  {$(\mu,P_{I},\lambda,\beta,\chi) \in \mathbb{R}^{5}$};

\coordinate (mlpX) at ($(scalar_in.south) + (0,-2.0)$);

\def\nsep{0.6}     
\def\lysep{1.2}    

\node[neuron] (in1) at ($(mlpX)+(-2*\nsep,0)$) {$\mu$};
\node[neuron] (in2) at ($(mlpX)+(-1*\nsep,0)$) {$P_{I}$};
\node[neuron] (in3) at ($(mlpX)+(0,0)$)        {$\lambda$};
\node[neuron] (in4) at ($(mlpX)+(\nsep,0)$)    {$\beta$};
\node[neuron] (in5) at ($(mlpX)+(2*\nsep,0)$)  {$\chi$};

\node[neuron] (hA1) at ($(mlpX)+(-3*\nsep,-\lysep)$) {};
\node[neuron] (hA2) at ($(mlpX)+(-2*\nsep,-\lysep)$) {};
\node[dots]   (hAdots) at ($(mlpX)+(0,-\lysep)$) {$\cdots$};
\node[neuron] (hA4) at ($(mlpX)+( 3*\nsep,-\lysep)$) {};

\node[dots] (vhdots) at ($(mlpX)+(0,-2.2*\lysep)$) {$\vdots$};

\node[neuron, fill=purple!12, label=below:$s_{1}$]   (o1) at ($(mlpX)+(-3*\nsep,-5.0*\lysep)$) {};
\node[neuron, fill=purple!12, label=below:$s_{2}$]   (o2) at ($(mlpX)+(-2*\nsep,-5.0*\lysep)$) {};
\node[dots] (oDots) at ($(mlpX)+(0,-5.0*\lysep)$) {$\cdots$};
\node[neuron, fill=purple!12, label=below:$s_{N_{\text{out}}}$]   (o4) at ($(mlpX)+( 3*\nsep,-5.0*\lysep)$) {};

\foreach \a in {in1,in2,in3,in4,in5}{
  \foreach \b in {hA1,hA2,hA4}{
    \draw[gray!60] (\a) -- (\b);
  }
}

\foreach \n in {hA1,hA2,hA4}{
  \draw[gray!60] (\n) -- ($(\n)+(-0.25,-1.0)$);
  \draw[gray!60] (\n) -- ($(\n)+( 0.00,-1.0)$);
  \draw[gray!60] (\n) -- ($(\n)+( 0.25,-1.0)$);
}

\node[neuron] (hB1) at ($(mlpX)+(-3*\nsep,-3.6*\lysep)$) {};
\node[neuron] (hB2) at ($(mlpX)+(-2*\nsep,-3.6*\lysep)$) {};
\node[dots]   (hBdots) at ($(mlpX)+(0,-3.6*\lysep)$) {$\cdots$};
\node[neuron] (hB4) at ($(mlpX)+( 3*\nsep,-3.6*\lysep)$) {};

\foreach \a in {hB1,hB2,hB4}{
  \foreach \b in {o1,o2,o4}{
    \draw[gray!60] (\a) -- (\b);
  }
}
\foreach \n in {hB1,hB2,hB4}{
  \draw[gray!60] (\n) -- ($(\n)+(-0.25,1.0)$);
  \draw[gray!60] (\n) -- ($(\n)+( 0.00,1.0)$);
  \draw[gray!60] (\n) -- ($(\n)+( 0.25,1.0)$);
  }

\node[box, fit=(in1)(in5)(o1)(o4), inner ysep=4mm] (mlpbox) {};

\node[below=8mm of scalar_in.south, align=center] (mlplabel) {%
  Scalar branch (MLP: fully connected)
};

\draw[arr] (scalar_in.south) -- ++(0,-4mm) -- ($(mlplabel.south)+(0,5mm)$);


\node[below=2mm of mlpbox.south, align=center] (mlplabel2) {%
  $\mathbf{s}=\mathcal{B}_{\text{scalars}}[\mu, P_{I}, \lambda, \beta, \chi] \in \RR^{N_{\text{out}}}$
};

\coordinate (outs_center) at (o2.south);
\coordinate (fusion_anchor) at ($(headlin.south)!0.5!(outs_center)$);

\node[prod] (prodnode) at ($(fusion_anchor)+(0,-4.2cm)$) {$\odot$};

\draw[arr, densely dotted] (unetlabel2) to[out=-90, in=180] (prodnode); 
\draw[arr, densely dotted] (mlplabel2) to[out=-90, in=0] (prodnode);

\node[below=3.2mm of prodnode] {\scriptsize elementwise product};
\node[blk, minimum width=5.4cm] (gates)
      at ($(prodnode.south) + (0,-8mm)$)
      {Branch output (gates) $\mathcal{B}[\bfu_{\text{FM}};\theta_{\text{branch}}] = \mathcal{B}_{\text{fields}}[\log K,\phi] \odot \mathcal{B}_{\text{scalars}}[\mu, P_{I}, \lambda, \beta,\chi]\in \RR^{N_{\text{out}}}$};
      
\draw[arr] (prodnode) -- (gates);


 \node[box, fit=(current bounding box)] {};

\end{tikzpicture}
\caption{Architecture of the DeepONet branch network. The field branch (left) is a U-Net encoder that processes spatially distributed inputs consisting of the log-permeability, porosity, and positional encodings of the spatial coordinates. Each \emph{DoubleConv} block denotes two successive convolutional layers with batch normalisation (BN) and ReLU activations. The scalar branch (right) is a fully connected multilayer perceptron (MLP) acting on global scalar parameters. The outputs of the two branches are combined via element-wise multiplication to form a gating vector that modulates the trunk network.}

\label{fig:deeponet_architecture}
\end{figure}

The two branch outputs are then combined multiplicatively:  
\begin{equation}
\mathcal{B}[\bfu_{\text{FM}};\theta_{\text{branch}}] 
= \mathcal{B}_{\text{fields}} \odot \mathcal{B}_{\text{scalars}},
\end{equation}
where $\odot$ denotes element-wise multiplication, and $\theta_{\text{branch}}$ is the combined set of trainable parameters for $\mathcal{B}_{\text{fields}}$ and $\mathcal{B}_{\text{scalars}}$. An overview of the full architecture is shown in Fig.~\ref{fig:deeponet_architecture}.


The use of two distinct branch networks is consistent with recent extensions of DeepONet for multi-modal and multi-input operators (e.g., MIONet \citep{MIONet}, U-DeepONet \citep{UDeepONet}, EDeepONet \citep{LI2025128675}). These studies highlight that treating heterogeneous inputs (fields vs.\ scalars, or functions vs.\ parameters) with separate encoders improves both expressivity and training efficiency, as each encoder can be tailored to the nature of the input. While both branch networks can, in principle, be chosen as multilayer perceptrons, for highly variable fields a CNN-based architecture is often preferable to efficiently extract spatial features. Here, the CNN-based encoder extracts spatial correlations from the field data $(\log K,\phi)$, while the MLP is well suited for the lower-dimensional scalar parameters. The multiplicative fusion then captures their joint effect in the latent representation. In summary, the branch network provides a \emph{global context} vector that conditions the surrogate model on the full set of physical parameters.  

\paragraph{Trunk network}  
The trunk network takes as input the space-time coordinates $(\bfx,t) \in D \times [0,T] \subset \mathbb{R}^3$, which are first mapped into a higher-dimensional Fourier feature space. This embedding enriches the representation of both spatial and temporal variations by providing sinusoidal basis functions at multiple frequencies. The embedded input is then processed by a sequence of fully connected layers.  

At each layer, the pre-activations are modulated by a shared gating signal provided by the branch network output $\mathcal{B}[\bfu_{\text{FM}};\theta_{\text{branch}}]$. The gating is applied in an affine, element-wise manner (FiLM-style) and followed by a nonlinear GELU activation. This procedure is repeated across all trunk layers, producing a hidden representation that integrates the coordinate information with the branch-supplied modulation.  

Finally, the hidden representation is split into two disjoint channel groups: one of dimension $N_{\text{out}}^{p}$ for the pressure head and one of dimension $N_{\text{out}}^{f}$ for the filling-factor head. The filling-factor output is obtained by applying a logistic sigmoid to the corresponding linear head, ensuring values lie in $[0,1]$. Each slice is thus mapped to a scalar output through its dedicated linear layer. The detailed forward pass of the trunk is summarised in Algorithm~\ref{alg:trunk}. Based on preliminary experiments, the values $L=6$ and $N_{\mathrm{freq}}=6$ were found to yield satisfactory performance across the considered test cases, although a comprehensive ablation study of these architectural choices is beyond the scope of the present work.

In summary, the trunk network provides a \emph{local coordinate embedding} that captures spatial and temporal variability, which is then modulated by the global context supplied by the branch.  

\SetAlgoNlRelativeSize{-1}
\SetKwInput{KwInput}{Input}
\SetKwInput{KwOutput}{Output}

\begin{algorithm}
\caption{Forward pass of trunk network}
\label{alg:trunk}
\LinesNumbered
\DontPrintSemicolon
\KwInput{
Raw trunk input $\mathbf{z}=(\bfx,t)\in\mathbb{R}^{3}$;
Branch input $\bfu_{\mathrm{FM}}=(K,\phi,\mu,P_{I},\lambda,\beta,\chi)$; Number of Fourier frequencies $N_{\mathrm{freq}}$;Number of layers $L$; Output dimensions $(N_{\mathrm{out}}^{p},N_{\mathrm{out}}^{f})$ with
$N_{\mathrm{out}}=N_{\mathrm{out}}^{p}+N_{\mathrm{out}}^{f}$.
}

\textbf{Fourier embedding of trunk input:} $\displaystyle \gamma(\mathbf{z}) \gets 
\Big[
\sin(2\pi\,2^k \mathbf{z}),
\cos(2\pi\,2^k \mathbf{z})
\Big]_{k=0}^{N_{\mathrm{freq}}-1}
\in \mathbb{R}^{6N_{\mathrm{freq}}}$\;
\BlankLine
\textbf{Augment trunk input:} $\displaystyle \tilde{\mathbf{z}} \gets [\mathbf{z},\gamma(\mathbf{z})]
\in \mathbb{R}^{3+6N_{\mathrm{freq}}}$\;
\BlankLine

\textbf{Get branch network:}  $\displaystyle g \gets\mathcal{B}[\bfu_{\text{FM}};\theta_{\text{branch}}]  \in \mathbb{R}^{N_{\text{out}}}$;\
\BlankLine
\textbf{Trunk encoder:} $\displaystyle       \mathbf{z}^{(0)} \gets W^{(\text{enc})}\tilde{\mathbf{z}} + b^{(\text{enc})}$\;
\BlankLine
\For{$\ell = 0$ to $L-1$}{
    \BlankLine
    Form per-layer gate: $\displaystyle \tilde{g}^{(\ell)} \gets a^{(\ell)} \odot g + b_g^{(\ell)}$\;
    \BlankLine
    Apply gating and nonlinearity: $\displaystyle   h^{(\ell)} \gets \textrm{GELU}\big(\mathbf{z}^{(\ell)} \odot \tilde{g}^{(\ell)}\big)$\;
    \BlankLine
    \If{$\ell < L-1$}{
    \BlankLine
        Next pre-activation: $\displaystyle \mathbf{z}^{(\ell+1)} \gets W^{(\ell)} h^{(\ell)} + b^{(\ell)}$;\
        \BlankLine
    }
}

\textbf{Slice hidden representation:}
    $\displaystyle    h_p \gets h^{(L-1)}_{1:N_{\text{out}}^{p}}, \qquad 
      h_f \gets h^{(L-1)}_{N_{\text{out}}^{p}+1:N_{\text{out}}}$\;
\BlankLine

\textbf{Heads:} $\displaystyle        p_{\text{out}}[\bfu_{\text{FM}}](\bfx,t) \gets w_p^\top h_p + b_p, \qquad
      f_{\text{out}}[\bfu_{\text{FM}}](\bfx,t) \gets \textrm{Sigmoid}(w_f^\top h_f + b_f)$\;
\BlankLine
\KwOutput{ Outputs $(p_{\text{out}}, f_{\text{out}}) \in \mathbb{R}^2$.}


\end{algorithm}

The set of trunk network parameters given in Algorithm~\ref{alg:trunk} is denoted by
\[
\theta_{\text{trunk}} 
= \Big\{ W^{(\text{enc})}, b^{(\text{enc})}, 
          \{ W^{(\ell)}, b^{(\ell)} \}_{\ell=1}^{L-1},
          \{ a^{(\ell)}, b_g^{(\ell)} \}_{\ell=0}^{L-1},
          w_p, b_p, w_f, b_f \Big\},
\]
and the total set of trainable parameters by $\theta = \theta_{\text{branch}} \cup \theta_{\text{trunk}}$. The surrogate is finally defined as
\begin{align}
\mathcal{F}_{s}[\bfu_{\text{FM}};\theta](\bfx,t)
:= 
\begin{bmatrix} 
p_{s}[\bfu_{\text{FM}}](\bfx,t) \\[0.3em]
f_{s}[\bfu_{\text{FM}}](\bfx,t)
\end{bmatrix} 
= 
\begin{bmatrix} 
p_{\text{out}}[\bfu_{\text{FM}}](\bfx,t)\;\mathbb{I}_{ f_{\text{out}}[\bfu_{\text{FM}}](\bfx,t) > \delta } \\[0.3em]
f_{\text{out}}[\bfu_{\text{FM}}](\bfx,t)
\end{bmatrix}.
\label{eq:deep10}
\end{align}
In words, surrogate predictions for the pressure field are obtained by masking the raw pressure output of the DeepONet using an indicator function that enforces $f_{\text{out}}[\bfu_{\text{FM}}](\bfx,t)>\delta$. As a result, pressure values are retained only in regions where the network predicts that the flow front has passed. This construction is motivated by the physical assumption that pressure is defined only within the saturated region of the domain, corresponding to $f>0$. In the present work, the threshold is set to $\delta=0.9$ which reflects a conservative criterion for classifying a point as saturated, ensuring that pressure predictions are retained only where the network predicts a high degree of confidence that the front has passed. Values of $\delta$ closer to zero were observed to admit spurious pressure predictions in regions close to the interface, while values too close to unity led to unnecessary erosion of the predicted saturated region. In practice, surrogate predictions were found to be robust to moderate variations of $\delta$ in the range $[0.8,0.95]$, with $\delta=0.9$ providing a good balance between robustness to noise in the filling-factor prediction and accurate localisation of the moving front.


\subsection{Training the DeepONet} 

For training, losses are based on the squared $L^{2}$-error on the spatio-temporal domain $D\times [0,T]$. For pressure, this is defined by 
\begin{eqnarray}\label{eq:train1}
\norm{p_{\mathrm{out}}- p}_{L^{2}}^2= \int_{0}^{T}\!\!\int_{D} 
\Big( p(\bfx,t) - p_{\mathrm{out}}(\bfx,t) \Big)^2 \, d\bfx \, dt,
\end{eqnarray}
and a similar expression holds for the filling factor. The overall loss function is given by
\begin{eqnarray}\label{eq:train2}
\mathcal{L}\big[p,f,p_{\text{out}},f_{\text{out}} ;\theta\big] := \kappa\,\norm{f-f_{\mathrm{out}}}_{L^{2}}^2
+ \norm{p_{\mathrm{out}}- p}_{L^{2}}^2,
\end{eqnarray}
where $\kappa > 0$ is a tunable weight that balances the contribution of the filling factor and pressure terms. In the present work, $\kappa$ is fixed to $\kappa=0.05$ throughout training. This value was selected empirically so that the magnitudes of the two loss contributions are of comparable order during training, preventing either term from dominating the optimisation. No significant sensitivity to moderate variations of $\kappa$ around this value was observed.

Let $\pi$ denote the joint law induced by sampling $\bfu_{\mathrm{FM}}\sim\mathbb{P}(\bfu_{\mathrm{FM}})$ and setting $(p,f)=\mathcal{F}(\bfu_{\mathrm{FM}})$, i.e.
\(
\pi := \mathbb{P}(\bfu_{\mathrm{FM}})\otimes \mathcal{F}_{*}\mathbb{P}(\bfu_{\mathrm{FM}}).
\)

The training process is then formulated as the minimisation of the expected loss:
\begin{eqnarray}\label{eq:train3}
\mathcal{L}_{\text{train}}(\theta):= \mathbb{E}^{\pi}\,\mathcal{L}\big[p,f,p_{\text{out}},f_{\text{out}} ;\theta\big].
\end{eqnarray}
where the dependence on $\theta$ arises through the DeepONet trainable parameters that produce the predictions $p_{\mathrm{out}}$ and $f_{\mathrm{out}}$.

To approximate Eq. \eqref{eq:train3}, a set of training samples of the form
\begin{eqnarray}\label{eq:train4}
\mathcal{T} := \Big\{ \big(\bfu_{\text{FM}}^{(j)}, p^{(j)}, f^{(j)}\big) \Big\}_{j=1}^{N_{\text{train}}}  \label{eq:training}    
\end{eqnarray}
is considered, where, for convenience of notation, the definitions $p^{(j)} := p[\bfu_{\text{FM}}^{(j)}]$ and $f^{(j)} := f[\bfu_{\text{FM}}^{(j)}]$ are used. 
Analogously, the network predictions are denoted by $p_{\mathrm{out}}^{(j)} := p_{\mathrm{out}}[\bfu_{\text{FM}}^{(j)}]$ and $f_{\mathrm{out}}^{(j)} := f_{\mathrm{out}}[\bfu_{\text{FM}}^{(j)}]$. In Eq.~\eqref{eq:training}, $\bfu_{\text{FM}}^{(j)}\sim \mathbb{P}(\bfu_{\text{FM}})$, i.e. $\bfu_{\text{FM}}^{(j)}$ are samples from the prior on $\bfu_{\text{FM}}$ and, consequently, $(p^{(j)}, f^{(j)})\sim \FF_{*}\mathbb{P}(\bfu_{\text{FM}})$.

With this training set, the expected loss is approximated via the empirical risk:
\begin{eqnarray}\label{eq:train5}
\mathcal{L}_{\text{train}}(\theta)\approx \frac{1}{N_{\mathrm{train}}}\sum_{j=1}^{N_{\text{train}}}\mathcal{L}\big[p^{(j)},f^{(j)},p_{\text{out}}^{(j)},p_{\text{out}}^{(j)}\big].
\end{eqnarray}


\subsubsection{Discretisation}


During training, evaluating the loss integrals in Eq.~\eqref{eq:train1} over the full space--time domain is computationally expensive. As is standard in FEM, the spatial $L^2$ inner product is 
approximated using a quadrature rule, 
i.e., a diagonal approximation of the $L^2$ inner-product matrix, 
yielding a nodal weighted sum over the mesh \citep{ZienkiewiczTaylor}. This deterministic quadrature is then approximated in a stochastic manner during training by subsampling space-time points, resulting in a Monte Carlo approximation of the loss integrals, as commonly employed in neural operator and physics-informed learning settings \citep{Raissi2019PINNs,Kovachki2023NeuralOperators}. Concretely, let $\{\bfx_i\}_{i=1}^{S}$ denote the mesh nodes with the associated weights $\{w_i\}_{i=1}^{S}$ of the 
diagonal approximation of inner-product matrix, where $w_i>0$ and $\sum_{i=1}^{S} w_i = |D|$, and let $\{t_\ell\}_{\ell=1}^{N_t}$ be uniformly sampled time instances in $[0,T]$. The $L^2$-error is then approximated as
\[
\norm{p_{\mathrm{out}}^{(j)}- p^{(j)}}_{L^{2}}^2
\;\approx\; \frac{T}{N_t}\sum_{\ell=1}^{N_t}\sum_{i=1}^{S} w_i\,
\bigl(p^{(j)}(\bfx_i,t_\ell)-p_{\mathrm{out}}^{(j)}(\bfx_i,t_\ell)\bigr)^2,
\]
with an analogous expression for the filling factor.


\subsubsection{Implementation details}
DeepONet models are implemented in PyTorch and trained with Distributed Data Parallel (DDP) across 3 NVIDIA L40 GPUs (48~GB each) on the Tier 2 Midlands Plus HPC. A batch size of 32 per GPU is used, with data loading handled by DistributedSampler to ensure disjoint, epoch-wise shuffling across ranks. Inputs to the field branch network are normalised to the interval $[0,1]$ on a per-channel basis, while the branch scalars and the pressure targets are standardised. The filling-factor targets are naturally bounded in the interval $[0,1]$ by construction, and therefore no additional scaling or transformation is applied to these outputs during training.

Training is performed with weighted Adam, using an initial learning rate of $3\times 10^{-4}$ and a scheduler that halves the learning rate when the validation curve plateaus. Network parameters are checkpointed every 50 epochs, and evaluation metrics are computed offline on the testing set. All models are trained up to a maximum of 450 epochs, which serves as a common upper bound for reporting training time across architectures. Three latent output dimensions are considered, $N_{\text{out}} \in \{200,400,800\}$, and for simplicity $N_{\text{out}}^{p} = N_{\text{out}}^{f} = N_{\text{out}}/2$. Training times and number of trainable parameters $\vert \theta\vert$ are reported in Table~\ref{table:metrics0}. As expected, increasing the latent output dimension $N_{\text{out}}$ raises both the parameter count and the wall-clock training time. In addition, it is observed that the scaling is not strictly linear: the quadratic growth of certain weight matrices with $N_{\text{out}}$ leads to a disproportionately large increase in memory usage and communication overhead in DDP. Similarly, increasing the number of training samples naturally prolongs each epoch, but also reduces the variance of gradient estimates, which can in practice accelerate convergence and improve generalisation.

\subsubsection{Surrogate evaluation metrics} 
The DeepONet-based surrogate performance is tested on an unseen set of $N_{\text{test}}=10^4$ samples of the form given in Eq.~\eqref{eq:train4}, disjoint from the training set. For each test sample, the relative $L^2$-error is computed, and the mean and standard deviation is reported over the test set:
\begin{eqnarray}\label{eq:metrics}
\varepsilon_{p}^{(j)}
= \frac{\bigl\| p^{(j)} - p_{s}^{(j)} \bigr\|_{L^2}}
       {\bigl\| p^{(j)} \bigr\|_{L^2}},
\qquad
\varepsilon_{f}^{(j)}
= \frac{\bigl\| f^{(j)} - f_{s}^{(j)} \bigr\|_{L^2}}
       {\bigl\| f^{(j)} \bigr\|_{L^2}},
\end{eqnarray}
where $p_{s}^{(j)}$ and $f_{s}^{(j)}$ are surrogate predictions (see Eq. \eqref{eq:deep10}) obtained using inputs from test samples. The relative $L^2$-error is a natural metric for operator surrogates, as it normalises by the signal energy of each test case, ensuring scale invariance and allowing for fair comparison across heterogeneous samples. For the evaluation of the metrics in Eq.~\eqref{eq:metrics}, the $L^2$ norms are computed using the full discretised $L^2$ inner product induced by the finite element mass matrix on the computational mesh, without stochastic subsampling or quadrature approximation. In this way, the reported errors reflect the surrogate accuracy with respect to the full numerical solution.

\subsection{DeepONet performance}

Results are reported for a model with latent dimension $N_{\text{out}}=400$, trained using datasets of size $N_{\text{train}} \in \{10{,}000, 20{,}000, 40{,}000\}$. The smaller datasets are obtained by random subsampling of the full $40{,}000$-sample training set. Generating the full training dataset required approximately $1.8$~hours of wall-clock time using the same computational resources described in Section~\ref{sec:virtual_EKI}.

For each dataset, the samples are randomly split into 90\% for training and 10\% for validation. The aim here is to understand the effect of training data size on surrogate accuracy, which will be investigated further in the context of the inverse problem. Training data size is practically relevant since, in realistic scenarios (e.g., 3D simulations), each training sample may be very costly to generate, and data availability can be sparse.  

\begin{figure*}
    \centering
    \includegraphics[width=0.6\linewidth]{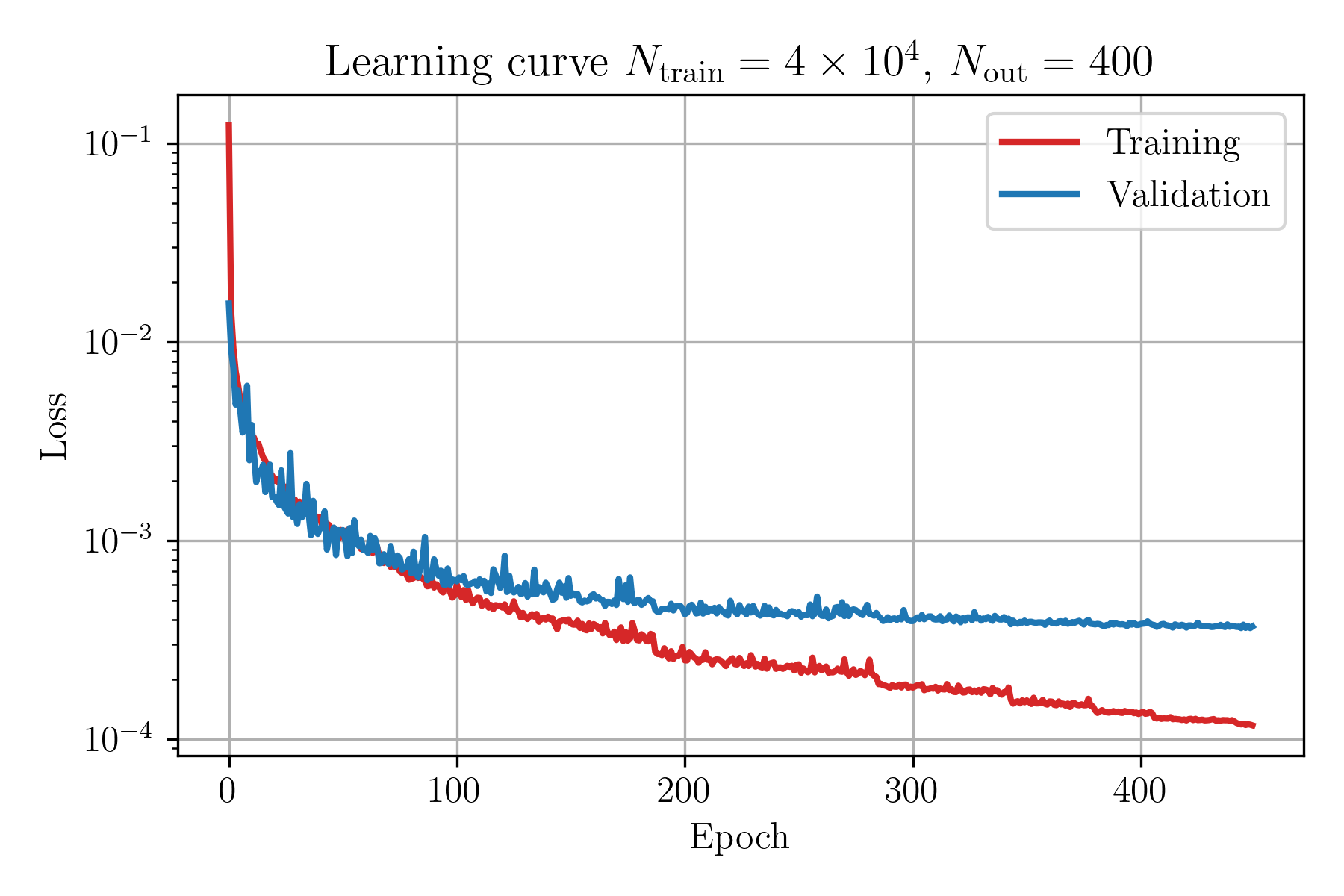}
    \includegraphics[width=0.6\linewidth]{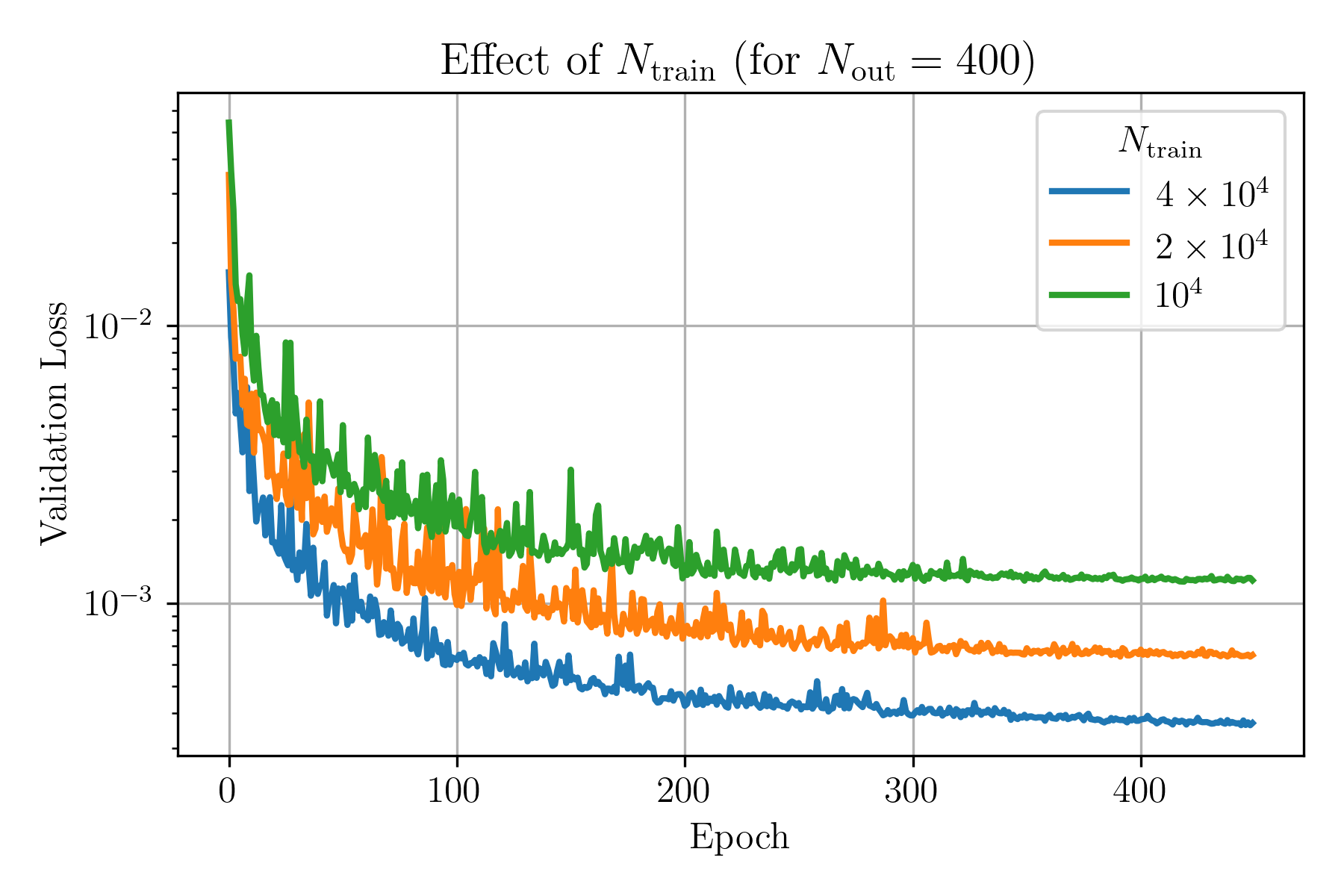}
    \includegraphics[width=0.6\linewidth]{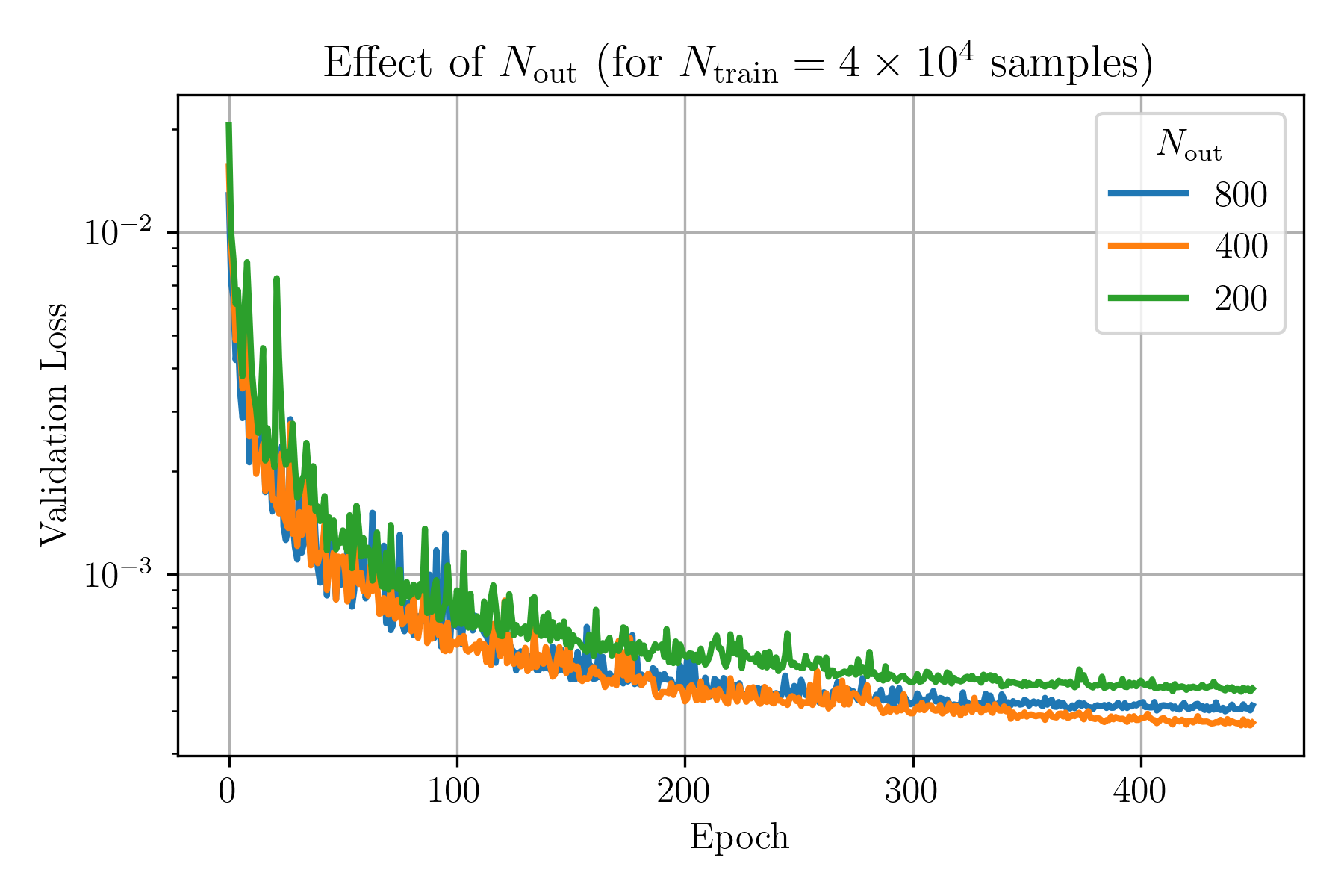}
    \caption{Top to bottom: Training and validation losses over 300 epochs with $N_{\text{train}}=4\times 10^4$ and $N_{\text{out}}=400$; validation loss for varied training set sizes with $N_{\text{out}}=400$ fixed; validation loss for varied latent dimensionality with $N_{\text{train}}=4\times 10^4$ fixed.}
    \label{fig:curves}
\end{figure*}

For the case with $N_{\text{train}}=40{,}000$, the top panel of Fig.~\ref{fig:curves} shows the training and validation curves of the loss per epoch. The validation plateaus after epoch 350 with no clear signs of overfitting. A similar trend is observed for the validation $L^2$-losses in the middle panel of Fig.~\ref{fig:curves}, where reducing the training set size leads to a higher plateau value of the loss. This is further confirmed in Table~\ref{table:metrics1}, which reports the mean $\pm$ standard deviation of the relative $L^2$-errors for the filling factor and pressure over the validation set. The larger training set achieves the best accuracy, with average errors of $1.19\%$ in pressure predictions and $3.73\%$ in the filling factor.

The higher error in the filling factor is expected: unlike pressure, which is a relatively smooth quantity across space and time, the filling factor represents the propagation of a sharp saturation front. Small misalignments in the predicted location of this front can result in disproportionately large $L^2$-errors, even if the overall dynamics are captured well. In other words, the surrogate must learn to reproduce highly non-smooth, discontinuous behaviour, which is inherently more challenging.  

Although the validation loss plateaus after approximately 350 epochs, the surrogate selection criterion is the relative $L^2$ error computed using the full discretised norm. For $N_{\mathrm{train}}=4\times10^{4}$, this test metric continues to decrease up to 450 epochs (e.g.\ $\epsilon_p$ from $1.26$ to $1.19$ and $\epsilon_f$ from $3.92$ to $3.73$ in Table~\ref{table:metrics1}), hence the 450-epoch checkpoint is retained. Training beyond 450 epochs was not pursued because improvements in the full test-set relative $L^2$ errors after 350 epochs were marginal compared with the variability across test samples, and additional optimisation may increase the risk of overfitting without commensurate gains. The 450-epoch model therefore represents a practical compromise between accuracy and computational cost.


The bottom panel of Fig.~\ref{fig:curves} shows validation curves obtained with three different choices of latent dimension. While increasing the latent dimension from $N_{\text{out}}=200$ to $400$ consistently improves surrogate accuracy, a further increase to $N_{\text{out}}=800$ leads to a slight degradation in test performance, despite continued decreases in training and validation loss. This behaviour suggests a transition to a variance-dominated regime, in which the model becomes over-parameterised relative to the available training data and begins to overfit subtle numerical and statistical features of the training distribution. The effect is modest and consistent across epochs, indicating that $N_{\text{out}}=400$ provides a favourable balance between expressive power and generalization.

The numerical results indicate that $N_{\text{train}}=40{,}000$ and $N_{\text{out}}=400$ provide a good compromise between accuracy, complexity, and training time. This configuration is adopted for the remainder of this work, although for the inverse problem the effect of training data size is further investigated. While further ablation studies are warranted, additional experiments suggest that the chosen architecture, consisting of a U-Net for the field branch, a 3-layer scalar branch, and a 6-layer trunk, strikes a reasonable balance between capacity and efficiency. This design enables the surrogate to encode fine-scale features in the input, which often include sharp discontinuities and very small regions of abrupt variation (e.g., due to race tracking).  

\begin{figure*}
    \centering
    \includegraphics[width=0.43\linewidth]{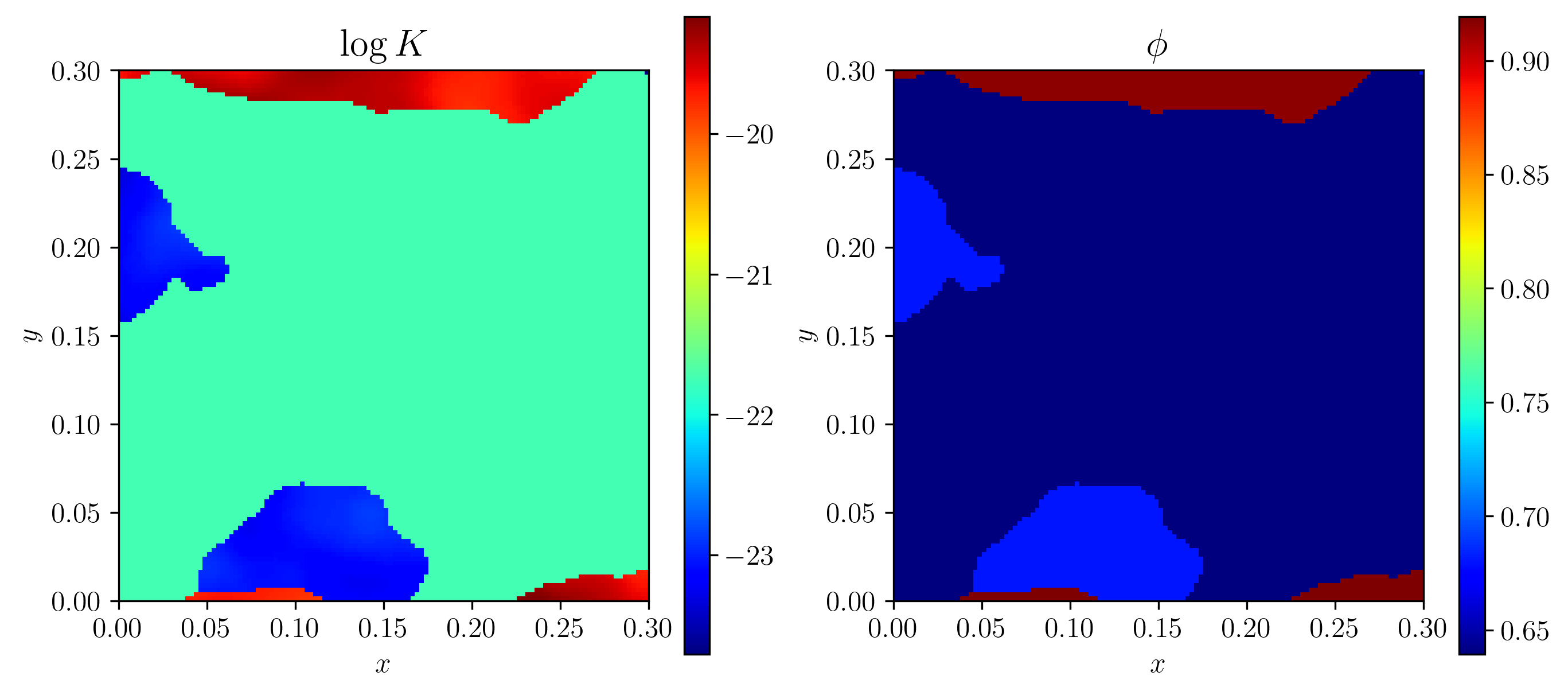}
    \includegraphics[width=0.86\linewidth]{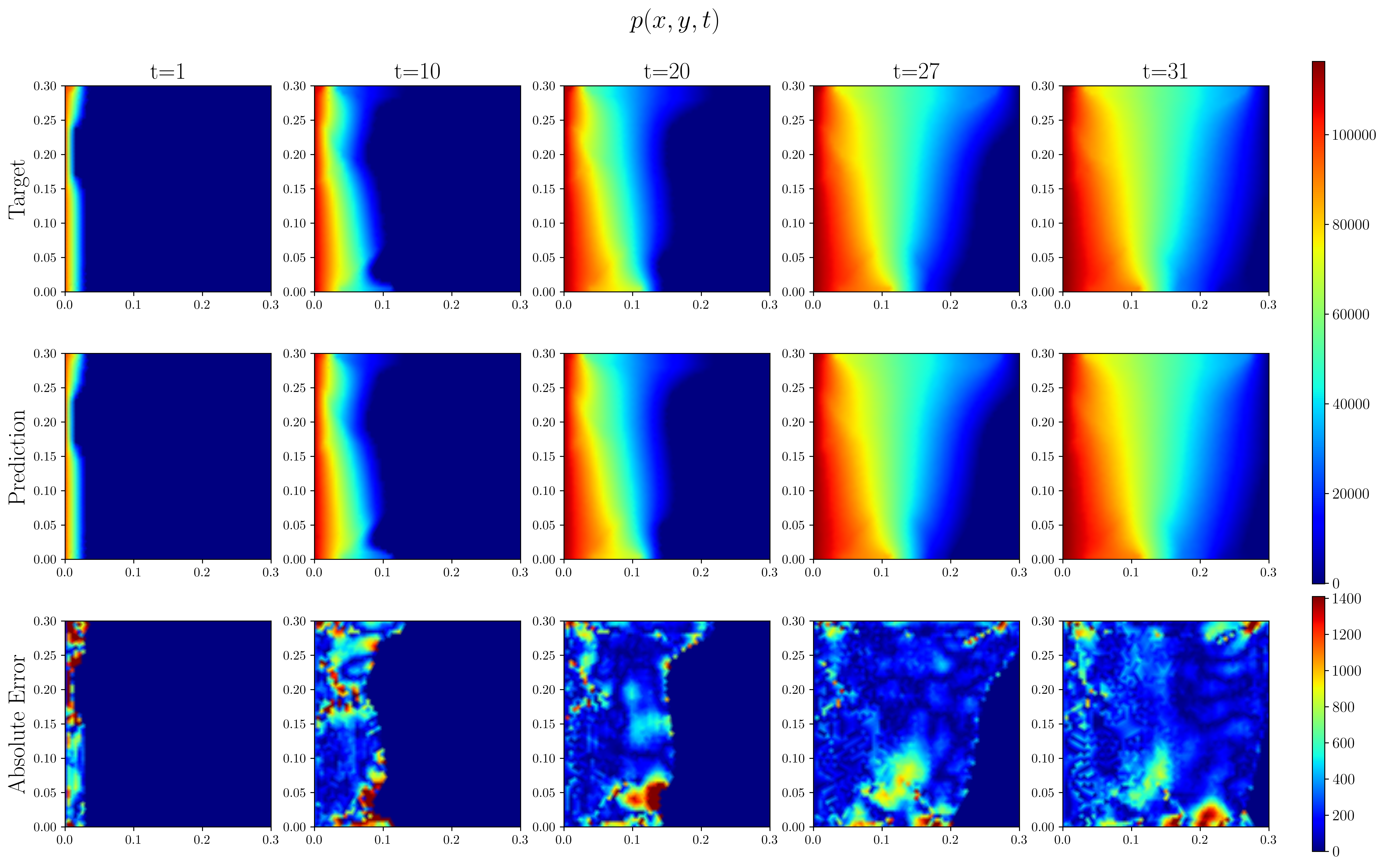}
    \includegraphics[width=0.86\linewidth]{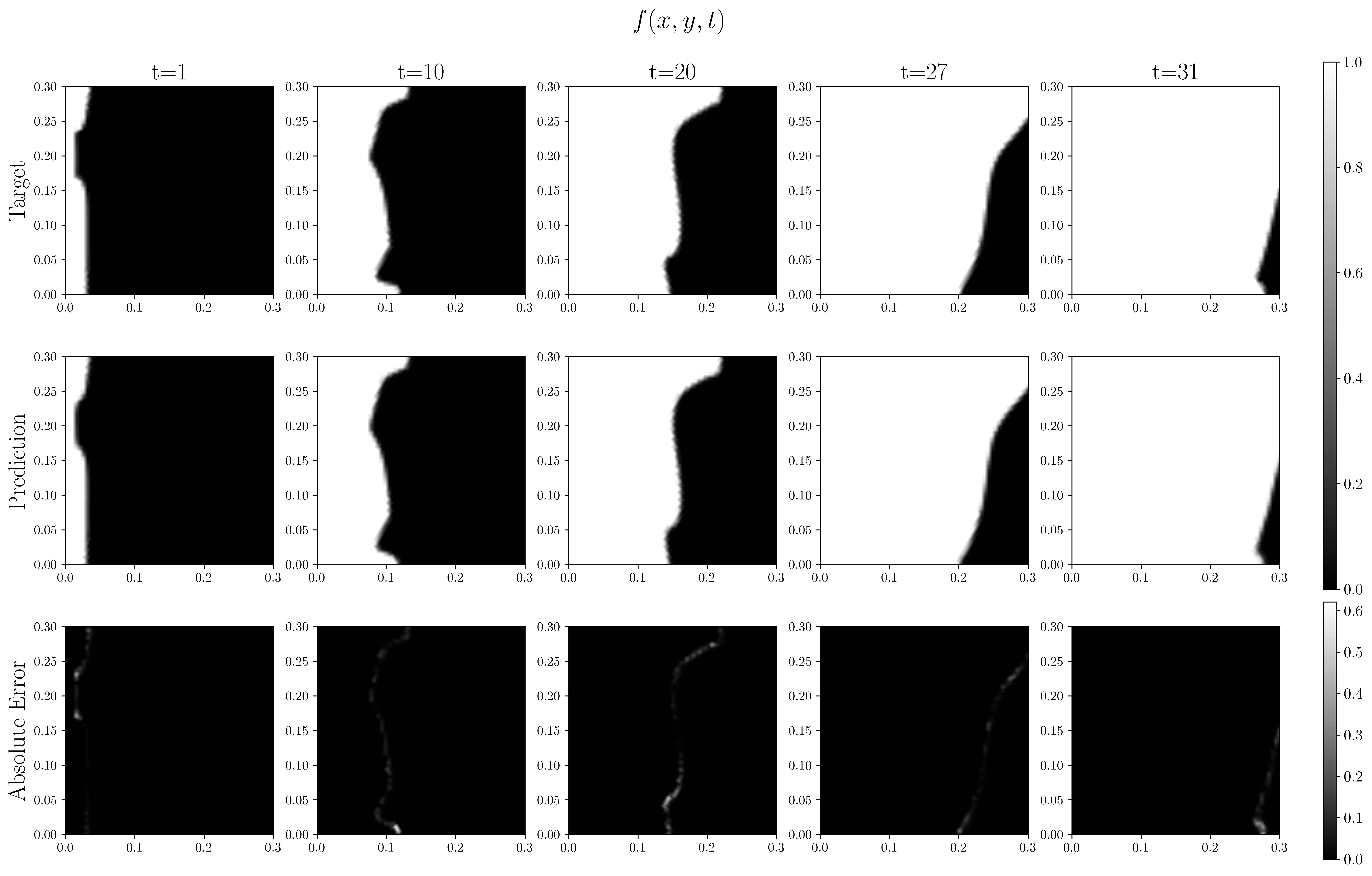}
    \caption{Top to bottom: Log-permeability and porosity, sampled from the chosen prior; CVFEM simulation and surrogate prediction for the pressure field at various observation times, along with the absolute error between them; CVFEM simulation and surrogate prediction for the filling factor at various observation times, along with the absolute error between them.}
    \label{fig:Em1}
\end{figure*}

To illustrate predictive capability, the top panel of Fig.~\ref{fig:Em1} shows $\log K$ and $\phi$ fields from a representative test case. Together with the scalar inputs 
$$(\mu, P_{I}, \lambda, \beta, \chi)=( 0.1074\,\text{Pa}\cdot \text{s}, 118.86\,\text{kPa},1.1358,  0.2625,  0.5537),$$
these fields produce the corresponding CVFEM simulator outputs (ground truth), surrogate predictions, and absolute errors for both pressure and filling factor, evaluated at the 2973 mesh nodes. For brevity, five representative time steps are shown. The surrogate captures the pressure distribution and the saturation front with high fidelity, demonstrating accurate recovery of both smooth fields and sharp fronts. In the next section, the performance of DeepONet-accelerated EKI is investigated.

\begin{table}[t]
\centering
\caption{Model complexity, $\vert \theta\vert$, and training time for varied training set size, $N_{\text{train}}$, and latent dimensionality, $N_{\text{out}}$.}
\label{tab:train_time_dataset}
\begin{tabular}{c c c c}
\toprule
$N_{\text{out}}$ &$N_{\text{train}}$ & $\vert \theta\vert$ & Training time \\
\midrule
200 &$4.0 \times 10^4$ & 52M & 9.6h \\
400 &$4.0 \times 10^4$ & 100M & 13.2h \\
800&$4.0 \times 10^4$ & 202M &  21.3h \\
400 &$1.0 \times 10^4$ & 100M & 3.6h \\
400 &$2.0 \times 10^4$ & 100M & 6.9h \\
\bottomrule
\end{tabular}\label{table:metrics0}
\end{table}

\begin{table}[t]
\centering
\caption{Mean/standard deviation of validation errors, defined in Eq.~\eqref{eq:metrics}, vs. epochs for varied dataset sizes ($N_{\text{out}}=400$).}
\label{tab:dataset_size}
\begin{tabular}{r c c c c c c }
\toprule
Epoch & \multicolumn{2}{c}{$N_{\text{train}}= 10^{4}$} & \multicolumn{2}{c}{$N_{\text{train}}= 2\times 10^{4}$} & \multicolumn{2}{c}{$N_{\text{train}}= 4\times 10^{4}$} \\

\cmidrule(lr){2-7}
 & $\epsilon_{p}\,\,[\times 10^{-2}]$ & $\epsilon_{f}\,\,[\times 10^{-2}]$& $\epsilon_{p}\,\,[\times 10^{-2}]$ & $\epsilon_{f}\,\,[\times 10^{-2}]$ & $\epsilon_{p}\,\,[\times 10^{-2}]$ & $\epsilon_{f}\,\,[\times 10^{-2}]$  \\
\midrule
50 & 3.75 $\pm$ 1.59 & 10.39 $\pm$ 2.85 & 3.15 $\pm$ 1.28 & 9.39 $\pm$ 2.56 & 2.39 $\pm$ 0.99 & 6.53 $\pm$ 2.05 \\
100 & 3.08 $\pm$ 1.48 & 8.83 $\pm$ 2.84 & 2.27 $\pm$ 1.10 & 6.94 $\pm$ 2.36 & 1.70 $\pm$ 0.85 & 5.16 $\pm$ 2.01 \\
150 & 2.89 $\pm$ 1.45 & 8.69 $\pm$ 2.79 & 2.32 $\pm$ 1.01 & 7.33 $\pm$ 2.19 & 1.71 $\pm$ 0.77 & 5.06 $\pm$ 1.87 \\
200 & 2.31 $\pm$ 1.51 & 6.88 $\pm$ 3.10 & 1.84 $\pm$ 1.03 & 5.52 $\pm$ 2.40 & 1.39 $\pm$ 0.74 & 4.31 $\pm$ 1.82 \\
250 & 2.46 $\pm$ 1.50 & 7.64 $\pm$ 3.09 & 1.74 $\pm$ 1.07 & 5.34 $\pm$ 2.53 & 1.34 $\pm$ 0.74 & 4.22 $\pm$ 1.83 \\
300 & 2.35 $\pm$ 1.58 & 7.10 $\pm$ 3.39 & 1.70 $\pm$ 1.07 & 5.39 $\pm$ 2.59 & 1.27 $\pm$ 0.73 & 3.99 $\pm$ 1.78 \\
350 & 2.24 $\pm$ 1.59 & 6.71 $\pm$ 3.49 & 1.64 $\pm$ 1.06 & 5.00 $\pm$ 2.50 & 1.26 $\pm$ 0.72 & 3.92 $\pm$ 1.81 \\
400 & 2.24 $\pm$ 1.59 & 6.74 $\pm$ 3.47 & 1.62 $\pm$ 1.07 & 5.04 $\pm$ 2.58 & 1.22 $\pm$ 0.71 & 3.80 $\pm$ 1.81 \\
450 & 2.22 $\pm$ 1.60 & 6.71 $\pm$ 3.57 & 1.62 $\pm$ 1.08 & 5.18 $\pm$ 2.62 & 1.19 $\pm$ 0.71 & 3.73 $\pm$ 1.83 \\

\bottomrule
\end{tabular} \label{table:metrics1}
\end{table}

\begin{table}[t]
\centering
\caption{Mean/standard deviation of validation errors, defined in Eq.~\eqref{eq:metrics}, vs. epochs for varied latent dimensionality ($N_{\text{train}}=4\times 10^4$).}
\label{tab:latent_dims}
\begin{tabular}{r c c c c c c }
\toprule
Epoch & \multicolumn{2}{c}{$N_{\text{out}}=200$} & \multicolumn{2}{c}{$N_{\text{out}}=400$} & \multicolumn{2}{c}{$N_{\text{out}}=800$} \\
\cmidrule(lr){2-7}
 & $\epsilon_{p}\,\,[\times 10^{-2}]$ & $\epsilon_{f}\,\,[\times 10^{-2}]$& $\epsilon_{p}\,\,[\times 10^{-2}]$ & $\epsilon_{f}\,\,[\times 10^{-2}]$ & $\epsilon_{p}\,\,[\times 10^{-2}]$ & $\epsilon_{f}\,\,[\times 10^{-2}]$  \\
\midrule
0 & 2.62 $\pm$ 1.10 & 7.63 $\pm$ 2.26 & 2.39 $\pm$ 0.99 & 6.53 $\pm$ 2.05 & 2.17 $\pm$ 0.96 & 6.48 $\pm$ 2.03 \\
100 & 1.79 $\pm$ 0.91 & 5.53 $\pm$ 2.12 & 1.70 $\pm$ 0.85 & 5.16 $\pm$ 2.01 & 1.89 $\pm$ 0.86 & 5.92 $\pm$ 1.90 \\
150 & 1.77 $\pm$ 0.85 & 5.38 $\pm$ 1.97 & 1.71 $\pm$ 0.77 & 5.06 $\pm$ 1.87 & 1.54 $\pm$ 0.83 & 4.83 $\pm$ 2.01 \\
200 & 1.62 $\pm$ 0.82 & 5.04 $\pm$ 1.94 & 1.39 $\pm$ 0.74 & 4.31 $\pm$ 1.82 & 1.43 $\pm$ 0.82 & 4.43 $\pm$ 1.96 \\
250 & 1.55 $\pm$ 0.78 & 4.70 $\pm$ 1.91 & 1.34 $\pm$ 0.74 & 4.22 $\pm$ 1.83 & 1.35 $\pm$ 0.83 & 4.18 $\pm$ 1.99 \\
300 & 1.42 $\pm$ 0.80 & 4.40 $\pm$ 1.97 & 1.27 $\pm$ 0.73 & 3.99 $\pm$ 1.78 & 1.34 $\pm$ 0.88 & 3.98 $\pm$ 1.95 \\
350 & 1.41 $\pm$ 0.78 & 4.44 $\pm$ 1.96 & 1.26 $\pm$ 0.72 & 3.92 $\pm$ 1.81 & 1.46 $\pm$ 1.01 & 3.94 $\pm$ 1.95 \\
400 & 1.41 $\pm$ 0.78 & 4.35 $\pm$ 1.94 & 1.22 $\pm$ 0.71 & 3.80 $\pm$ 1.81 & 1.36 $\pm$ 0.89 & 3.97 $\pm$ 1.95 \\
450 & 1.37 $\pm$ 0.78 & 4.29 $\pm$ 1.97 & 1.19 $\pm$ 0.71 & 3.73 $\pm$ 1.83 & 1.50 $\pm$ 1.06 & 3.94 $\pm$ 2.01 \\

\bottomrule
\end{tabular}\label{table:metrics2}
\end{table}


\section{DeepONet-accelerated inversion} \label{subsec: SurrogateEKI}

With the DeepONet surrogate forward model $\mathcal{F}_{s}(\bfu)$, and thus the surrogate parameter-to-measurements map $\mathcal{G}_{s}=\OO\circ \FF_{s}\circ \mathcal{P}$, EKI is used to infer the posterior distribution of $\bfu = (\bfu_{K,\phi},\,\mu,\,P_{I},\,\lambda,\,\beta)$. Recall that the posterior distribution of the reinforcement porosity and permeability can be obtained by pushing forward the posterior on $\bfu_{K,\phi}$ under $\mathcal{P}$. In inverse problems, naively replacing the true parameter-to-measurements map $\mathcal{G}$ with a surrogate $\GG_{s}$ can introduce bias and overconfidence in the inferred parameters. To address this, surrogate–model discrepancy is explicitly incorporated into the inversion rather than treating $\mathcal{G}_{s}$ as exact. Specifically, the enhanced modelling-error framework of  \citet{OfflineUQ} is adopted, in which the discrepancy between the high-fidelity forward map and $\mathcal{G}_{s}$ is characterised offline and propagated through the likelihood during EKI.

For consistency with the inversion performed using the full forward model, EKI is also employed in the surrogate-accelerated setting. It is emphasised, however, that the DeepONet surrogate is not intrinsically tied to EKI and could, in principle, be combined with alternative Bayesian inversion methodologies. Despite the differentiability of the DeepONet architecture itself, the overall parameter-to-observable map remains non-differentiable due to the parametrisation of the permeability and porosity fields, which involves level-set representations and piecewise-defined random fields. Consequently, derivative-based sampling methods, such as gradient-informed or geometric MCMC techniques, are not directly applicable without additional smoothing or reformulation, making a derivative-free ensemble-based approach a natural choice in the present setting.

When employing the surrogate within the inversion, the observation model of Eq.~\eqref{eq: inverse_problem_unparameterised} is rewritten in terms of the surrogate:  
\begin{align}
    \mathbf{d} =  \mathcal{G}_{s}(\bfu)  + \varepsilon + \eta, \qquad \eta \sim N(0,\Gamma),
\end{align}
where $\varepsilon := \GG(\bfu) - \GG_{s}(\bfu)$ denotes the surrogate error. The enhanced modelling-error framework works under the assumption that this surrogate error is independent of the inputs $\bfu$, and is normally distributed, i.e.\ $\varepsilon \sim N(\bar{\varepsilon},\Sigma)$. By defining $\zeta := \varepsilon - \bar{\varepsilon} + \eta$, the observation model can be equivalently expressed as  
\begin{align}
    \mathbf{d} - \bar{\varepsilon} = \GG_{s}(\bfu) + \zeta, \qquad \zeta \sim N(0,\Gamma+\Sigma). \label{eq: surr_aug_inv}
\end{align}
In practice, $\bar{\varepsilon}$ and $\Sigma$ are estimated using the sample mean and covariance of surrogate errors computed over a sufficiently large collection of test cases.  

In the framework proposed here, the unseen testing data are used to compute surrogate errors and their corresponding statistics at the measurement points for each sensor configuration. Since the testing and training data are drawn from the same prior distribution of inputs, these errors can be readily obtained by evaluating  
\[
\GG_{s}(\bfu^{(j)}) = \OO \circ \mathcal{F}_{s} \circ \mathcal{P}(\bfu^{(j)})
\]
for each test sample and comparing with the corresponding high-fidelity outputs, generating the set $\{\GG(\bfu^{(j)})-\GG_{s}(\bfu^{(j)})\}_{j=1}^{N_{\text{test}}}$. In other words, error statistics are built directly from test-set predictions without requiring additional forward simulations.  


The surrogate-augmented inverse formulation in Eq.~\eqref{eq: surr_aug_inv} is solved in exactly the same way as the original inverse problem described in Section \ref{subsec:BayesApproach}, using Algorithm~\ref{al: EKI} in \ref{app: EKI}. The only changes are the use of surrogate evaluations $\GG_s$, the shifted data $\mathbf{d} - \bar{\varepsilon}$, and the inflated covariance $\Gamma + \Sigma$. The inclusion of $\Sigma$ effectively down-weights information from sensor locations, where the surrogate is less accurate, thereby mitigating bias and overconfidence in the inferred posterior.

\subsection{Synthetic experiments with DeepONet-accelerated EKI} \label{sec: Virtual}
EKI is applied using the surrogate with $N_{\text{out}}=400$ and varying training set sizes. The same synthetic data and initial ensemble of $J=5000$ from Section~\ref{sec:virtual_EKI} are employed. To ensure seamless integration with the surrogate, both DeepONet and EKI are implemented in PyTorch. Unlike the full-model inversions (which exploited CPU parallelism across $\sim$90 cores), all surrogate-based inversions are executed on a single NVIDIA L40 GPU (48~GB); parallelism arises from GPU tensor operations rather than multi-core CPUs. Although the computational bottleneck is GPU-bound, 20 CPU cores were requested solely to provide sufficient host memory under the system’s \texttt{mem-per-cpu} policy (3.85~GB per core, i.e., $\approx 77$~GB total). This headroom was required for data loading and assembling ensemble/observation covariance objects; the extra CPU cores were not used for algorithmic speed-up.

\begin{figure*}
    \centering
    \includegraphics[width=0.9\linewidth]{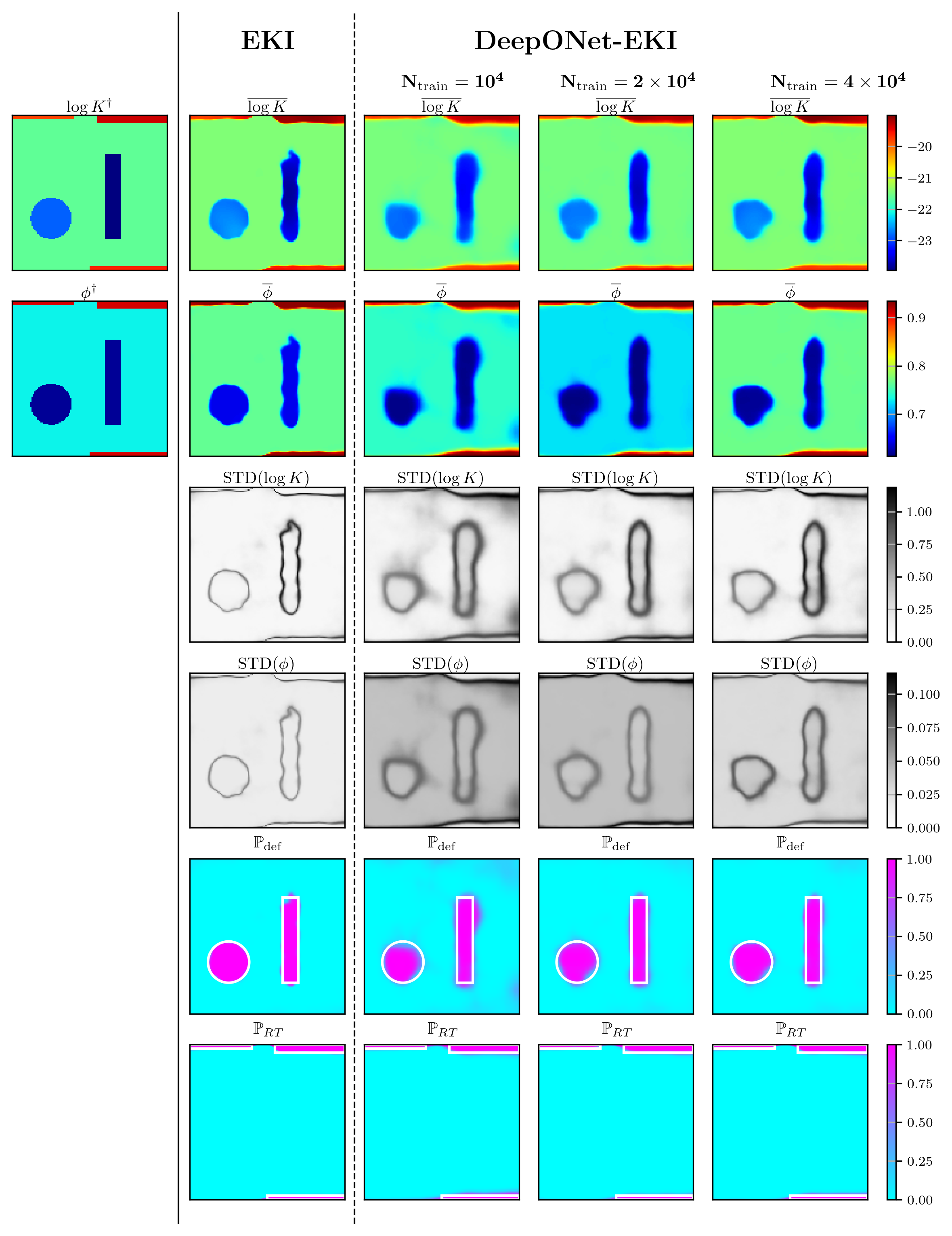}
    \caption{Results for the $M=100$ sensor configuration with $J=5000$. First column: ground–truth log-permeability, $\log K^{\dagger}$, and porosity, $\phi^{\dagger}$. Second column: posterior statistics obtained with full-model EKI (see Fig.~\ref{fig:full_EKI1}). Third to fifth columns: posterior statistics obtained with DeepONet-EKI for training sizes $N_{\text{train}}=10^4$, $N_{\text{train}}=2\times 10^4$, and $N_{\text{train}}=4\times 10^4$, respectively. From top to bottom: mean of log-permeability, $\overline{\log K}$, mean of porosity, $\overline{\phi}$, standard deviation of log-permeability, $\mathrm{STD}(\log K)$, standard deviation of porosity, $\mathrm{STD}(\phi)$, probability of central defects, $\mathbb{P}_{\mathrm{def}}$, and probability of RT, $\mathbb{P}_{\mathrm{RT}}$. The geometry of the true central defects is shown in white in the fifth row, while the true RT geometry is shown in the bottom row.}
    \label{fig:example_test1}
\end{figure*}

\begin{figure*}
    \centering
    \includegraphics[width=1.0\linewidth]{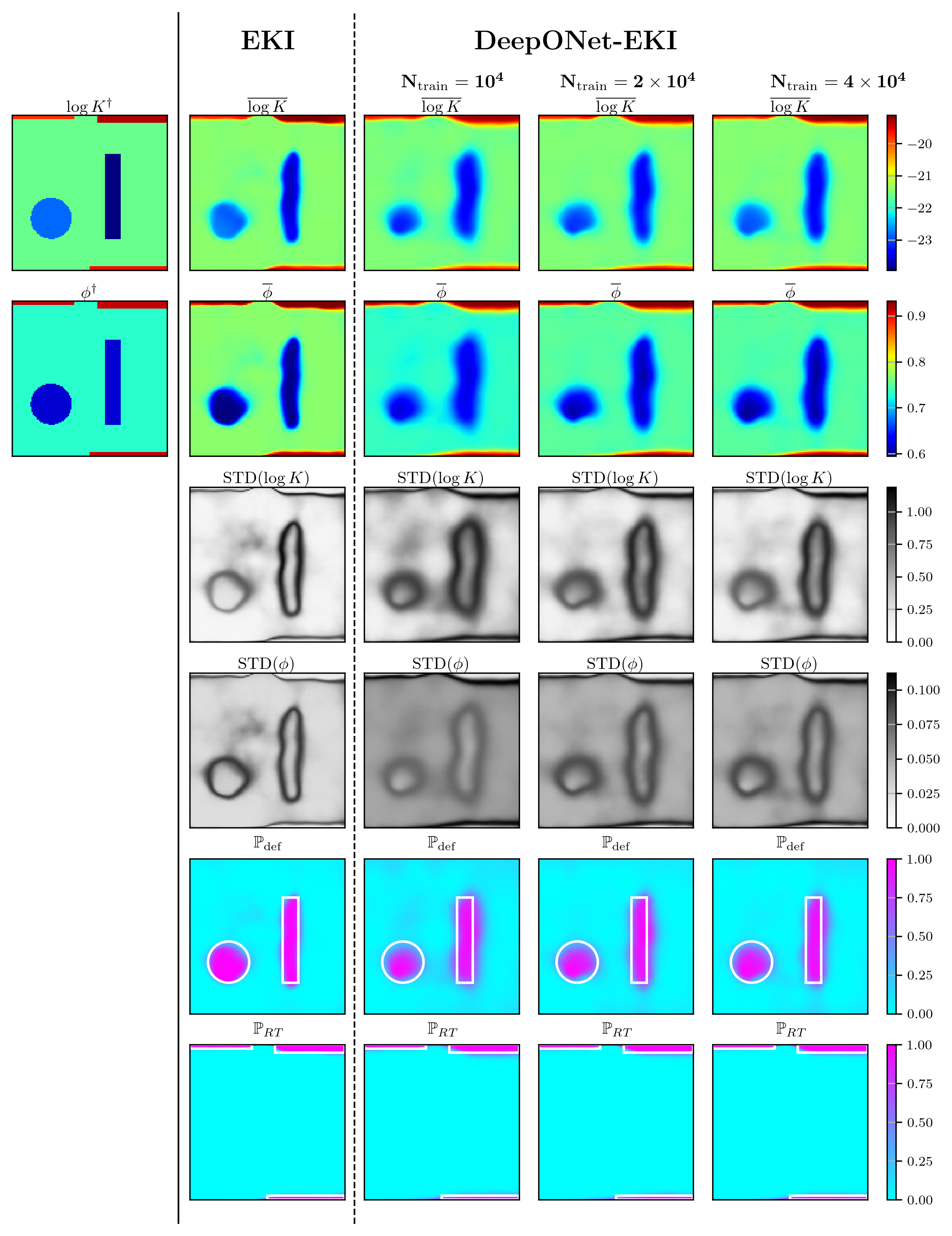}
    \caption{Results for the $M=23$ sensor configuration with $J=5000$. The columns are as described in Fig.~\ref{fig:example_test1}, except for the second column, which presents the posterior statistics obtained from the full-model EKI shown in Fig.~\ref{fig:full_EKI2}.}
    \label{fig:example_test2}
\end{figure*}

As in Section~\ref{subsec:BayesApproach}, posterior statistics are computed for all inferred variables. Fig.~\ref{fig:example_test1} shows the results of the DeepONet-accelerated EKI inversion for the sensor configuration with $M=100$, while Fig.~\ref{fig:example_test2} displays results for $M=23$ across different surrogate training sizes. For reference, the posterior results from the full-model EKI with $J=5000$ are also included.  The results show that log-permeability and porosity can be inferred very accurately with the DeepONet surrogate when a sufficiently large training set is employed. This effect is most pronounced for the case with fewer sensors ($M=23$): here the inversion problem is more ill-posed, so surrogate accuracy plays a greater role, and larger training sets lead to visibly improved reconstructions. For $M=100$, even $N_{\text{train}}=10{,}000$ produced satisfactory reconstructions. In all cases, race tracking is well recovered, and central defects are identified within the regions of highest posterior probability.  

Despite the accuracy of the surrogate, the inherent discrepancy reduces overall inversion accuracy compared to the full model. Moreover, the Gaussian assumption underlying the enhanced modelling–error framework is imperfect, since surrogate errors are not exactly Gaussian. This mismatch can lead to slight underestimation of uncertainty and reduced reconstruction quality, especially for filling-factor predictions.  

\begin{figure*}
    \centering
    \includegraphics[width=1.0\linewidth]{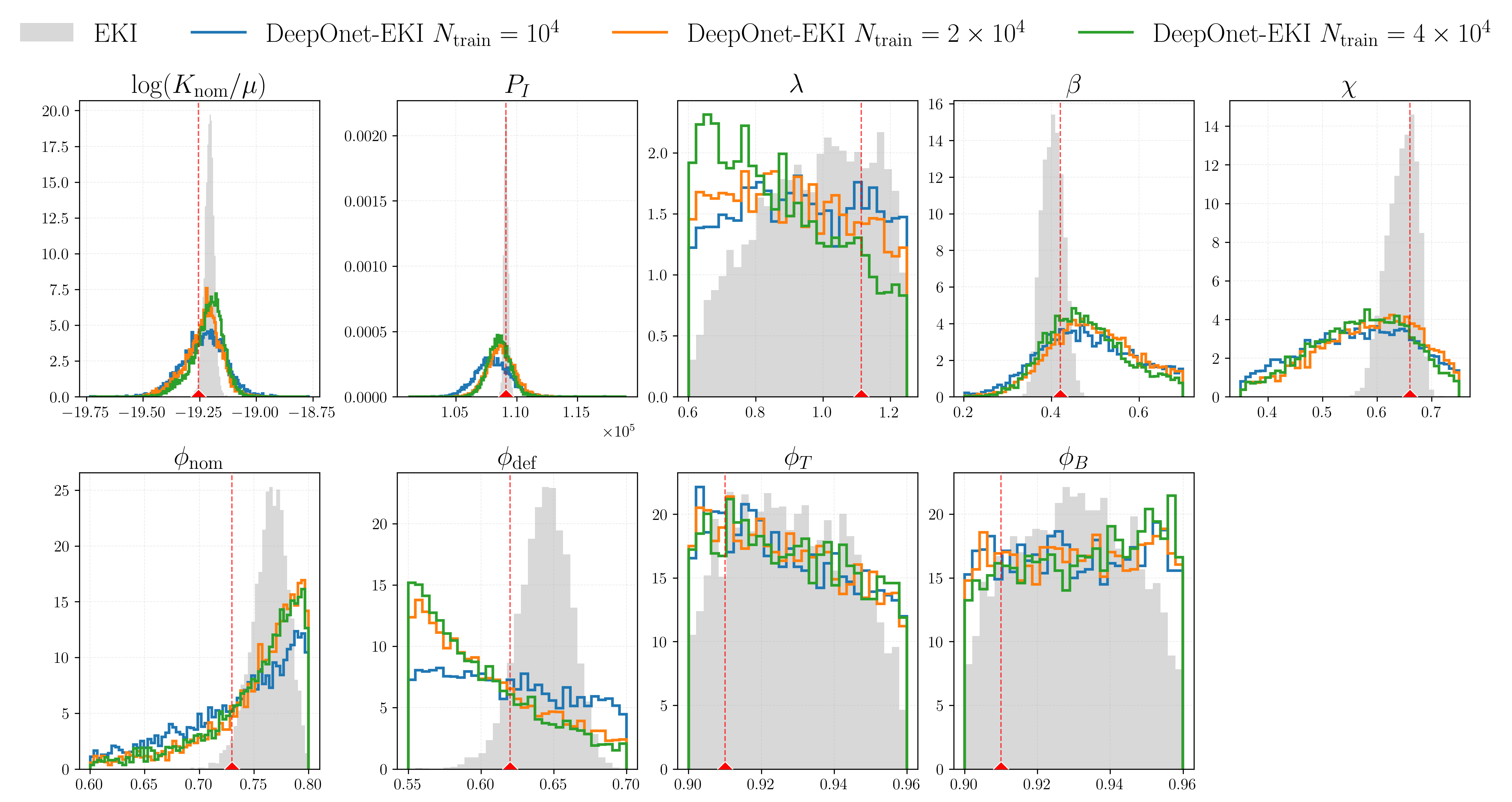}
    \caption{Results for $M=100$. Histograms showing marginal posterior distributions for the scalar parameters, using full-model EKI and DeepONet-EKI with training dataset sizes $N_{\text{train}}=10^4$, $N_{\text{train}}=2\times 10^4$, and $N_{\text{train}}=4\times 10^4$. True values are shown in red.}
    \label{fig:example_test3}
\end{figure*}

\begin{figure*}
    \centering
    \includegraphics[width=1.0\linewidth]{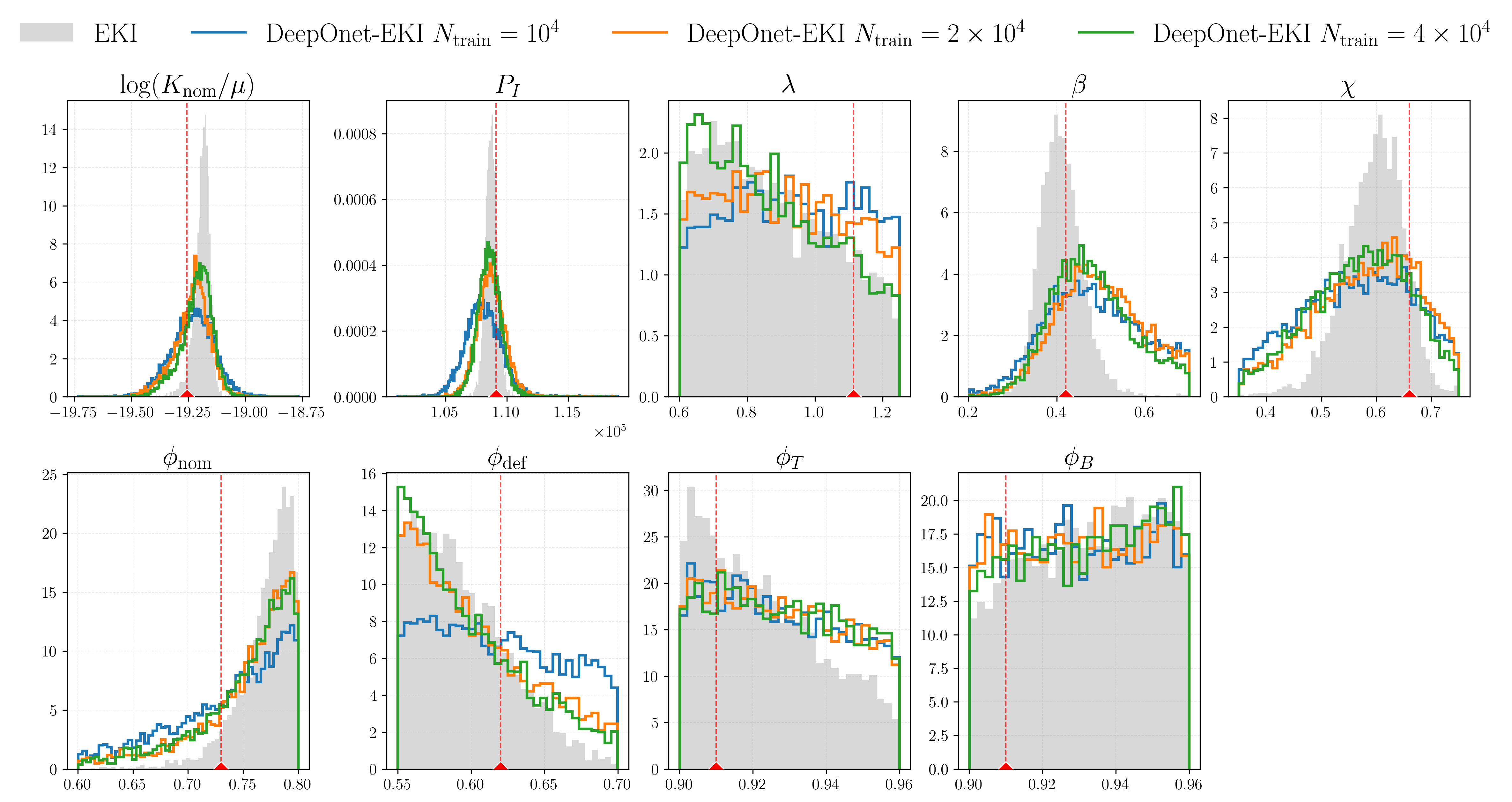}
    \caption{Results for $M=23$. See description in Fig.~\ref{fig:example_test3}.}
    \label{fig:example_test4}
\end{figure*}

\begin{figure*}
    \centering
    \includegraphics[width=1.0\linewidth]{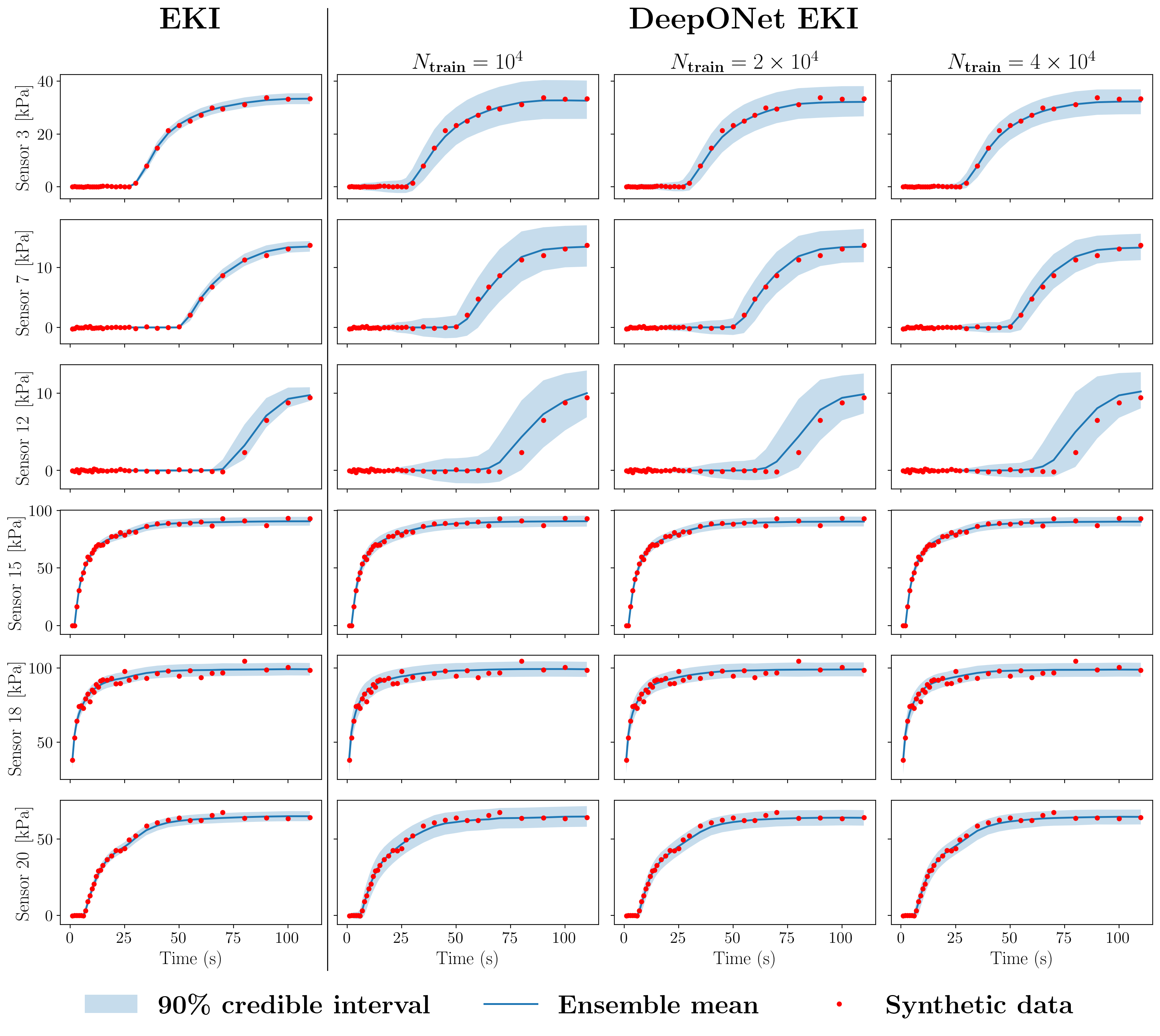}
    \caption{Results for $M=23$ sensors. Synthetic data at the sensor locations are shown in red. Posterior mean predictions and associated credible regions (blue) are obtained from the DeepONet surrogate. For comparison, the corresponding quantities obtained using the full EKI approach are shown in the left panels.}    \label{fig:example_test5}
\end{figure*}

Histograms of the inferred scalar variables are shown in Figs.~\ref{fig:example_test3} to \ref{fig:example_test4}, while posterior predictive distributions are reported in Fig.~\ref{fig:example_test5}. The effect of training data size is also evident in the scalar parameters: with larger training sets, the surrogate-based posteriors align more closely with those obtained from full-model EKI. It is also to be noted that the surrogate predictive distributions exhibit larger uncertainties compared to the full-model EKI. While this added uncertainty is in some sense desirable as it prevents overconfidence in the surrogate, it can also be detrimental to the overall inversion by inflating posterior variance.

The overall inversion performance is summarised in Table~\ref{table:deep-EKI}. Replacing the high-fidelity forward model with the DeepONet surrogate results in speed-ups of approximately two orders of magnitude compared with full-model EKI. For $M=100$ sensors, the DeepONet-accelerated inversion achieves speed-up factors between approximately $100$ and $200$ for $M=100$ sensors, and between $100$ and $150$ for $M=23$ sensors, depending on the training-set size. In the most realistic configurations, the full inversion is completed within $20$ seconds to $30$~seconds of wall-clock time, compared with several hours for the full-model EKI. The increase in runtime with larger $M$ is expected: a greater number of sensors increases the dimensionality of the observation space, which in turn enlarges the tensors involved in surrogate evaluation and covariance updates within EKI. This leads to higher memory and computational demands, even though each individual evaluation of the surrogate forward map $\mathcal{G}_s$ remains inexpensive.

Although generating the training dataset and training the DeepONet model incur a substantial offline computational cost (which may, in principle, be comparable to several full EKI inversions), this cost is incurred only once. Once trained, the surrogate enables near–real-time Bayesian inversion, making it suitable for time-critical applications such as adaptive process monitoring and control. Furthermore, the same trained DeepONet model can be reused across different sensor configurations, as demonstrated here for two distinct layouts, without retraining. In an engineering process scenario, the offline training effort is therefore amortised as additional experiments and monitoring campaigns are performed.


In summary, although the surrogate achieves less than $1.2\%$ average error in pressure metrics, the point-wise errors can be large and non-Gaussian. As a result, when the surrogate is used within the inversion, the accuracy of the reconstructions degrades in the sense that posterior uncertainties are inflated. Nevertheless, defects (including race-tracking) are still accurately identified within seconds on a single GPU, which highlights the practical usefulness of the approach. While the virtual experiments considered here represent an idealised best-case setting, in which the only source of error arises from the surrogate approximation, additional uncertainties are unavoidable in realistic scenarios: even with a near-perfect surrogate, the underlying forward model may not capture the true physics due to unaccounted source terms, discretisation error in the simulator, or incomplete physical modelling.

\begin{table}[ht]
\centering
\caption{EKI performance, with $J=5000$, $N_{\text{out}}=400$ for two sensor layouts.}
\label{table:deep-EKI}
\begin{threeparttable}
\begin{tabular}{@{}l cc cc cc@{}}
\toprule
& \multicolumn{2}{c}{Resources} & \multicolumn{2}{c}{$M=100$ sensors} & \multicolumn{2}{c}{$M=23$ sensors} \\
\cmidrule(lr){2-3}\cmidrule(lr){4-5}\cmidrule(lr){6-7}
Method & CPU cores & GPUs & Time (s) & $N_{\text{iter}}$ & Time (s) & $N_{\text{iter}}$ \\
\midrule
DeepONet-EKI, $N_{\text{train}}=10^4$        & 20 & 1 & 51.976  & 4 & 17.753 & 5 \\
DeepONet-EKI, $N_{\text{train}}=2\times 10^4$& 20 & 1 & 72.840 & 6 & 23.652 & 7 \\
DeepONet-EKI, $N_{\text{train}}=4\times 10^4$& 20 & 1 & 83.736 & 7 & 26.664 & 8 \\
EKI (full model)                              & 90 & 0 & $9.614\times 10^3$ & 12 & $1.278\times 10^4$ & 16 \\

\bottomrule
\end{tabular}
\end{threeparttable}
\end{table}


\subsection{Lab experiments with DeepONet-accelerated EKI} \label{sec: Lab}


The experimental setup here is the same as in the previous work \citep{MC}, where further details can be found. For the laboratory experiments, an injection tool 
consisting of a steel base plate and a transparent PMMA cover plate of 90~mm thickness was used. The tool cavity accommodates square specimens with dimensions of $300~ mm \times 300~ mm$. To create unidirectional flow-driving pressure gradients, galleries are machined in the base of the fixture along two opposing edges, serving as the linear inlet and outlet, respectively. A total of 23 pressure sensors were embedded within the tool, supplemented by an additional sensor at the inlet to monitor the injection pressure. All sensors were connected to a data acquisition unit, recording fluid pressures at a rate of 10~Hz. In parallel, a video camera was used to capture the progression of the flow front.


All experiments employed a continuous random glass-fibre reinforcement with an areal density of $(210 \pm 15)$~g/m$^2$. The reference lay-up contained seven plies of this material. The overall fibre volume fraction was controlled using a spacer frame positioned between the base and the cover of the mould tool; the frame and sealing elements provided a nominal cavity height of 2.1~mm. Measurements of mould deflection with the reinforcement inserted showed displacements below 0.1~mm for the seven-ply baseline, leading to a porosity of approximately 0.73. 

To generate inclusions, circular patches of 80~mm diameter were cut from the same reinforcement. Several of these patches were stacked at different through-thickness positions within a lay-up to modify the local porosity and permeability. Adding three extra plies locally reduced the porosity to approximately 0.62. RT channels were introduced by removing narrow strips of reinforcement at prescribed locations along the specimen edges, and different defect geometries were investigated.  

As the test fluid, a synthetic oil was used. Its viscosity was characterised across a range of temperatures using a Brookfield viscometer, and the viscosity was determined from the temperature–viscosity calibration after measuring the fluid temperature directly before each injection. Experiments were conducted at an inlet (gauge) pressure close to $10^{5}$~Pa. The inlet pressure (after sufficient time) and resin viscosity associated with each run are summarised in Table~\ref{tab:lab-details}. 

Four experimental configurations are considered here:  
\emph{Experiment~1 (no defects):} a reference case with no intentional variations in reinforcement properties;  
\emph{Experiment~2 (circular defect; no race tracking):} a specimen containing a centrally located low-porosity circular inclusion;   
\emph{Experiment~3 (race tracking only on top and bottom edges):} a specimen where 5~mm wide strips were removed along the edges in the second half of the domain to create RT channels;
\emph{Experiment~4 (circular defect with race tracking at the top edge):} a specimen combining a central circular defect with an additional 3~mm wide RT channel cut along the top edge.  

\subsubsection{Inversion outcomes}
DeepONet-EKI is applied to each of the recorded experimental datasets, assuming measurement noise of $2.5\%$ relative to the signal magnitude. The same initial ensemble ($J=5000$) as in the synthetic experiments (Section~\ref{sec:virtual_EKI}) is employed. For all real-data inversions, the DeepONet surrogate is used with latent dimension $N_{\text{out}}=400$ trained on the largest dataset ($N_{\text{train}}=40{,}000$). The DeepONet-acceleration inversion time for each experiment is shown in Table~\ref{table:deep-EKI2}, with each inversion requiring less than 42 s.

Although the reported wall-clock times are of the order of tens of seconds, it is important to note that these inversions are performed in an \emph{all-at-once} setting, in which the complete measurement time series is assimilated simultaneously after completion of the injection process. This choice is made for simplicity and to enable direct comparison with full-model EKI. In practice, however, the proposed framework naturally supports \emph{sequential} or online data assimilation, in which measurements are incorporated incrementally as they become available. In such a setting, the computational cost per assimilation step is significantly reduced, since only a small number of new observations are processed at each time instance. Consequently, when used in a sequential inversion mode, the DeepONet-accelerated EKI is expected to operate close to real time relative to the data acquisition rate, making it suitable for online monitoring and adaptive control during the injection process.

\begin{figure}
    \centering
    \includegraphics[width=\linewidth]{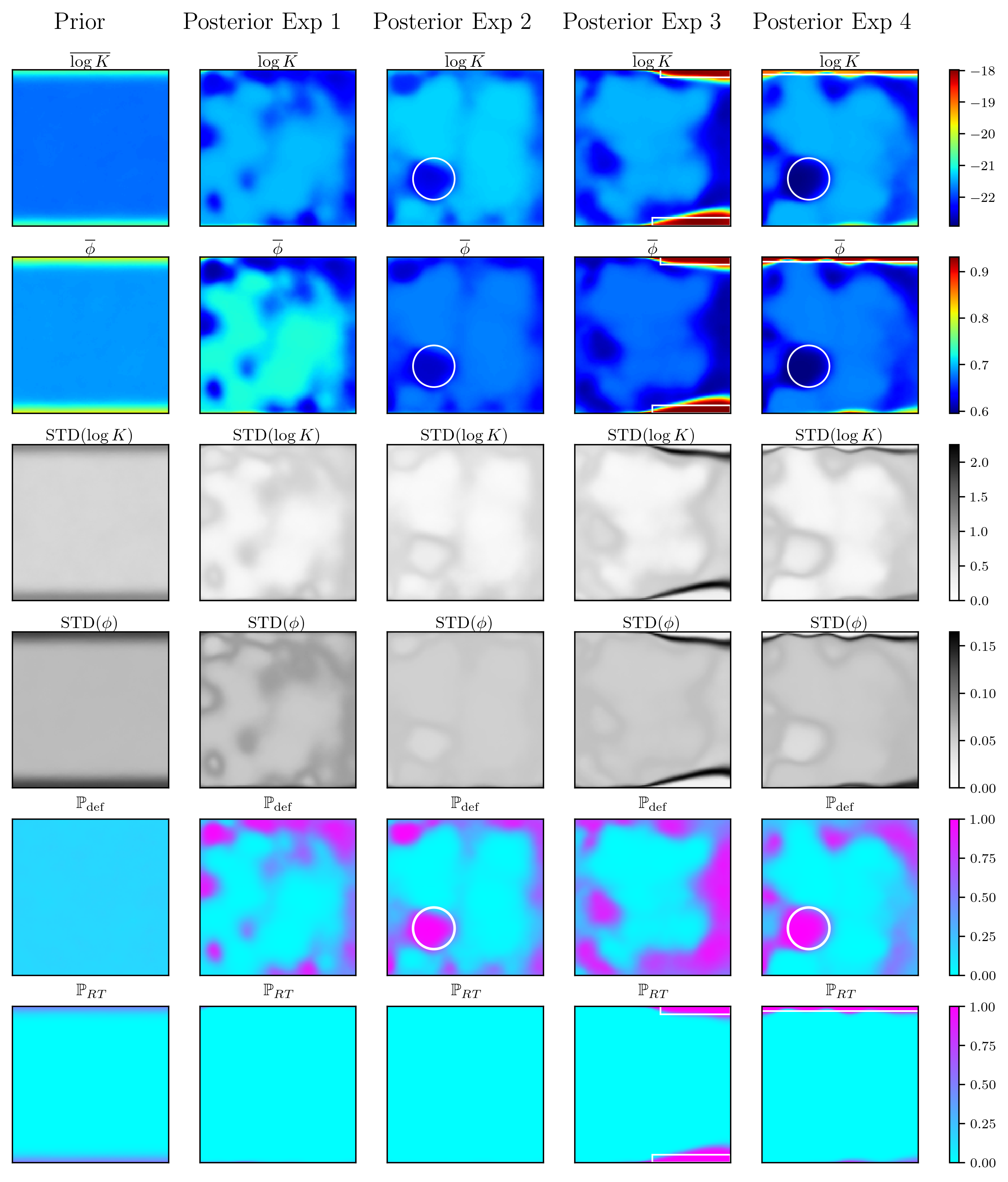}
    \caption{Results for the lab experiments with $J=5000$. First column: prior statistics. Second to fifth columns: posterior statistics obtained with DeepONet-EKI for experiments 1, 2, 3, and 4, respectively. From top to bottom: mean of log-permeability, $\overline{\log K}$, mean of porosity, $\overline{\phi}$, standard deviation of log-permeability, $\mathrm{STD}(\log K)$, standard deviation of porosity, $\mathrm{STD}(\phi)$, probability of central defects, $\mathbb{P}_{\mathrm{def}}$, and probability of RT, $\mathbb{P}_{\mathrm{RT}}$. The geometry of the intended central defects and RT regions (where applicable) are shown in white.}
    \label{fig:real1}
\end{figure}

\begin{figure}
    \centering
    \includegraphics[width=\linewidth]{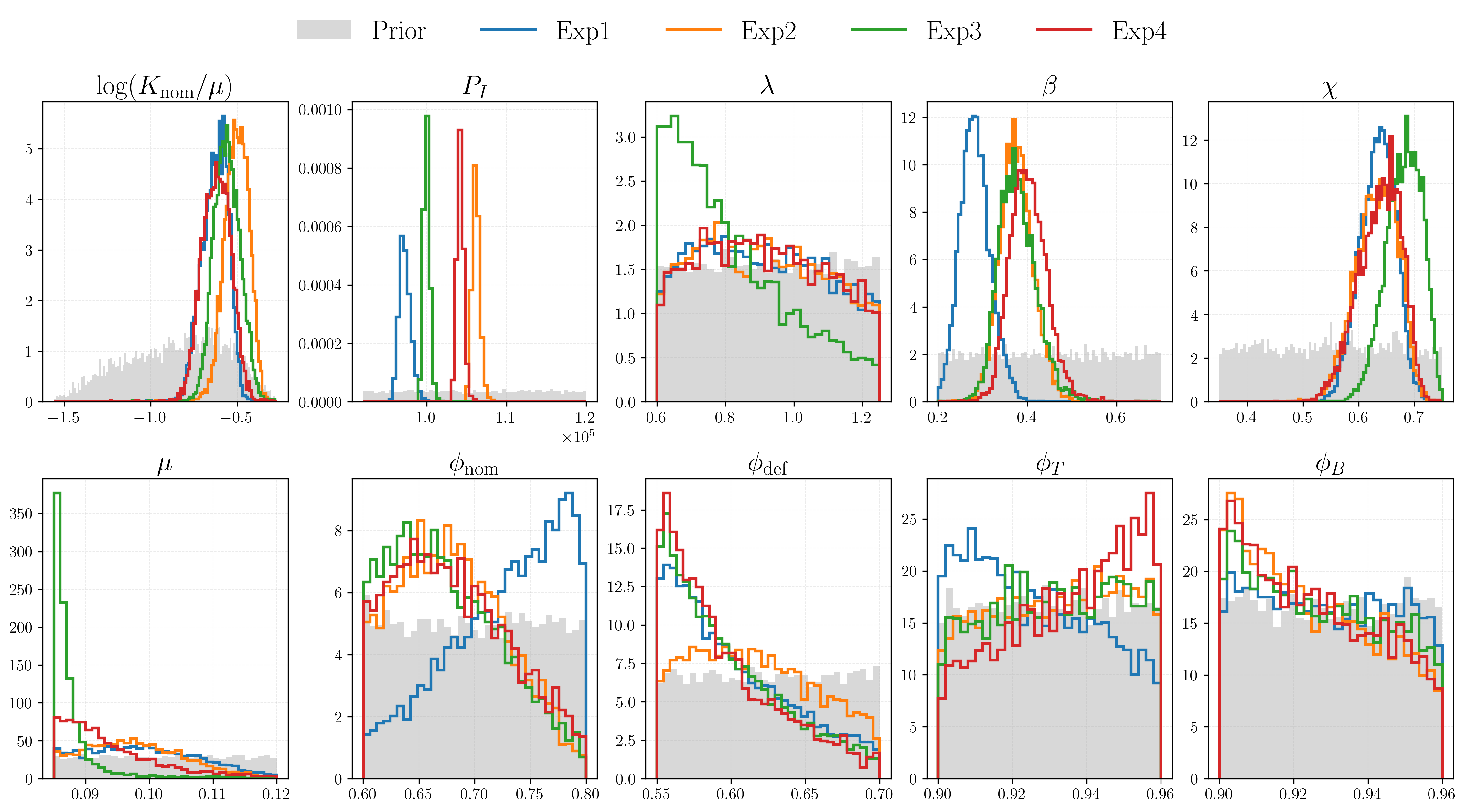}
    \caption{Histograms showing the marginal priors and DeepONet-EKI posteriors for the scalar parameters in each experiment.}
    \label{fig:real2}
\end{figure}

\begin{figure}
    \centering
    \includegraphics[width=\linewidth]{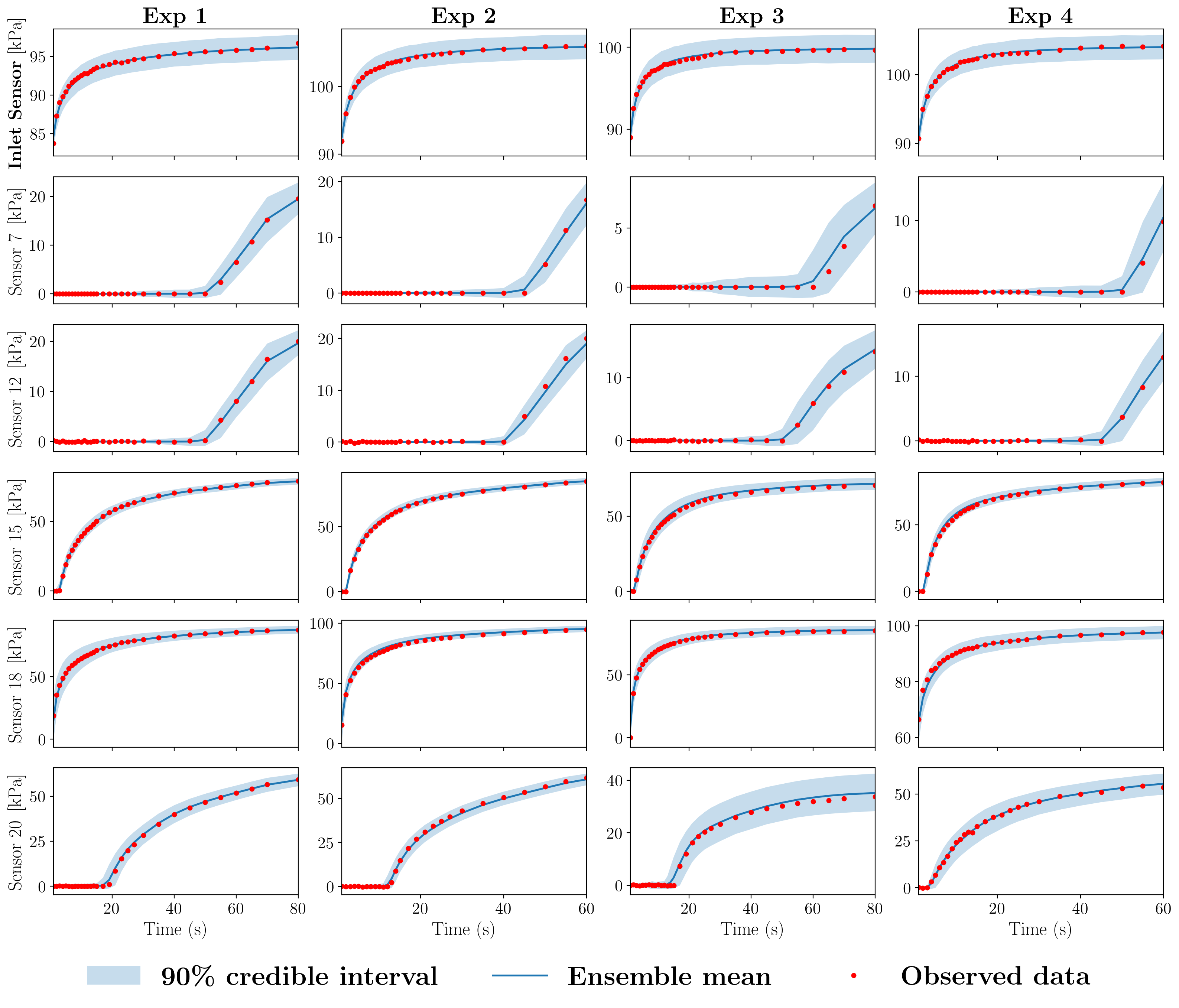}
    \caption{Top row: experimental data at an additional inlet sensor (not used in the inversion), along with mean and credible regions for $P_I(t)$, constructed by pushing DeepONet-EKI posterior samples of $(P_{I},\lambda,\beta,\chi)$ through the relation in Eq.~\eqref{eq:press}. Remaining rows: experimental data for each lab experiment at various sensors (red), along mean estimates and credible regions (blue), generated by the DeepONet surrogate using samples from the joint DeepONet-EKI posterior distributions.}
    \label{fig:real3}
\end{figure}

\begin{figure}[htbp]
    \centering
    \begin{subfigure}[b]{\textwidth}
        \centering
        \includegraphics[width=\textwidth]{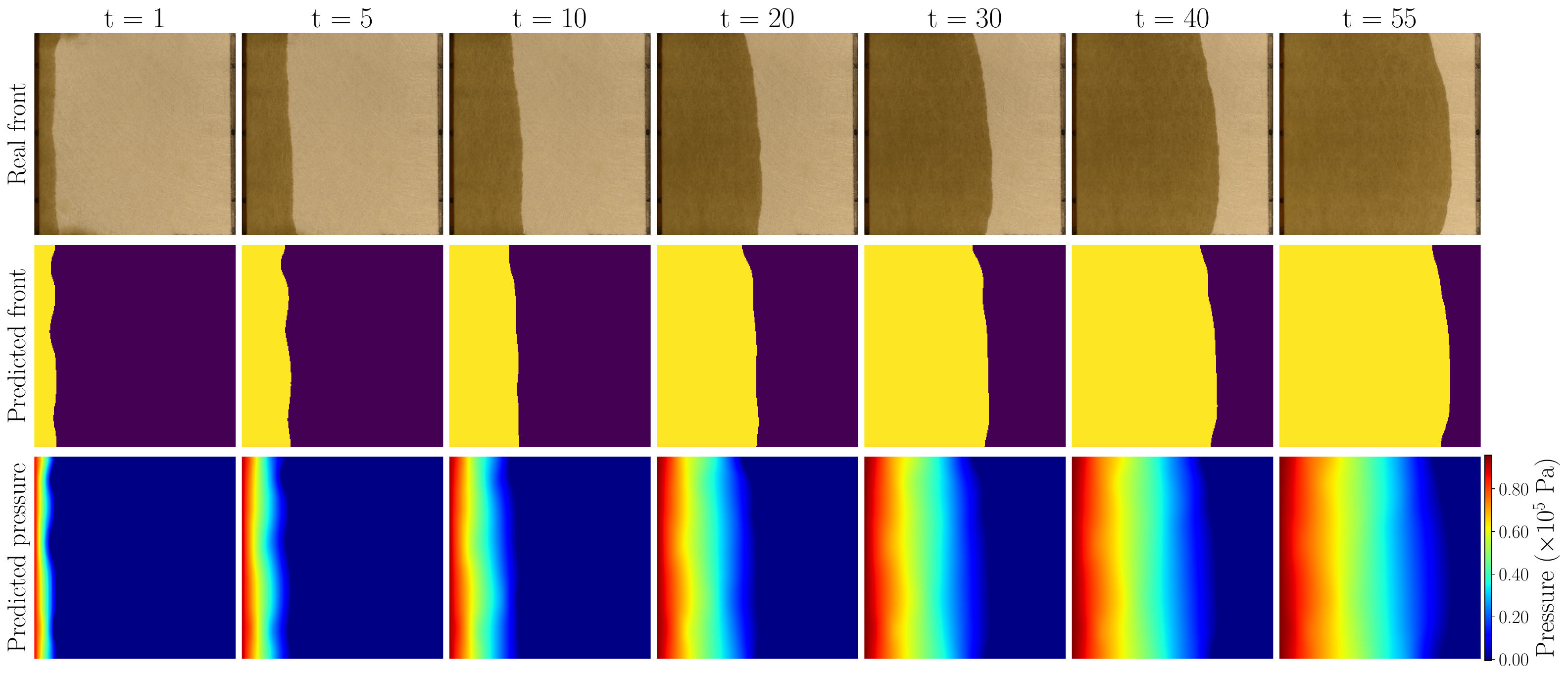}
        \caption{Experiment 1: no defect}
        \label{fig:exp1}
    \end{subfigure}
    \hfill
    \begin{subfigure}[b]{\textwidth}
        \centering
        \includegraphics[width=\textwidth]{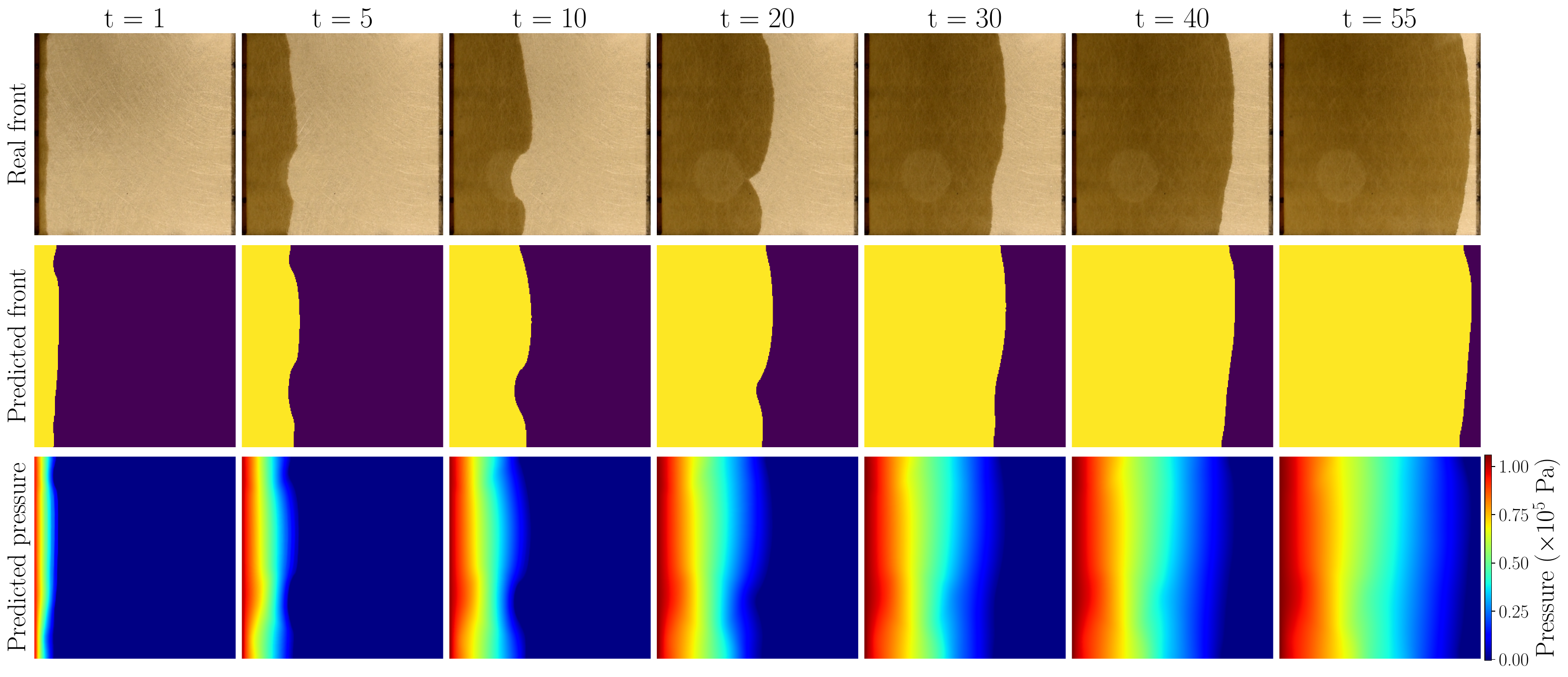}
        \caption{Experiment 2: circular defect}
        \label{fig:exp2}
    \end{subfigure}
    \caption{(Top of each) Images of the experiment at various times, (middle and bottom of each) filling factor and pressure attained by DeepONet surrogate using the posterior mean of inferred parameters as inputs.}
    \label{fig:real4}
\end{figure}

\begin{figure}[htbp]
    \centering
    \begin{subfigure}[b]{\textwidth}
        \centering
        \includegraphics[width=\textwidth]{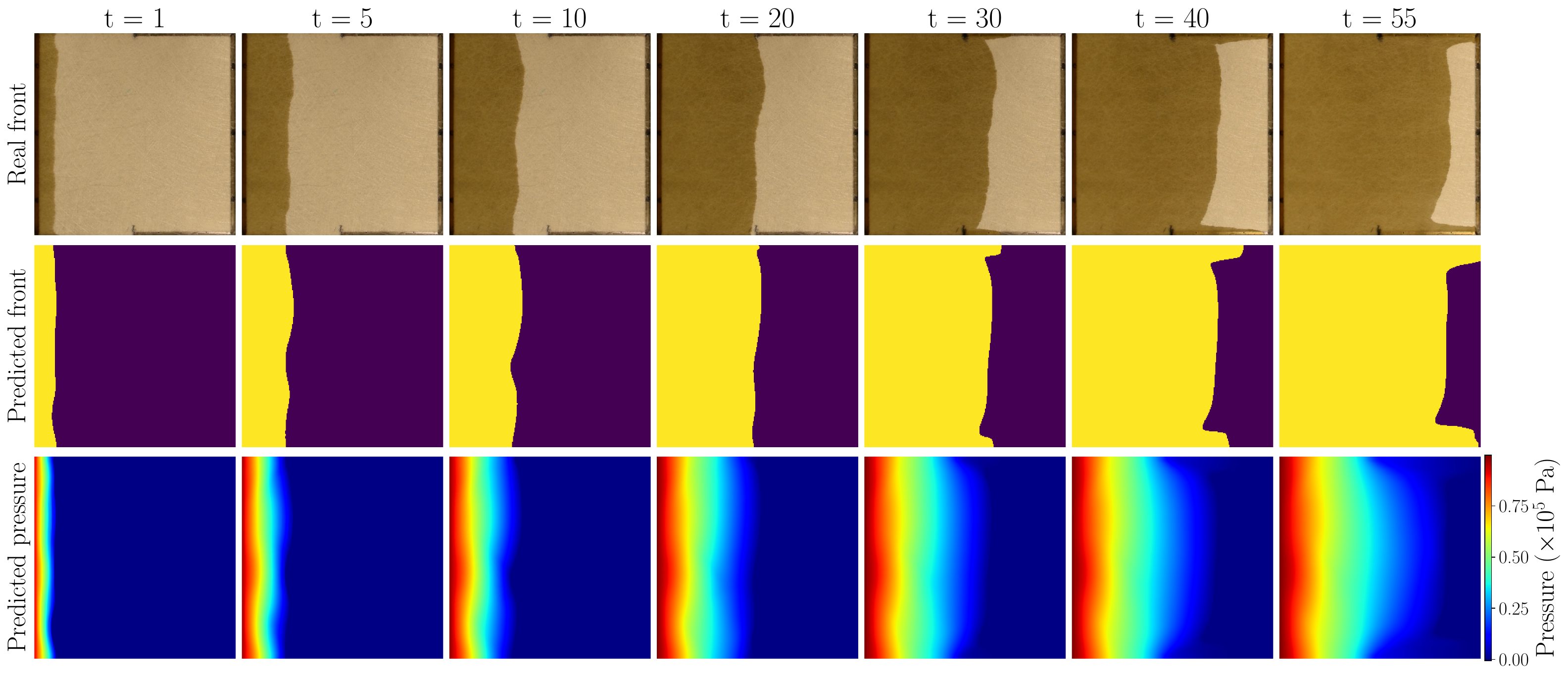}
        \caption{Experiment 3: race tracking only}
        \label{fig:exp3}
    \end{subfigure}
    \hfill
    \begin{subfigure}[b]{\textwidth}
        \centering
        \includegraphics[width=\textwidth]{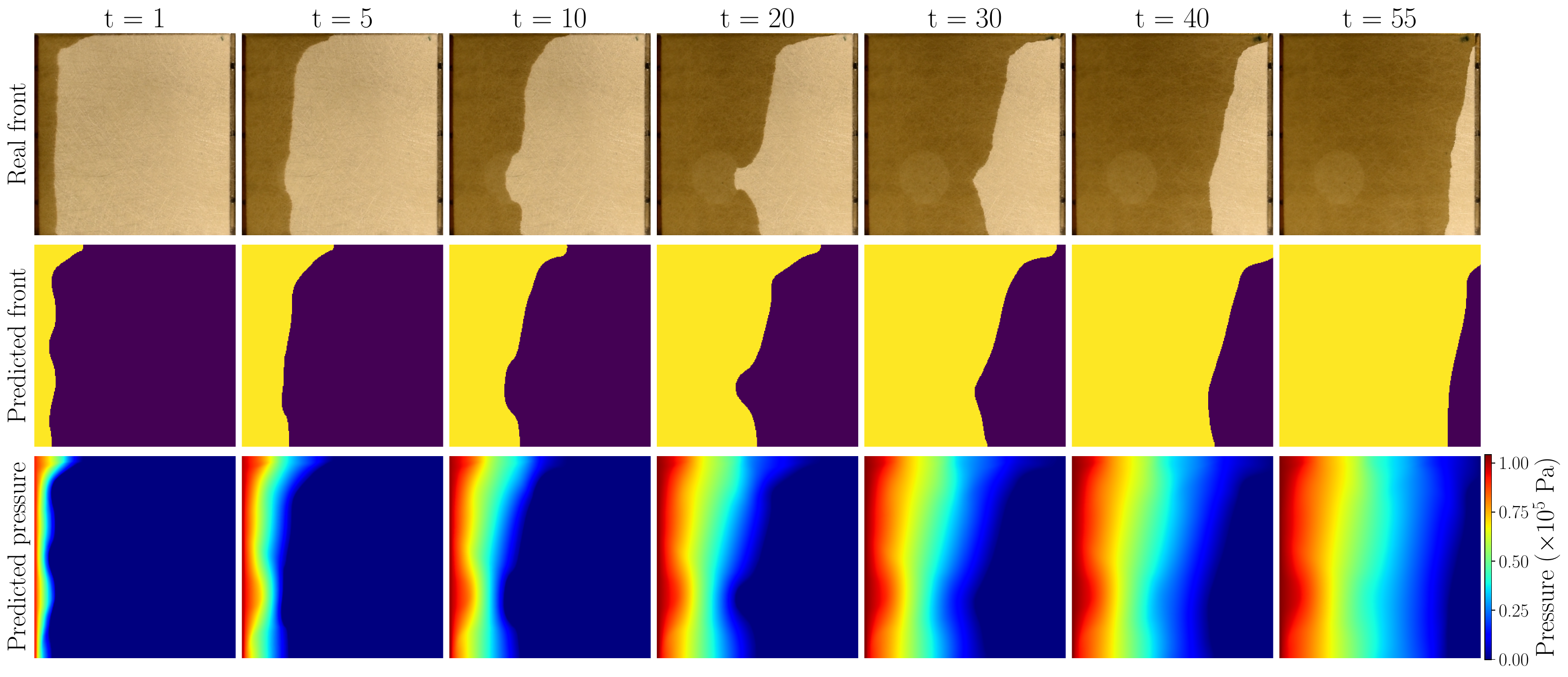}
        \caption{Experiment 4: circular defect with race tracking}
        \label{fig:exp4}
    \end{subfigure}

    \caption{(Top of each) Images of the experiment at various times, (middle and bottom of each) filling factor and pressure attained by DeepONet surrogate using the posterior mean of inferred parameters as inputs.}
    \label{fig:real5}
\end{figure}

Posterior statistics for log-permeability, porosity, and defect probabilities for each experiment are shown in Fig.~\ref{fig:real1}, with prior ensemble statistics included for reference. Histograms of the inferred scalar variables are given in Fig.~\ref{fig:real2}, and posterior predictions at sensor locations are presented in Fig.~\ref{fig:real3}. For the latter, it is worth noticing that the posterior predictions are computed using the DeepONet surrogate with inputs obtained via the DeepONet-EKI framework. The top panel of Fig.~\ref{fig:real3} corresponds to the inlet sensor, where predictions are obtained via the inlet pressure model given by Eq.~\eqref{eq:press} using the ensemble of inferred parameters $(P_{I},\lambda,\beta,\chi)$. To further validate the approach, the full pressure and filling-factor fields are computed using the posterior ensemble means as input to the DeepONet surrogate. These predictions are evaluated on a $256 \times 256$ grid, which is finer than the training mesh, and selected snapshots are displayed in Figs.~\ref{fig:real4} to \ref{fig:real5}.

Fig.~\ref{fig:real1} shows that, although the prior assigns some probability to RT in Experiment~1 (no defects) and Experiment~2 (circular defect without race tracking), the posterior probability of race tracking is nearly zero across the domain. However, in Experiment~1, the posterior identifies spurious regions of high defect probability near the edges, and even the centre of the specimen, despite the absence of intentional defects. This suggests that the reinforcement itself is not perfectly homogeneous. Indeed, Fig.~\ref{fig:real4}(a) shows that the experimental fluid front is not straight: at certain times, it lags behind near the edges. The surrogate reproduces this behaviour by assigning lower permeability and porosity in these regions. Similarly, for Experiment~2, the circular defect is correctly identified, but additional low-permeability regions at the top and bottom are again inferred (Fig.~\ref{fig:real4}(b)), compensating for observed lag in the flow front. For Experiments~3 and~4, the posterior distributions clearly capture the intended RT channels with high probability, and in Experiment~4, the central defect is also accurately localized. Yet, in both cases, additional unintended regions of reduced permeability and porosity are detected. Inspection of Figs.~\ref{fig:real4}(c) and (d) suggests that these are again associated with local lagging of the experimental front. In this sense, the inversion interprets delayed propagation as defects, even if they were not manufactured intentionally.

It is important to emphasise, however, that the unintended ``defects'' identified in Experiments~1 to 4 may not correspond to true material flaws. Plausible explanations could be a small amount of deflection of the transparent lid of the injection tool, which would result in increased porosity and faster fluid propagation along the centre line of the tool, or a local decrease in porosity along the reinforcement edges, resulting from the reinforcement being forced into the tool cavity and the occurrence of local fibre rearrangement.

Together, Fig.~\ref{fig:real2} and Table~\ref{tab:post_scalars_real} demonstrate that the scalar parameters are estimated effectively using DeepONet-EKI. Viscosity and asymptotic inlet pressure, in particular, correspond well with the estimated values outlined in Table~\ref{tab:lab-details}, and the porosity estimates are close to the nominal values provided at the beginning of this section. It should also be noted that the distributions of porosity of the RT channels stay close to the priors. This is consistent with the earlier observations. Figure ~\ref{fig:real3} confirms that posterior predictions of sensor measurements remain consistent with the experimental data: all observations fall within the credible intervals of the surrogate-accelerated inversion. In particular, the inlet sensor readings are well reproduced by the inlet pressure model, validating that the inferred parameters $(P_{I},\lambda,\beta,\chi)$ effectively calibrate the model.

\begin{table}[ht]
\centering
\caption{EKI performance on lab experiments ($J=5000$, $N_{\text{out}}=400$).}
\label{table:deep-EKI2}
\begin{threeparttable}
\begin{tabular}{@{}l c c c c@{}}
\toprule
Experiment & CPU cores & GPUs & Time (s) & $N_{\text{iter}}$ \\
\midrule
Experiment 1 & 20 & 1 &  41.894 & 14 \\  
Experiment 2 & 20 & 1 &  38.862  & 14 \\
Experiment 3 & 20 & 1 & 33.602  & 11 \\
Experiment 4 & 20 & 1 & 28.547   & 10 \\
\bottomrule
\end{tabular}
\begin{tablenotes}\footnotesize
\item \emph{Resource note:} Surrogate-based runs are GPU-bound on a single NVIDIA L40 (48\,GB). 
Twenty CPU cores were allocated to satisfy the cluster’s \texttt{mem-per-cpu} policy (host-memory provisioning), not for algorithmic speedup.
\end{tablenotes}
\end{threeparttable}
\end{table}

\begin{table}[!hb]
    \centering
    \caption{Setup for each lab experiment.}
    \begin{tabular}{c|c|c|l}
          Experiment & $P_I$ (Pa) & $\mu$ (Pa$\cdot$s)& Defect type\\
         \hline
         1 & $96066$ & $0.1088$ & None\\
         2 & $106050$ & $0.0922$ & Central (circular inclusion)\\
         3 & $97633$ & $0.1088$ & RT only (top and bottom)\\
         4 & $104410$ & $0.0922$ & RT (top) and central (circular inclusion)
    \end{tabular}
    \label{tab:lab-details}
\end{table}

\begin{table}[ht]
\centering
\caption{Posterior ensemble means and standard deviations ($N=5000$) for scalar parameters.}
\label{tab:post_scalars_real}
\begin{tabular}{@{}lcccc@{}}
\toprule
Parameter & Experiment 1 & Experiment 2 & Experiment 3 & Experiment 4 \\
\midrule
$K_{\rm nom}$ ($10^{-10}m^2$) & 5.294 $\pm$ 0.505 & 5.913 $\pm$ 0.403 & 5.045 $\pm$ 0.448 & 5.067 $\pm$ 0.520 \\
$\phi_{\rm nom}$ & 0.727 $\pm$ 0.051 & 0.681 $\pm$ 0.047 & 0.676 $\pm$ 0.049 & 0.681 $\pm$ 0.050 \\
$\phi_{\rm def}$ & 0.602 $\pm$ 0.039 & 0.616 $\pm$ 0.040 & 0.598 $\pm$ 0.037 & 0.595 $\pm$ 0.037 \\
$\phi_T$ & 0.926 $\pm$ 0.017 & 0.931 $\pm$ 0.017 & 0.931 $\pm$ 0.017 & 0.935 $\pm$ 0.017 \\
$\phi_B$ & 0.929 $\pm$ 0.017 & 0.925 $\pm$ 0.017 & 0.927 $\pm$ 0.017 & 0.926 $\pm$ 0.017 \\
$\mu$ ($10^{-2}$Pa.s) & 9.903 $\pm$ 0.849 & 9.781 $\pm$ 0.757 & 8.796 $\pm$ 0.476 & 9.393 $\pm$ 0.749 \\
$P_I$ (kPa)& 97.254 $\pm$ 0.728 & 106.186 $\pm$ 0.505 & 99.982 $\pm$ 0.394 & 104.223 $\pm$ 0.452 \\
\bottomrule
\end{tabular}
\end{table}

\section{Conclusions}\label{sec: Conclusions}
This work demonstrates that the EKI algorithm is capable of simultaneously reconstructing spatially varying and discontinuous permeability and porosity fields in porous media, together with scalar process variables such as viscosity and time-dependent boundary conditions, including the inlet pressure. The spatially varying fields are parametrised using Gaussian random fields, enabling the joint characterisation of both central defects and race-tracking regions. Although EKI with a relatively small ensemble size (e.g., $J=500$) yields reasonably accurate reconstructions within 13 minutes to 19 minutes using 90 cores of an HPC cluster, this computational cost remains suboptimal for real-time digital twinning applications in more practical settings using standard computing resources.

A DeepONet surrogate model was trained to alleviate the computational bottleneck associated with repeated evaluations of the forward operator. The surrogate was designed to approximate the nonlinear mapping from permeability and porosity fields, together with scalar parameters describing fluid viscosity and inlet pressure, to the resulting pressure and filling-factor fields. For a training dataset of size $N_{\text{train}} = 4\times 10^{4}$ and a latent output dimensionality of $N_{\text{out}} = 400$, the surrogate achieved mean relative errors of $1.2\%$ and $3.7\%$ for the pressure and filling-factor predictions, respectively. The DeepONet surrogate model was then embedded within EKI and compared against the EKI approach using the full forward model. The results indicate that the DeepONet-accelerated inversion yields accurate posterior estimates of reinforcement properties and process parameters, while achieving speedups of up to 200$\times$. The integration of the surrogate error within the Bayesian framework introduces a slight reduction in reconstruction quality and increase in posterior uncertainty, effects that become more pronounced when fewer sensors or smaller training datasets are used.

Lab experiments demonstrated that the DeepONet-accelerated framework can accurately characterise the shape, size and severity of defects in real reinforcements, including central inclusions and RT regions, within 42 s. Crucially, the posterior approximations for permeability and porosity are reconstructed on a high-dimensional mesh, rather than the low-dimensional representation used in the previous work \citep{MC}. Predictions of pressure at sensor locations, obtained via DeepONet surrogate simulation of approximate posterior samples, showed strong agreement with the true data recorded in each experiment. Furthermore, DeepONet simulation of the posterior mean approximations produced filling factors that closely matched flow front patterns observed in each experiment. While the lab experiments were conducted using a fixed sensor configuration, this approach can be applied to arbitrary configurations without the need to retrain the surrogate.

Despite the strong performance demonstrated in this work, several limitations should be acknowledged and addressed in future research. Although the DeepONet architecture enables evaluation of the pressure and filling factor at arbitrary spatial locations beyond those used during training, the permeability and porosity fields provided as inputs to the surrogate are restricted to the discretisation employed in the training data. When material properties are specified on a different grid, an interpolation or projection step is required prior to evaluation, which may introduce additional approximation error. In this regard, alternative neural operator architectures that naturally accommodate discretisation-independent inputs, such as Fourier Neural Operators \citep{li2021fourierneuraloperatorparametric}, represent an important direction for future work.

A natural extension of the proposed methodology is to three-dimensional domains, as well as to more complex geometries. The present surrogate, however, relies on a U-Net--based encoder to process the input fields, which is well suited to structured grids and relatively simple geometries. For more complex component geometries or unstructured meshes, this architecture may become restrictive. Geometry-aware operator-learning approaches, including graph- and mesh-based neural operators as well as geometry-informed neural operators designed to explicitly account for complex domain representations, offer promising avenues to overcome these limitations and broaden the applicability of the proposed framework \citep{10.5555/3666122.3667678,li2025geometricoperatorlearningoptimal,park2024pointdeeponetdeepoperatornetwork}.

Along these lines, it is worth noting that, for moving boundary problems, the DeepONet surrogate based on the filling-factor formulation is not the only viable approach. Alternative strategies that explicitly learn the evolution of the moving front itself may also be pursued. In particular, mesh-based graph neural network approaches, such as those recently developed for free-surface fluid flows \citep{LANTERI2025106773}, provide a promising direction for learning interface dynamics directly.

Overall, these considerations highlight the potential of operator learning as a powerful tool for constructing fast and accurate emulators for moving boundary problems, while also underscoring the need for further methodological developments. The generality of the proposed formulation makes it applicable to a wide range of moving boundary problems beyond RTM, including flow through geological and cementitious porous media \citep{1DSol}, where similar challenges in inverse modelling and flow prediction arise. The present work thus paves the way for further exploration of operator-learning-based approaches in complex, evolving-domain settings.

\section*{Data Availability}

The code used to generate the results presented in this manuscript is available at
\url{https://github.com/Marco-Iglesias-Nottingham/DeepONetEKI_RTM.git} and will be made publicly accessible upon publication.
The training data and trained PyTorch models will be deposited in the University of Nottingham’s Research Data Management Repository and made publicly available upon publication.

\section*{Acknowledgments}
This work was supported by the Engineering and Physical Sciences Research Council [grant number EP/P006701/1]; through the EPSRC Future Composites Manufacturing Research Hub. Some of the simulations were performed using the Sulis Tier 2 HPC platform hosted by the Scientific Computing Research Technology Platform at the University of Warwick. Sulis is funded by EPSRC Grant EP/T022108/1 and the HPC Midlands+ consortium. We are also grateful for access to the University of Nottingham's Ada HPC service, where some of the simulations were performed.

\appendix 
\section{Priors}\label{app: Priors}

For the scalar parameters $(K_{\mathrm{nom}}, \phi_{T}, \phi_{B}, \phi_{\mathrm{nom}}, \phi_{\mathrm{def}})$, uniform priors are assigned over the ranges reported in Table~\ref{tab:prior_scalar}. These ranges are chosen to cover the physically plausible operating envelope observed in the experimental settings. For the functional parameters $(L, \xi_{T}, \xi_{B}, \log K_{T}, \log K_{B}, \log K_{\mathrm{def}})$, Gaussian random–field (GRF) priors are used on the spatial domain. For simplicity, the geometric fields $(L,\xi_{T},\xi_{B})$ are modelled as zero–mean GRFs with isotropic Mat\'ern covariance \citep{matern1,matern2}:
\begin{equation}\label{eq:param11}
C_{\sigma,\ell,\nu}(r)
=\sigma^2\,\frac{2^{1-\nu}}{\Gamma(\nu)}
\left(\frac{\sqrt{2\nu}\,r}{\ell}\right)^{\nu}
K_{\nu}\!\left(\frac{\sqrt{2\nu}\,r}{\ell}\right),
\qquad r=\|\bfx-\bfx'\|,\qquad \bfx,\bfx'\in D,
\end{equation}
where $\sigma^2$ is the marginal variance (amplitude), $\ell$ is the correlation length, $\nu$ is the smoothness parameter, $\Gamma$ is the Gamma function, and $K_{\nu}(\cdot)$ is the modified Bessel function of the second kind of order $\nu$.

The priors for the log-permeability fields are also defined as GRFs with a hierarchical structure. Specifically, for each region $\alpha\in\{T,B,\mathrm{def}\}$, a baseline permeability level $K_{0,\alpha}$ is first drawn from a uniform distribution over the physically admissible ranges reported in Table~\ref{tab:prior_fields}. The corresponding mean level of the log-permeability field is then defined as
\[
m_\alpha = \log K_{0,\alpha}.
\]
Conditionally on $m_\alpha$, the log-permeability field is modelled as
\[
\log K_\alpha \,\big|\, m_\alpha \;\sim\; \mathcal{N}\!\big(m_\alpha,\, C_{\sigma,\ell,\nu}\big),\qquad \alpha\in\{T,B,\mathrm{def}\},
\]
so that $K_\alpha(\bfx)=\exp(\log K_\alpha(\bfx))$ is a log-normal random field. This construction enforces positivity of the permeability, allows sample–to–sample variability in the baseline (through $m$), and captures spatial correlations through the covariance $C_{\sigma,\ell,\nu}$.

The hyperparameters of the Gaussian random field priors reported in Table~\ref{tab:prior_fields} are chosen to reflect physically plausible spatial variability and geometric feature sizes consistent with typical RTM processes, while remaining compatible with the numerical resolution used in this work. In particular, the correlation length $\ell$ is selected as a fraction of the mould length $D_x$ to control the characteristic size of spatial variations. For the geometric fields $(L,\xi_T,\xi_B)$, the chosen values of $\ell$ correspond to the expected spatial extent of central defects (for $L$) and to the characteristic length scales over which the widths of the upper and lower RT boundaries vary. For the log-permeability fields, $\ell = 0.1\,D_x$ allows for moderate spatial heterogeneity within each region while avoiding unrealistically rapid spatial fluctuations.

The variance parameter $\sigma$ controls the amplitude of spatial fluctuations about the mean. For the log-permeability fields, $\sigma=0.3$ permits small variations within each region. For the geometric fields, the selected variances yield realistic magnitudes for boundary perturbations and defect shapes without inducing nonphysical geometries. The smoothness parameter $\nu$ controls the regularity of prior realisations. Since the priors are imposed on both two-dimensional fields (e.g., $L$ and the log-permeability fields) and one-dimensional boundary fields (e.g., $\xi_T$ and $\xi_B$), the same value of $\nu$ does not correspond to the same degree of smoothness across all quantities. The values of $\nu$ are chosen to balance physical realism and numerical robustness: large $\nu$ produces overly smooth, unrealistic realisations, while small $\nu$ yields overly rough fields with rapid oscillations that can cause numerical difficulties.

For the log-permeability fields, a hierarchical prior is adopted in which the mean level $m$ is drawn from a uniform distribution over physically admissible ranges informed by prior experimental data and manufacturing specifications. This construction allows for sample-to-sample variability in baseline permeability while preserving spatial correlation through the GRF structure.

\begin{table}
    \centering
    \caption{Range for priors on the scalar parameters describing permeability and porosity.}
    \begin{tabular}{c|c}
         Parameter & Range\\
         \hline
         $K_{\text{nom}}$ &  $[2\times 10^{-10},6.5\times 10^{-10}]$ $\text{m}^2$\\
         $\phi_{\text{nom}}$ &  $[0.6,0.8]$\\
         $\phi_{\text{def}}$ &  $[0.55,0.7]$\\
         $\phi_{\text{T}}, \phi_{\text{B}} $ &  $[0.9,0.96]$
    \end{tabular}
    \label{tab:prior_scalar}
\end{table}

\begin{table}
    \centering
    \caption{Hyperparameters for Gaussian random field priors. For the permeability fields, the baseline level is sampled as $K_{0,\alpha}\sim U[\cdot]$ and the mean of the log-field is set to $m_\alpha=\log K_{0,\alpha}$.}
    \begin{tabular}{c|c|c|c|c}
        & $\ell$ & $\sigma$ &$\nu$ & baseline level \\
        \hline
     $L$ & $0.075\,D_{x}$& 1.0&2.0& $m=0$\\
     $\log(K_{T})$ & $0.1\,D_{x}$& 0.3&2.0 & $K_{0,T}\sim U[2.0\times 10^{-9},5.0\times 10^{-9}]~\mathrm{m}^2$\\
     $\log(K_{B})$ & $0.1\,D_{x}$& 0.3&2.0 & $K_{0,B}\sim U[2.0\times 10^{-9},5.0\times 10^{-9}]~\mathrm{m}^2$\\
     $\log(K_{\text{def}})$ & $0.1\,D_{x}$& 0.3&2.0 & $K_{0,\text{def}}\sim U[2.5\times 10^{-11},2.5\times 10^{-10}]~\mathrm{m}^2$\\
     $\xi_{T}$  & $0.15\,D_{x}$& $0.0135$ &1.5& $m=0$\\
     $\xi_{B}$  & $0.15\,D_{x}$& $0.0135$ &1.5& $m=0$
    \end{tabular}
    \label{tab:prior_fields}
\end{table}

\section{The EKI algorithm} \label{app: EKI}

The unknown parameter vector to be inferred via ensemble Kalman inversion (EKI) is
\[
\bfu = (\bfu_{K,\phi},\,\mu,\,P_{I},\,\lambda,\,\beta,\,\chi),
\]
where the permeability–porosity parameter vector $\bfu_{K,\phi}$, defined in Eq.~\eqref{eq:param4}, is recalled here as
\[
\bfu_{K,\phi}
:= \big(\log K_{T},\,\log K_{B},\,\log K_{\text{def}},\,K_{\text{nom}},\,
\phi_{T},\,\phi_{B},\,\phi_{\text{nom}},\,\phi_{\text{def}},\,L,\,\xi_{T},\,\xi_{B}\big).
\]

For the application of EKI, it is convenient to partition the unknowns into
\emph{field-valued parameters}
\[
\boldsymbol{\Theta}
:= \big(\log K_{T},\,\log K_{B},\,\log K_{\text{def}},\,L,\,\xi_{T},\,\xi_{B}\big),
\]
and \emph{scalar parameters}
\[
\boldsymbol{\theta}
:= \big(K_{\text{nom}},\,\phi_{T},\,\phi_{B},\,\phi_{\text{nom}},\,\phi_{\text{def}},\,\mu,\,P_{I},\,\lambda,\,\beta,\,\chi\big).
\]

To ensure that the scalar parameters remain within the support of their respective uniform prior distributions during the EKI iterations, a bounded-to-unbounded reparametrisation is employed, following the approach described in Appendix~A of \citet{MC}. This transformation prevents ensemble members from leaving the region of parameter space on which the surrogate model is trained.

Specifically, for each scalar component $\theta_m \in (a_m,b_m)$, the transformation
\[
\mathcal{T}_m(\theta_m)
= \log\!\left(\frac{b_m-\theta_m}{\theta_m-a_m}\right),
\qquad
\mathcal{T}_m^{-1}(\zeta_m)
= \frac{b_m + e^{\zeta_m} a_m}{1 + e^{\zeta_m}},
\]
is applied, where $(a_m,b_m)$ denotes the support of the corresponding prior. Denoting
\[
\mathcal{T}(\boldsymbol{\theta})
= \big(\mathcal{T}_1(\theta_1),\ldots,\mathcal{T}_{10}(\theta_{10})\big),
\]
the EKI algorithm is applied to the transformed variables $(\boldsymbol{\Theta},\,\boldsymbol{\zeta})$, with $\boldsymbol{\zeta}=\mathcal{T}(\boldsymbol{\theta})$. The inferred parameters $\bfu$ are obtained by mapping back via $\mathcal{T}^{-1}$.

The ensemble Kalman inversion scheme employed in this work is summarised in Algorithm~\ref{al: EKI}. We adopt the formulation proposed by \citet{MarcoYang}, in which EKI is interpreted as an iterative algorithm that constructs a sequence of Gaussian approximations to intermediate posterior distributions within a tempering framework. The regularisation parameter $\alpha$ controls the transition between successive intermediate measures.

The initial choice of $\alpha$ follows the strategy described in \citet{MarcoYang}. In addition, we incorporate an extra damping mechanism based on a discrepancy principle introduced by \citet{Iglesias_2016}, itself inspired by the regularised Levenberg--Marquardt method of \citet{Hanke1}. This damping depends on a tuning parameter $\rho\in(0,1)$ and was found to be particularly important in the present setting, as it prevents overly abrupt updates of the level-set–based geometric parameters. For all numerical experiments reported in this work, the value $\rho=0.65$ is employed.


\begin{algorithm}
\caption{The EKI algorithm.}\label{al: EKI}
\LinesNumbered
\DontPrintSemicolon
\KwInput{Forward map $\FF(\bfu)$; Measurements $\mathbf{d}^{\dagger}\in \mathbb{R}^{MN}$;Noise covariance $\Gamma\in\mathbb{R}^{MN\times MN}$; Prior ensemble $\{\bfu^{(j)}\}_{j=1}^J$}; tuning parameter $\rho\in (0,1)$.
$n \gets 0$\;
$s \gets 0$\;
\While{$s < 1$}{
  Evaluate forward map: $\GG^{(j)} \gets  \GG(\bfu^{(j)})$ for $j \in \{1,\ldots,J\}$\;
  Transform scalar variables $\zeta^{(j)}\gets \mathcal{T}^{-1}(\bftheta^{(j)})$ (for $j \in \{1,\ldots,J\}$).\;
  \BlankLine
  Build $\omega^{(j)}=(\Theta^{(j)}, \zeta^{(j)})$
  Compute $\displaystyle\CC^{\GG\GG} = \frac{1}{J-1}\sum_{j=1}^J\big(\GG^{(j)}- \bar{\GG}\big)\otimes\big(\GG^{(j)}-\bar{\GG}\big)$ and 
  $\displaystyle\CC = \frac{1}{J-1}\sum_{j=1}^J\big(\omega - \bar{\omega}\big)\otimes\big(\GG^{(j)}-\bar{\GG}\big)$
  where $\displaystyle\bar{\GG} = \frac{1}{J}\sum_{j=1}^J \GG^{(j)}$ and $\displaystyle \bar{\omega} = \frac{1}{J}\sum_{j=1}^J \omega^{(j)}$\;
  \BlankLine
  Calculate 
  $\displaystyle\alpha = \frac{1}{MN}\frac{1}{J}\sum_{j=1}^J \lVert \Gamma^{-1/2}(\mathbf{d} - \GG^{(j)}) \rVert^2$
  \BlankLine
  \While{$\alpha\norm{\Gamma^{1/2}(\CC^{\GG\GG}+\alpha \Gamma)^{-1}(\mathbf{d}^{\dagger} - \overline{\GG})}\ge \rho\norm{\Gamma^{-1/2}(\mathbf{d}^{\dagger} - \overline{\GG})}$} {
  \BlankLine
     $\alpha\gets 2\alpha$\;
     \BlankLine
   }
  \eIf{$\displaystyle s + \frac{1}{\alpha} \geq 1$}{
  \BlankLine
      $\alpha \gets \frac{1}{1-\rf{s}}$\;
      \BlankLine
      $s\gets 1$\;      
      \BlankLine
    }{
    $s\gets s + \frac{1}{\alpha}$\;
    
    }
    \BlankLine
    Sample $\eta^{(j)} \sim N(0,\Gamma)$ \;
    \BlankLine
    Update the ensemble via $\displaystyle\omega^{(j)} \gets \omega^{(j)} + \CC(\CC^{\GG\GG}+\alpha \Gamma)^{-1}(\mathbf{d}^{\dagger} - \GG^{(j)} + \sqrt{\alpha}\eta^{(j)})$,  $j \in \{1,\ldots,J\}$\;
   \BlankLine
    Transform back: $(\Theta^{(j)},\theta^{(j)})=\bfu^{(j)}=(\Theta^{(j)},\mathcal{T}^{-1}(\zeta^{(j)})$\;
    
  }
  \BlankLine
\KwOutput{Posterior ensemble $\{\bfu^{(j)}\}_{j=1}^J$}  
\end{algorithm}

\newlength{\bibitemsep}\setlength{\bibitemsep}{.1\baselineskip } %
\newlength{\bibparskip}\setlength{\bibparskip}{0pt} \let\oldthebibliography%
\thebibliography
\renewcommand\thebibliography[1]{  \oldthebibliography{#1}
\setlength{\parskip}{\bibitemsep}  \setlength{\itemsep}{\bibparskip}}
\bibliographystyle{unsrtnat}
\bibliography{references}

\newpage
\pagenumbering{gobble}

\end{document}